\documentclass{article}

\usepackage[main, final]{NeurIPS/neurips_2026}

\usepackage[utf8]{inputenc} 
\usepackage[T1]{fontenc}    
\usepackage[hidelinks]{hyperref}
\usepackage{url}            
\usepackage{booktabs}       
\usepackage{amsfonts}       
\usepackage{nicefrac}       
\usepackage{microtype}      
\usepackage{xcolor}         
\usepackage{svg} 
\usepackage{rotating}
\usepackage{wrapfig}
\usepackage[table]{xcolor}

\usepackage{microtype}
\usepackage{graphicx}
\usepackage{subcaption}
\usepackage{booktabs} 
\usepackage{tabularx}

\usepackage{hyperref}
\usepackage{makecell}
\usepackage{adjustbox}
\usepackage{pifont}
\newcommand{\xmark}{ {\color{red}\ding{55}} }

\usepackage{fontawesome5}
\usepackage{amsmath}
\usepackage{amssymb}
\usepackage{mathtools}
\usepackage{amsthm}

\usepackage{float}     
\usepackage{placeins}  

\usepackage[capitalize,noabbrev]{cleveref}

\usepackage{listings}
\usepackage{xcolor}

\usepackage[textsize=tiny]{todonotes}
\usepackage{multirow}
\usepackage{enumitem}

\usepackage[dvipsnames]{xcolor}
\usepackage[most]{tcolorbox}
\tcbuselibrary{listings, breakable, skins}
\usepackage{listings}
\usepackage{paracol}

\usepackage{tabularx}
\newcolumntype{Y}{>{\centering\arraybackslash}X}

\lstdefinestyle{mono}{
  basicstyle       = \ttfamily\tiny,
  breaklines       = true,
  showstringspaces = false,
  tabsize          = 4,
  numbers          = none,
  xleftmargin      = 0pt,
  aboveskip        = 4pt,
  belowskip        = 0pt,
  frame            = none,
}
 
\newtcolorbox{testbox}[1][]{%
  colback=blue!3, colframe=RoyalBlue!70!black,
  coltitle=white, fonttitle=\bfseries\sffamily\scriptsize,
  title={Test Set Sample},
  left=4pt, right=4pt, top=4pt, bottom=4pt, boxrule=0.6pt,
  skin=enhanced jigsaw, breakable,
  #1}
 
\newtcolorbox{corpusbox}[1][]{%
  colback=orange!3, colframe=BurntOrange!80!black,
  coltitle=white, fonttitle=\bfseries\sffamily\scriptsize,
  title={Corpus Sample},
  left=4pt, right=4pt, top=4pt, bottom=4pt, boxrule=0.6pt,
  skin=enhanced jigsaw, breakable,
  #1}

\usepackage{CJKutf8}

\title{Soft Contamination Means Benchmarks Test \\ Shallow Generalization}

\author{%
  Ari Spiesberger\thanks{Equal contribution.} \\
  Arb Research \\
  {\footnotesize\texttt{arispiesberger@gmail.com}} \\
  \And
  Juan J Vazquez\footnotemark[1] \\
  Arb Research \\
  {\footnotesize\texttt{juan@arbresearch.com}} \\
  \And
  Nicky Pochinkov \\
  Arb Research \\
  \And
  Tomáš Gavenčiak \\
  Charles University \\
  \AND
  Peli Grietzer \\
  Arb Research \\
  \And
  Gavin Leech\thanks{Joint supervision.} \\
  Arb Research \\
  \And
  Nandi Schoots\footnotemark[2] \\
  University of Oxford \\
}

\begin{document}

\maketitle

\begin{abstract}
Training on benchmark test items is a classic way to overestimate generalization to unseen problems. Standard decontamination methods miss \textit{semantic duplicates} (SemDs): training items equivalent in content to a test item, but distant in string space. We call training on such items \textit{soft contamination}; when the resulting gains extend to held-out items of the same benchmark (but not to related benchmarks), we call the result \textit{shallow generalization}. We first embed Olmo 3's entire post-training dataset and search it for SemDs of benchmark items; we then finetune Olmo 3 and Qwen models on natural and synthetic semantic duplicates. We find (1) that soft contamination is widespread: the Olmo corpus contains SemDs of 76\% of the CodeForces test set and 100\% of MBPP's; (2) that finetuning on SemDs of a benchmark raises accuracy on its held-out items by up to 22pp in controlled experiments and 10--11pp at realistic contamination rates, though the effect varies across benchmarks; (3) that the gains from training on SemDs mostly do not transfer to conceptually similar benchmarks, i.e. we find shallow generalization; (4) that finetuning on items with similar embeddings that are not SemDs produces no such gain, indicating that the mechanism is not mere topical similarity; and (5) that at realistic contamination rates, soft contamination can raise held-out performance about as much as exact duplicates (or more). Benchmark gains here are therefore confounded - in a way that standard decontamination does not address.
\end{abstract}

\section{Introduction}

LLMs have rapidly solved many hard reasoning benchmarks, some of which are nearing saturation even for smaller models \citep{epoch2025aicapabilitiesprogresshasspedup,maslej2025artificialintelligenceindexreport}. But LLM training corpora have also grown by a factor of thousands since 2020 \citep{EpochAIModels2025}. As corpora grow - in the absence of proportionate filtering effort - we should expect more benchmark test items to be included in them. While AI labs make good-faith efforts to remove syntactic duplicates of benchmark items from their corpora \citep{openai2024gpt4technicalreport,olmo2025olmo3, anthropic2025opus45systemcard}, `softer' forms of contamination  are  hard to detect. These `semantic duplicates' are data distant from test data in string space, but (as we will show) still containing learnable structure about the test data. 
This paper evaluates this potential confounder of benchmark scores.\footnote{The question
is sharpened by the ``approximate retrieval'' view of LLMs \citep{kambhampati2023tweet, kambhampati2024llms, meursault2025approximate}, on which some outputs are similarity-weighted combinations of nearby training documents.}

We combine existing contamination detection methods (embedding search and LLM judges) with novel finetuning experiments. When models train on semantic duplicates of some benchmark items, their accuracy on the benchmark's held-out items can rise too, while their accuracy on related control benchmarks mostly does not: \textit{shallow generalization} from semantic duplicates of test data to test data. Using the open-data model Olmo 3, we show that modern LLM training corpora include data that are \textit{effectively} samples from major reasoning benchmark test sets, which would make benchmark scores partly a measure of shallow generalization - rather than of the intended general reasoning capability. We hypothesize that increased LLM performance on reasoning benchmarks partly reflects an increase in soft contamination. 
\begin{table*}[t]
\centering
\scriptsize
\setlength{\tabcolsep}{4pt}
\caption{An example of a semantic duplicate of test data found by our pipeline.}
\renewcommand*{\arraystretch}{0.5}
\begin{tabular}{ll}
\toprule
\textbf{Original MBPP sample $297$}
& \textbf{Semantic duplicate from \texttt{dolmino}} \\
\midrule
\makecell[l]{
\emph{Write a function to flatten a given nested list structure.} \\[4pt]
\texttt{def flatten\_list(list1):} \\
\texttt{\ \ \ \ result\_list = []} \\
\texttt{\ \ \ \ if not list1: return result\_list} \\
\texttt{\ \ \ \ stack = [list(list1)]} \\
\texttt{\ \ \ \ while stack:} \\
\texttt{\ \ \ \ \ \ \ \ c\_num = stack.pop()} \\
\texttt{\ \ \ \ \ \ \ \ next = c\_num.pop()} \\
\texttt{\ \ \ \ \ \ \ \ if c\_num: stack.append(c\_num)} \\
\texttt{\ \ \ \ \ \ \ \ if isinstance(next, list):} \\
\texttt{\ \ \ \ \ \ \ \ \ \ \ \ if next: stack.append(list(next))} \\
\texttt{\ \ \ \ \ \ \ \ else: result\_list.append(next)} \\
\texttt{\ \ \ \ result\_list.reverse()} \\
\texttt{\ \ \ \ return result\_list}
}
&
\makecell[l]{
\texttt{:param nested\_list: A list that} \\
\texttt{\ \ \ \ may contain other lists} \\
\texttt{:return: A flattened list of} \\
\texttt{\ \ \ \ all elements} \\
\texttt{"""} \\[4pt]
\texttt{flattened\_list = []} \\
\texttt{for element in nested\_list:} \\
\texttt{\ \ \ \ if isinstance(element, list):} \\
\texttt{\ \ \ \ \ \ \ \ \# Recursively flatten the sublist} \\
\texttt{\ \ \ \ \ \ \ \ flattened\_list.extend(} \\
\texttt{\ \ \ \ \ \ \ \ \ \ \ \ flatten\_nested\_list(element))} \\
\texttt{\ \ \ \ else:} \\
\texttt{\ \ \ \ \ \ \ \ \# Append non-list elements} \\
\texttt{\ \ \ \ \ \ \ \ flattened\_list.append(element)} \\
\texttt{return flattened\_list}
} \\
\bottomrule
\end{tabular}
\label{tab:eg}
\end{table*}
%


We define an \textit{exact} duplicate of test data to be a training datapoint which contains a test set example verbatim, up to formatting. A \textit{semantic} duplicate of test data is an example in the training corpus that shares the algorithmic core or logical structure of a test item - despite different surface framing, categories or wording \citep{riddell-etal-2024-quantifying} - so that learning to solve it suffices to solve the test item. Our LLM judges count a pair as semantic duplicates when the training example states the same \textit{task} as the test item, or a task that subsumes it, or if it supplies the insight for the test problem (Appendix~\ref{app:annot}).
We call contamination \textit{soft} when it involves semantic duplicates. We call generalization \textit{shallow} when gains from training on duplicates of some benchmark items extend to the benchmark's held-out items, but \textit{not} to related benchmarks.


Our contributions:

$\bullet$ \textbf{High rates of soft contamination in the wild:} We use embedding distances to screen for natural semantic duplicates in 1\% of the pretraining data and all of the finetuning data of Olmo 3 -- around 260GB of text, significantly more than previous studies. Despite the industry-standard decontamination performed by the Olmo team, we find more contamination than previous studies.

$\bullet$ \textbf{Shallow generalization:} We finetune Olmo 3 on a random split of MuSR, ZebraLogic, and MBPP. We find that performance improves on the unseen half, while control benchmarks drawn from the same domain generally do not improve. 
On MuSR, both semantic and exact duplicates add $\sim$20 pp to seen and unseen items alike. On MBPP, exact duplicates help seen items more, (17pp vs 9pp), while semantic duplicates help unseen items more (7pp vs 2pp). 

$\bullet$ \textbf{Duplication drives gains, not mere similarity:} Finetuning on the corpus items closest to benchmark items in embedding space that are not duplicates does not improve benchmark scores (Table~\ref{tab:big_deltas}), so topical similarity alone does not explain the gains. 

$\bullet$ \textbf{Natural levels of soft contamination have a substantial effect:} We estimate the natural rate of semantic duplication: 4 items in 10,000. We run the first ecologically-valid finetuning study of exposure to semantic duplicates: at this natural rate, synthetic semantic duplicates inflate scores on held-out MuSR items by 10--11pp ($p < 0.05$) in Olmo 3, Qwen3 and Qwen3.5. 

$\bullet$ \textbf{True hold-out has never been tried:}  We view this work as a step towards evaluating LLMs in the empirical risk minimization setting \citep{vapnik1991principles}: a strict separation between training and test sets, verified on open training data. 


\section{Related Work}

\begin{table*}[t]
\centering
\caption{\textbf{The literature on data contamination by semantic duplicates}, whether it: studies \textit{semantic} duplicates of test data ( \faIcon[regular]{clone} ); estimates the prevalence of contamination in training corpora (\faIcon[regular]{divide}); has automated screening for semantic overlap ( \faIcon[regular]{layer-group} ); quantifies the effects of contamination on downstream performance (\faIcon[regular]{ruler}); employs finetuning (\faIcon[regular]{wrench}) and whether the FT data are ecologically valid (\faIcon[regular]{globe}); tests \textit{in-benchmark generalization} (whether training on duplicates of some benchmark items improves performance on held-out items,\faIcon[regular]{hubspot}); and whether it does training data attribution ( \faIcon[regular]{file} ).}

\resizebox{\textwidth}{!}{
\renewcommand*{\arraystretch}{0.5}
\begin{tabular}{l c c c c c c c c c c l}
\toprule
\textbf{Paper} &
\makecell{\faIcon[regular]{clone}} & 
\makecell{\faIcon[regular]{divide}} & 
\makecell{\faIcon[regular]{layer-group}} &
\makecell{\faIcon[regular]{ruler}} & 
\makecell{\faIcon[regular]{wrench}} & 
\makecell{\faIcon[regular]{globe}} & 
\makecell{\faIcon[regular]{hubspot}} & 
\makecell{\faIcon[regular]{file}} & 
\makecell{\textbf{Data}\\\textbf{scale}} &
\makecell{\textbf{Model}\\\textbf{scale}} &
\makecell{\textbf{Major}\\\textbf{evals}} \\
\midrule
\textit{Ours}
& \checkmark
& \checkmark
& \checkmark
& \checkmark
& \checkmark
& \checkmark
& \checkmark
& \checkmark
& 260G
& \textit{7B}
& \textit{CodeForces, MBPP (et al.)} \\
\midrule
\citet{magar2022data}
& \xmark
& \xmark
& -
& \checkmark
& \checkmark
& \xmark
& \xmark
& \checkmark
& 2.5G
& 0.3B
& SST-5, SST-2, SNLI \\
\midrule
\citet{yang2023rethinking}
& \checkmark
& \checkmark
& \checkmark
& \checkmark
& \checkmark
& \xmark
& \xmark
& \checkmark
& 23G
& 13B
& MMLU, GSM8K \\
\midrule
\citet{riddell-etal-2024-quantifying}
& \checkmark
& \checkmark
& \checkmark
& \checkmark
& \xmark
& -
& \xmark
& \checkmark
& 155G
& 16B
& MBPP, HumanEval \\
\midrule
\citet{shilov2025mosaic}
& \xmark
& \xmark
& -
& \checkmark
& \checkmark
& \xmark
& \xmark
& \xmark
& 45G
& 2.7B
& MIA \\
\midrule
\citet{xu-etal-2025-ssa}
& \checkmark
& \xmark
& \checkmark
& \checkmark
& \xmark
& -
& \xmark
& \xmark
& 1G
& 72B
& LIAR2 \\
\bottomrule
\end{tabular}
}
\label{relateddiff}
\end{table*}



Contamination of LLM training corpora by benchmark test items is a large field. Table~\ref{relateddiff} covers key differences between our work and prior studies. \citet{yang2023rethinking}, \citet{riddell-etal-2024-quantifying}, and \citet{xu-etal-2025-ssa} establish that semantic duplicates exist and raise the scores of duplicated items. We differ from them as follows: 1) Our definition of duplication is \textit{functional}: a training item $x$ is a duplicate of a test item $t$ if learning to solve $x$ suffices to solve $t$. This admits training items that pose a more general problem solved by the same algorithm, which other definitions exclude. 2) We search a larger corpus and estimate natural contamination rates. 3) We evaluate on held-out benchmark items and against control benchmarks. 

The first wave of contamination studies \citep{magar2022data, elazar2024whats,
jiang2024investigating, dong2024generalization} focused on the prevalence and impact of exact syntactic duplicates (i.e., word-for-word matches in string space), with later work \citep{dekoninck2024constat, zhou2025lessleak,shilov2025mosaic} extending the analysis to partial syntactic duplicates. More recently, researchers have begun to study indirect or `semantic' contamination \citep{yang2023rethinking,riddell-etal-2024-quantifying,xu-etal-2025-ssa}, where data in the training corpora are equivalent to benchmark-items in substantive content without sharing $n$-gram sequences or other syntactical properties. 

The literature often treats semantic contamination as a form of memorization, focusing on how exposure in training to near-duplicates improves a model's performance on those specific benchmark items. Our work instead studies semantic duplicates as a source of \textit{shallow generalization}: gains that extend from duplicated benchmark items to \textit{unduplicated} items from the same benchmark, but not to related benchmarks. Because the gains reach unduplicated items, we still call this `generalization'. \citet{dominguez2024training} study a related phenomenon they call `training on the test \textit{task}'.




To identify exact duplicates of benchmark data, prior work deploys two approaches: searching for duplicates of benchmark items in open datasets \citep{magar2022data, elazar2024whats,
jiang2024investigating, dong2024generalization}, and testing for memorization using membership inference techniques \citep{shi2023detecting}. But memorization tests (which typically depend on matching strings) are unlikely to capture the effect of exposure to semantic duplicates, and so searching open datasets is preferred. To find semantic duplicates, a heuristic semantic distance is often used to guide the search, alongside human judgment, LLM judges, or plagiarism-detection software. Previous work by \citet{yang2023rethinking} and \citet{riddell-etal-2024-quantifying} has provided high-quality estimates of the prevalence of semantic duplicates of items from major reasoning benchmarks - including HumanEval, MMLU, and GSM8K \citep{chen2021evaluating,hendrycks2020measuring,cobbe2021training} - in widely-used training corpora, e.g. The Pile, StarCoderData, and RedPajama \citep{gao2020pile,li2023starcoder,weber2024redpajama}.


To measure the effect of contamination, prior work either 1) finetunes on contaminated data to simulate exposure \citep{yang2023rethinking} or 2) correlates how often items are duplicated with model performance on them, i.e.\ finds the `dosage' of exact duplicates \citep{riddell-etal-2024-quantifying,pmlr-v267-kocyigit25a}. We combine the two approaches by finetuning on semantic duplicates at natural rates and testing the gains on held-out benchmark items. 


\section{Methodology}

\begin{figure*}[!h]
    \centering
\includegraphics[width=\linewidth]{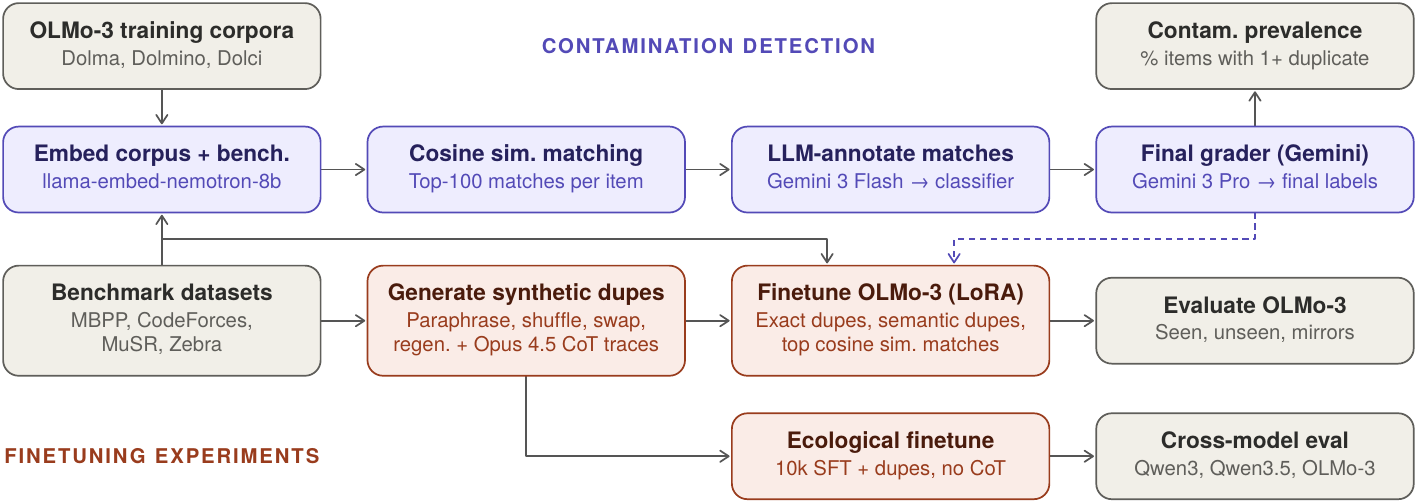}
    \caption{\textbf{Our full pipeline from input data to experimental outcomes.}}
    \label{fig:method_pipeline}
\end{figure*}

Figure~\ref{fig:method_pipeline} shows our full pipeline. To know what exactly the model has seen in training, our main experiments use \texttt{Olmo-3-7B}, an open-\textit{data} model \citep{olmo2025olmo3}. We use the closed-data Qwen3 and Qwen3.5 only in our ecological experiment (Section~\ref{sec:ecological}), where the clean and contaminated runs start from the same model (so that any prior exposure will affect both runs). Appendix~\ref{app:benchmarks} describes our four benchmarks: we search the Olmo 3 corpus for duplicates of all four, and finetune on duplicates of MuSR, ZebraLogic and MBPP. All code on GitHub: 

\begin{center}
\texttt{\href{https://github.com/AriSpiesberger/Soft-Contamination-Prevelance}{https://github.com/AriSpiesberger/Soft-Contamination-Prevelance}}
\end{center}

\subsection{Finding semantic duplicates in the wild}\label{sec:find-sem-dupes}
We find semantic duplicates in the wild in two stages: 1) we retrieve the training items with the highest cosine similarity to each benchmark item, then 2) we annotate a sample of these matches with an LLM judge. The second stage is needed because high cosine similarity does not imply algorithmic duplication. 

\textbf{Olmo 3 corpus}. 
We embed a 1\% random sample of Olmo 3's pretraining mix (Dolma 3 and Dolmino) and all of the Olmo 3 post-training datasets (Dolci SFT, Dolci Instruct DPO and Dolci Instruct RL). To sample the pretraining data, we use stratified reservoir sampling, which preserves the corpus's hierarchical structure. 

See Appendix \ref{app:olmo3-data} for more dataset details.

\textbf{Embeddings}. 
We embed the above datasets with \texttt{llama-embed-nemotron-8B} \citep{babakhin2025llama}, which was ranked second on the Massive Text Embedding Benchmark (MTEB) leaderboard \citep{muennighoff2023mteb} at the time of writing.
Embedding 260GB of chunked raw text at FP16 precision took about two days on 8 H100 GPUs. We describe our end-to-end discovery of semantic duplicates below.

\textbf{Cosine similarity}. 
To find candidate duplicates, we compute the cosine similarity between each benchmark item and every embedded corpus item (Appendix~\ref{app:cos-sim}).

\textbf{Semantic and exact duplicates among high cosine similarity matches}. For MuSR, the highest cosine similarity is around $0.4$, and on manual review we find no duplicates or MuSR-like problems in the post-training data or in the pretraining sample.
For ZebraLogic, we find many exact duplicates and a few semantic duplicates; most other high-similarity matches are easy `Zebra puzzles', far simpler than ZebraLogic items.
For MBPP and CodeForces, after manually inspecting the top matches, we sample matches and annotate them with an LLM.

\emph{Finding exact duplicates of the ZebraLogic test set in Dolma.}
We observed several ZebraLogic tasks in the training corpus, verbatim. We discuss this auxiliary result in Appendix~\ref{app:exact_zebra}.

\textbf{LLM annotation among high cosine similarity matches.}
For each MBPP and CodeForces item and each of the five training datasets, we take candidates in the top 0.1\% of cosine similarity, and from these either the top 100 or a random sample of 100. We prompt \texttt{gemini-3-flash-preview} \citep{google2025gemini3flash} to classify each pair. It outputs a boolean isSemanticDuplicate label, a relationship type and an explanation (definitions in Appendix~\ref{app:human-validation}). 

To validate this pipeline, we manually annotate a stratified sample of 190 items across CodeForces and MBPP (released on GitHub; also Appendix~\ref{app:human-validation}).
The judge counts at least 96\% of the pairs we label equivalent or exact as duplicates, misses 30.4\% of the pairs we label as CodeForces subsets and 16.5\% of MBPP subsets, and counts at most 5.5\% of our non-duplicate pairs as duplicates (Appendix~\ref{app:human-validation}). Our prevalence estimates are thus lower bounds. To reduce false positives, we run a second pass with \texttt{gemini-3-pro} on pairs flagged by the first judge, checking if the corpus item supplies the algorithm needed to solve the test problem.
Appendix~\ref{app:more-sem-dupe-examples} gives examples of semantic duplicates found by our pipeline, and Appendix~\ref{app:annot} details the annotation process. Appendix~\ref{app:synthetic-data} covers our generation of synthetic semantic duplicates for each benchmark.

\subsection{Finetuning on Duplicates}\label{sec:finetune-dupes}

Our experiments distinguish three kinds of gains: 1) gains on a benchmark item from training on its own semantic duplicates \citep{yang2023rethinking,riddell-etal-2024-quantifying}; 2) gains on a benchmark item from training on exact duplicates of \textit{other} items in the benchmark; and 3) gains on a benchmark item from training on semantic duplicates of other items in the benchmark. Gains of kinds 2 and 3 are shallow generalization when they do not extend to related benchmarks.

We finetune Olmo 3 Instruct on either exact duplicates or synthetic semantic duplicates (Appendix~\ref{app:synthetic-data}) of MuSR, ZebraLogic and MBPP items. Because Olmo 3 Instruct is a non-reasoning model, we first have a teacher model, Claude Opus 4.5, generate chain-of-thought (CoT) reasoning traces \citep{wei2022chain} for these items (Appendix~\ref{app:finetuning-results}). We finetune Olmo 3 Instruct with LoRA \citep{DBLP:conf/iclr/HuSWALWWC22} on the formatted questions and their CoT answers (Appendix~\ref{app:lora-defense}).
To estimate accuracy, we sample 8 generations per item at temperature 0.7 from the base model, the teacher and the finetuned model.

We split each benchmark in half, finetune on duplicates of one half (the \textit{seen} items), and evaluate on the original items of both halves (\textit{seen} and \textit{unseen}).
To test whether gains extend to related benchmarks, we use control benchmarks: TrueDetective \citep{del2023true} for MuSR, ARC-Challenge \citep{clark2018think} for ZebraLogic and HumanEval \citep{chen2021evaluating} for MBPP. These controls are imperfect. Olmo 3 scores near chance on TrueDetective (28\%, against about 24\% for random guessing), and ARC-Challenge tests science questions rather than logic puzzles.
We also evaluate on ARC-Easy, BoolQ \citep{clark2019boolq}, HellaSwag \citep{zellers2019hellaswag}, PIQA \citep{bisk2020piqa} and WinoGrande \citep{sakaguchi2021winogrande} to check for any accidental degradation in off-target capabilities.


\section{Results}

\subsection{Decontamination methods fail}
Consistent with \citet{yang2023rethinking}, we find that standard decontamination methods do not catch soft contamination. We apply the decontamination methods from recent model releases, plus standard baselines, to the semantic duplicate pairs we found (Table~\ref{tab:baselines}). No single method removes more than 6.5\% of them.

\begin{table}[t]
  \centering
  \caption{\textbf{Standard decontamination removes few of the semantic
  duplicates we find.} Percentage of confirmed in-the-wild duplicate pairs
  removed when we reproduce published decontamination pipelines and standard
  baselines. The most effective method removes 6.5\%; all methods
  combined remove 10.87\% (CodeForces) and 0.57\% (MBPP). The code-clone and
  code-embedding methods are not run on CodeForces, whose samples are mostly
  text. See Appendix~\ref{app:decontam_methods} for a more complete overview of known decontamination pipelines from popular model releases.}
  \label{tab:baselines}
  \small
  \begin{tabular}{lcc}
    \toprule
    Method & CodeForces & MBPP \\
    \midrule
    Opus 4.5 (full pipeline) (\cite{anthropic2025opus45systemcard}, \S2.2)      & 0.78\% & 0.04\% \\
    GPT-4 50-char substring match~\citep{openai2024gpt4technicalreport}          & 6.52\% & 0.27\% \\
    GPT-3: 13-gram overlap~\citep{brown2020language}                   & 6.09\% & 0.10\% \\
    Olmo 3 (full pipeline) (\cite{olmo2025olmo3}, \S3.5.3)           & 5.83\% & 0.10\% \\
    MinHash (Jaccard $\geq$ 0.8) + 13-gram containment             & 0.09\% & 0.00\% \\
    Code-clone: MinHash + Type-2 + CodeBLEU  \citep{ren2020codebleu}                      & n/a    & 0.21\% \\
    Code embedding: UniXcoder ($\geq$ 0.8) \citep{guo2022unixcoder}                         & n/a    & 0.20\% \\
    \midrule
    All combined                                                   & 10.87\% & 0.57\% \\
    \bottomrule
  \end{tabular}
  \label{tab:decontamintion_methods}
\end{table}

\subsection{There are exact duplicates in training corpora}

Surprisingly, we found that the Olmo Instruct RL dataset contains exact duplicates for ZebraLogic, one of the benchmarks the Olmo team reports: at least 49.5\% of its test set (Appendix~\ref{app:exact_zebra}). (The Olmo team have since fixed this error.)

\subsection{There are natural semantic duplicates in training corpora}\label{sec:results-inthewild}
\begin{figure}[b!]
  \centering
  \begin{subfigure}[b]{0.44\columnwidth}
        \includegraphics[width=\linewidth]{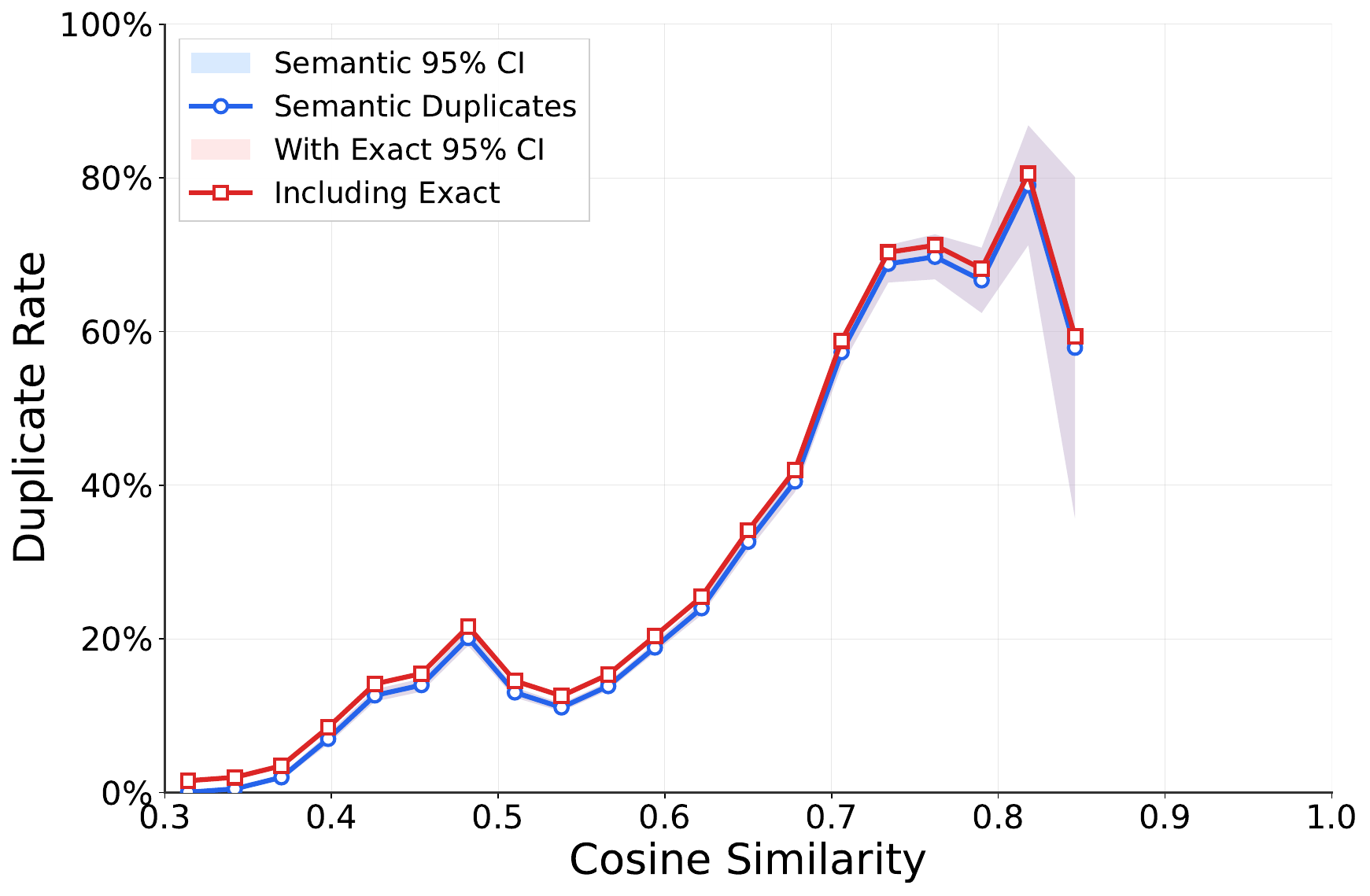}
    \caption{MBPP}
    \label{fig:mbpp-corr-cos-sim-sem-dupe}
  \end{subfigure}
  \hfill 
  \begin{subfigure}[b]{0.44\columnwidth}
    \includegraphics[width=\linewidth]{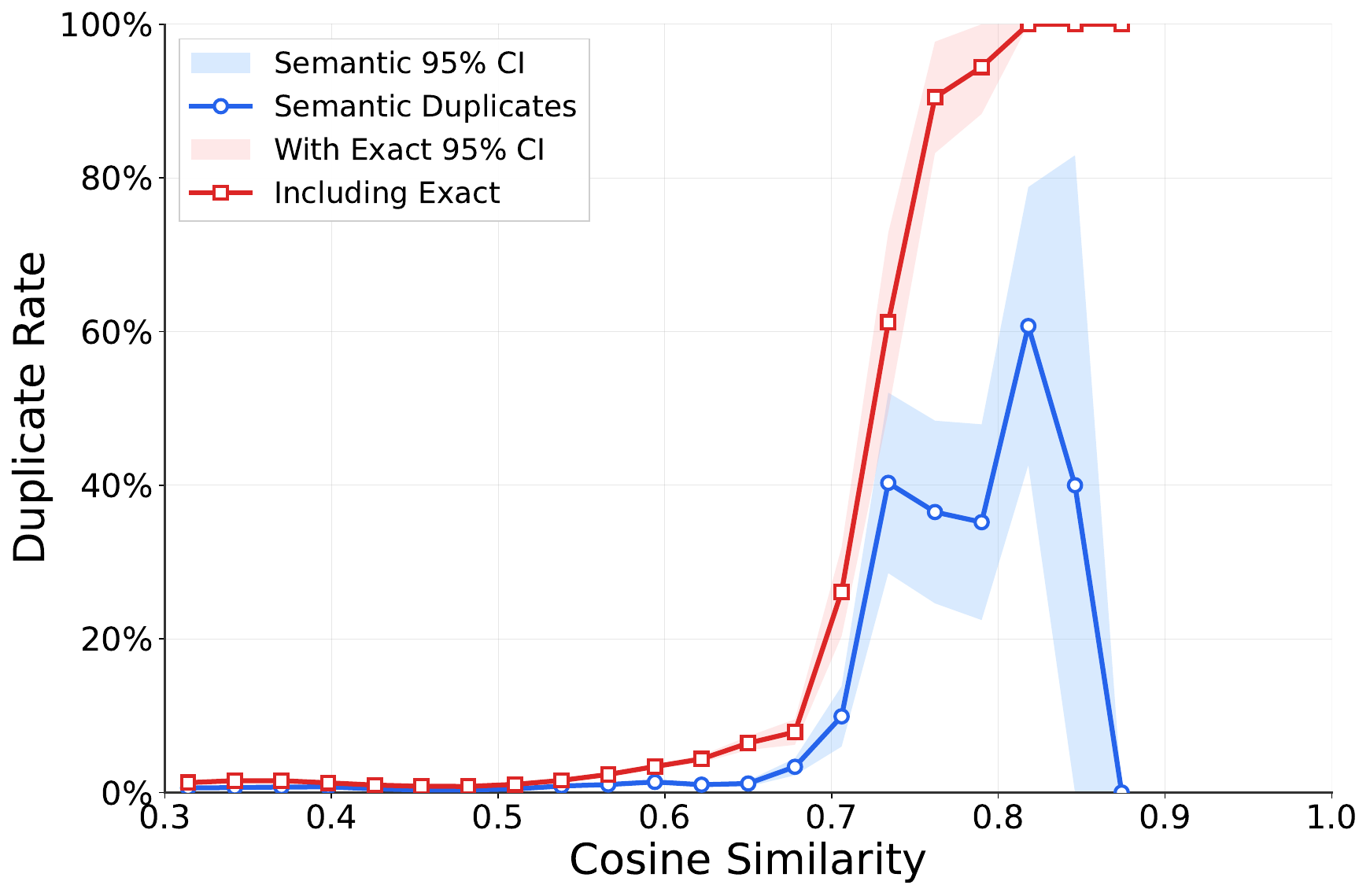}
    \caption{CodeForces}
    \label{fig:codeforces-corr-cos-sim-sem-dupe}
  \end{subfigure}
  \caption{\textbf{The relationship between cosine similarity and rates of semantic duplicates.}
  For each benchmark datapoint we sample 100 matches from the top 0.1\% cosine similarity matches in the training data.
  We plot the fraction of cosine similarity matches that are true semantic duplicates.  
  the rate of semantic \textit{or} exact duplicates are plotted in {\color{red}red}, while {\color{blue}blue} is the semantic-duplicate-only rate.
  }
  \label{fig:corr-cos-sim-vs-sem-dupe}
\end{figure}

\textbf{Relationship of cosine similarity to semantic duplicate status.}
Figure~\ref{fig:corr-cos-sim-vs-sem-dupe} plots the rate of semantic duplicates
across cosine similarity bands, for pairs sampled from the top 0.1\% of cosine similarity. The duplicate rate rises steeply with similarity: it is near zero at the low end and reaches 60--85\% above 0.8. This supports restricting LLM annotation to the extreme right tail, but makes our prevalence figures lower bounds (since duplicates below the cut are not counted).

\textbf{Benchmark examples with at least one semantic duplicate.}
Every MBPP problem has at least one semantic duplicate among its top 100 matches per training dataset, and 99.6\% have one among 100 matches sampled from the top 0.1\%.
For CodeForces, 75.9\% of problems have at least one semantic duplicate among their top 100 matches.
\textit{Contamination by semantic duplicates is therefore widespread.}
These figures are lower bounds: checking beyond the top 100 matches would very likely find more semantic duplicates for CodeForces.
Easier CodeForces problems (those with lower Elo ratings) are more likely to have a semantic duplicate (Appendix~\ref{app:cos-sim}).


\begin{figure}[ht]
  \centering
  \begin{subfigure}[b]{0.44\columnwidth}
    \includegraphics[width=\linewidth]{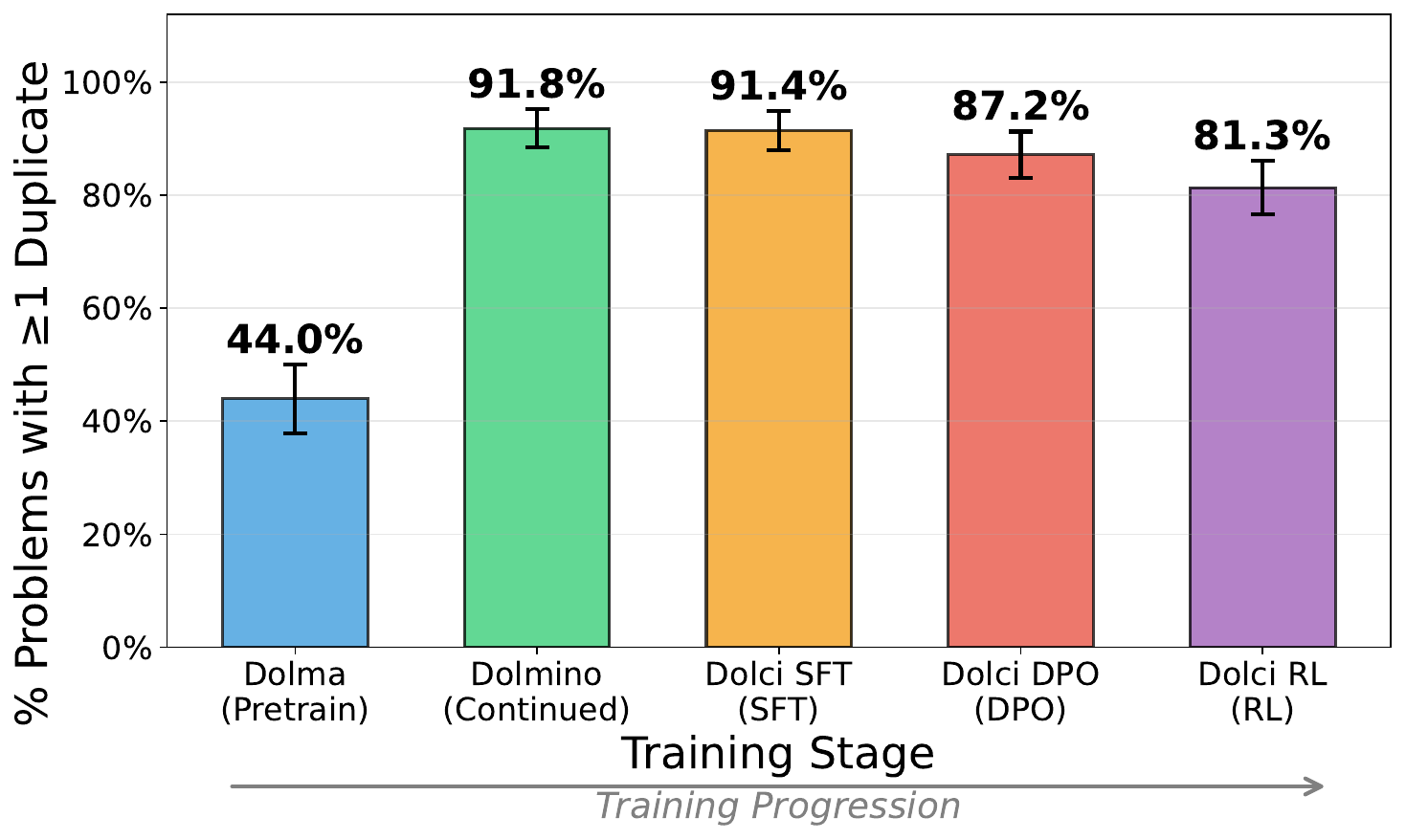}
    \caption{MBPP}
    \label{fig:mbpp-sem-dupe-propensity-training-scheme}
  \end{subfigure}
  \hfill 
  \begin{subfigure}[b]{0.44\columnwidth}
  \includegraphics[width=\linewidth]{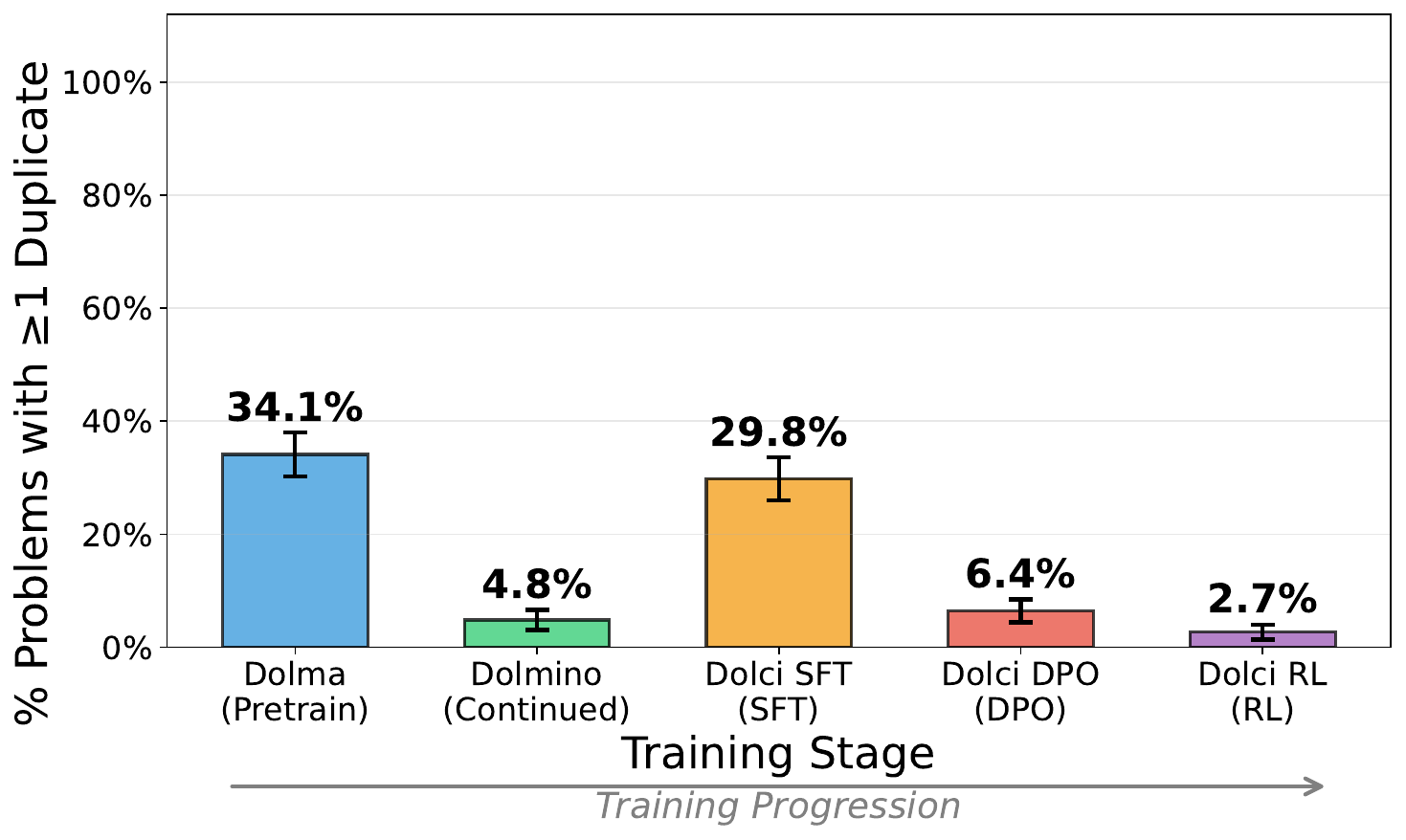}
    \caption{CodeForces}
    \label{fig:codeforces-sem-dupe-propensity-training-scheme}
  \end{subfigure}
  \caption{\textbf{Occurrence of semantic duplicates by Olmo training phase for two benchmarks: (a) MBPP, (b) CodeForces.} 
  For each benchmark sample, we review the top 100 cosine similarity training datapoints to identify if any of the samples are semantic duplicates. We then calculate the proportion of all benchmark datapoints that have at least one semantic duplicate.
  Error bars are $\pm 1$ std.
  }
  \label{fig:propensity-training-scheme}
\end{figure}

\textbf{Semantic duplicate occurrence stratified by training dataset.}
In Figure \ref{fig:propensity-training-scheme}, we investigate how the prevalence of semantic duplicates varies across different stages of the training pipeline.
Figure \ref{fig:mbpp-sem-dupe-propensity-training-scheme} shows that all training datasets have semantic duplicates for MBPP. 
In Figure \ref{fig:codeforces-sem-dupe-propensity-training-scheme},  we find a similar effect for CodeForces: semantic duplicates of problems exist in each dataset, particularly in Dolma, Dolci SFT, and Dolci DPO. 

\textbf{Retrieving more matches finds more duplicates.}
Semantic duplicates are rare within the corpus, even though most test items have at least one such duplicate.
In Appendix~\ref{app:cos-sim}, we show that the proportion of benchmark problems with at least one semantic duplicate grows with the number of matches we check per item.
For instance, only 28.4\% of CodeForces problems have a semantic duplicate as their top-1 match. Searching deeper, and in a larger corpus than previous studies (Table~\ref{relateddiff}), may explain why we find more semantic duplicates; differences in benchmarks and in how we define a duplicate may also contribute. 

\subsection{Finetuning on Semantic Duplicates }\label{sec:finetuning-sem-dupes}





\textbf{MuSR. }
We experimented with three levels of synthetic semantic duplication, with increasing dissimilarity, generated as detailed in Appendix~\ref{app:synthetic-data}.
Table \ref{tab:big_deltas} shows that when we finetune on duplicates of half of the benchmark ($n = 125$), accuracy rises equally on the seen and unseen halves.
Finetuning on exact duplicates of MuSR benchmark data leads to similar gains as finetuning on a variety of semantic duplicates: in both cases accuracy rises by about 20pp. 
Finetuning instead on the corpus items with the highest cosine similarity to MuSR items barely changes accuracy.
Accuracy on the TrueDetective control stays flat, though Olmo 3 scores near chance on it (Section~\ref{sec:finetune-dupes}).
Appendix \ref{app:finetuning-results} also shows that using a better teacher model to generate CoT reasoning traces results in further gains for finetuning.

\textbf{ZebraLogic. }
Table \ref{tab:big_deltas} shows that finetuning on exact duplicates of half of the benchmark ($n = 500$) raises accuracy on both halves (seen +11.5pp, unseen +6.5pp). Unlike on MuSR, semantic duplicates barely help ZebraLogic solutions, and one combination (shuffling, substitution and paraphrasing) actually lowers accuracy by 8.5--8.9pp on both halves, while the ARC-Challenge control-benchmark score stays flat. We tentatively attribute this to compounding distribution shift in both the puzzles and their CoT traces: ZebraLogic's substitution operates on domains (e.g., color→shape) in a way MuSR's tree-based regeneration does not, so the teacher's reasoning traces shift accordingly, and stacking all three transformations likely pushes training too far from the test distribution to support transfer.
Finetuning on the most cosine-similar corpus items lowers accuracy from 36.9\% to 22.9\%.

(These results come with a caveat: Olmo 3 Instruct's RL data already contains exact duplicates of about half of ZebraLogic, see Appendix~\ref{app:exact_zebra}: many `unseen' items were thus in the model's training data before our finetuning.)

\textbf{MBPP. }
Table \ref{tab:big_deltas} shows that finetuning on exact duplicates of half of the benchmark ($n = 210$) raises accuracy on that half by 16.6pp, while accuracy on the unseen half barely changes (+2.4pp).
Semantic duplicates have a smaller effect on seen items (+8.7pp) but also raise unseen accuracy (+7.2pp).
Finetuning on cosine-similar non-duplicates does not substantially improve accuracy.
Unlike on the other two benchmarks, semantic duplicates of MBPP also raise accuracy on the HumanEval control, from 55.3\% to 67.0\%.
As with ZebraLogic, the model was already contaminated: the Olmo 3 corpus contains semantic duplicates of every MBPP test item (Section~\ref{sec:results-inthewild}).

\begin{table}[h]
\centering
\caption{\textbf{Finetuning Olmo 3 on semantic duplicates}. We see large gains from training on semantic duplicates, in MuSR's case statistically equivalent to training on exact duplicates (See Section~\ref{app:synthetic-data} for synthetic generation details).   The rightmost column is the control, showing accuracy on second similar-domain benchmarks: for MuSR this is `True Detective' \citep{del2023true}; for ZebraLogic this is the ARC Challenge \citep{clark2018think} and for MBPP this is HumanEval \citep{chen2021evaluating}. We see no significant improvement to the control scores when finetuning on any setting of MuSR or ZebraLogic duplicates; we do see a surprising leap in HumanEval (MBPP's control) for semantic duplicates.  Accuracy (\%) $\pm$ stderr. Abbreviations: P=Paraphrase, SH=Shuffle, SU=Substitution, $\cos$ similarity=high cosine similarity corpus matches. Experiments were performed on 8 A100 GPUs, each experiment taking around 2-3 hours.}

\small
\setlength{\tabcolsep}{5pt}
\renewcommand{\arraystretch}{1.05}

\textbf{MuSR}\par\smallskip
\begin{tabularx}{\linewidth}{p{1.25in}YYY}
\toprule
Degree of similarity & Seen accuracy & Unseen accuracy & Acc. on TrueDetective \\
\midrule
No finetuning & -- & $66.0 \pm 3.0$ & $28.3 \pm 3.3$ \\
\midrule
Exact dupes & $87.9 \pm 2.1$ & $87.3 \pm 2.1$ & $27.7 \pm 3.2$ \\
Sem. dupes {\footnotesize (regen)} & $85.8 \pm 2.2$ & $86.2 \pm 2.2$ & $29.3 \pm 3.3$ \\
Sem. dupes {\footnotesize (1-branch)} & $85.7 \pm 2.2$ & $86.0 \pm 2.2$ & $28.3 \pm 3.3$ \\
Sem. dupes {\footnotesize ($n$-branches)} & $87.5 \pm 2.1$ & $87.9 \pm 2.1$ & $29.8 \pm 3.3$ \\
\midrule
$\cos$ similarity {\footnotesize (SFT)} & -- & $68.6 \pm 2.1$ & $25.1 \pm 3.1$ \\
$\cos$ similarity {\footnotesize (RL)} & -- & $65.3 \pm 2.1$ & $26.7 \pm 3.2$ \\
\bottomrule
\end{tabularx}
\medskip
\vspace{0.25em}

\textbf{ZebraLogic}\par\smallskip
\begin{tabularx}{\linewidth}{p{1.25in}YYY}
\toprule
Degree of similarity & Seen accuracy & Unseen accuracy & Acc. on ARC Challenge \\
\midrule
No finetuning & -- & $36.9 \pm 1.5$ & $50.1 \pm 1.5$ \\
\midrule
\textbf{Exact dupes} & $\mathbf{48.4} \pm 1.6$ & $\mathbf{43.4} \pm 1.6$ & $49.5 \pm 1.5$ \\
Sem. dupes {\footnotesize (P)} & $39.2 \pm 1.5$ & $36.2 \pm 1.5$ & $49.3 \pm 1.5$ \\
Sem. dupes {\footnotesize (SH,SU)} & $36.0 \pm 1.5$ & $36.8 \pm 1.5$ & $50.7 \pm 1.5$ \\
Sem. dupes {\footnotesize (SH,P)} & $38.0 \pm 1.5$ & $36.0 \pm 1.5$ & $49.4 \pm 1.5$ \\
Sem. dupes {\footnotesize (SH,SU,P)} & $28.0 \pm 1.4$ & $28.4 \pm 1.4$ & $50.4 \pm 1.5$ \\
\midrule
$\cos$ similarity {\footnotesize (SFT)} & -- & $22.9 \pm 1.3$ & -- \\
\bottomrule
\end{tabularx}
\medskip
\vspace{0.25em}

\textbf{MBPP}\par\smallskip
\begin{tabularx}{\linewidth}{p{1.25in}YYY}
\toprule
Degree of similarity & Seen accuracy & Unseen accuracy & Acc. on HumanEval \\
\midrule
No finetuning & -- & $46.4 \pm 3.4$ & $55.3 \pm 3.9$ \\
\midrule
\textbf{Exact dupes} & $\mathbf{63.0} \pm 3.3$ & $\mathbf{48.8} \pm 3.5$ & $49.2 \pm 3.9$ \\
Sem. dupes {\footnotesize (Python)} & $55.1 \pm 3.4$ & $53.6 \pm 3.4$ & $\mathbf{67.0} \pm 3.7$ \\
\midrule
$\cos$ similarity {\footnotesize (SFT)} & -- & $48.8 \pm 3.5$ & $53.1 \pm 3.9$ \\
\bottomrule
\end{tabularx}
\label{tab:big_deltas}
\end{table}

\begin{figure}[ht]
  \centering
  \begin{subfigure}[b]{0.32\columnwidth}
    \includegraphics[width=\linewidth]{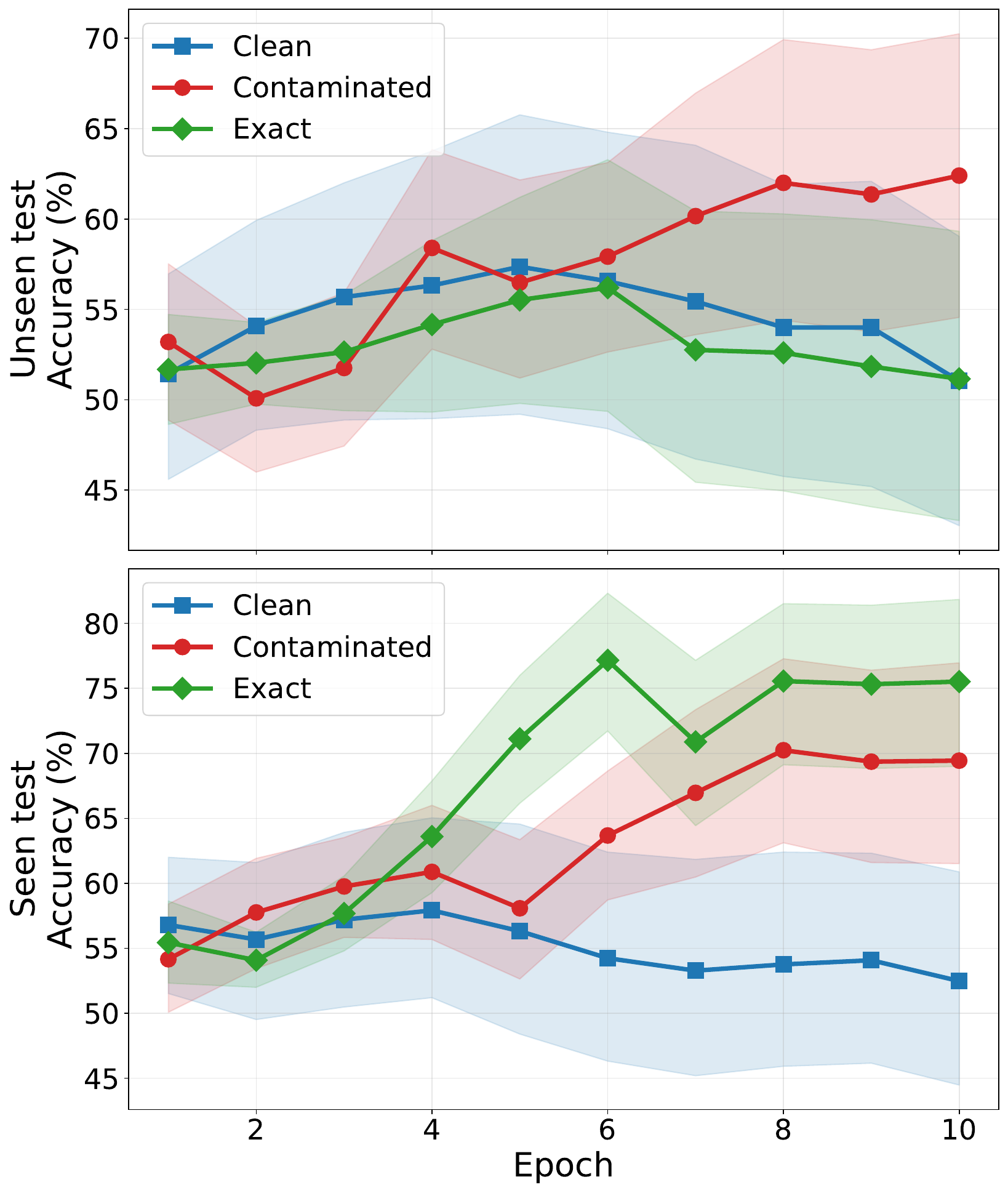}
    \caption{Olmo 3}
    \label{fig:olmo3-accuracy-curves}
  \end{subfigure}
  \hfill
  \begin{subfigure}[b]{0.32\columnwidth}
    \includegraphics[width=\linewidth]{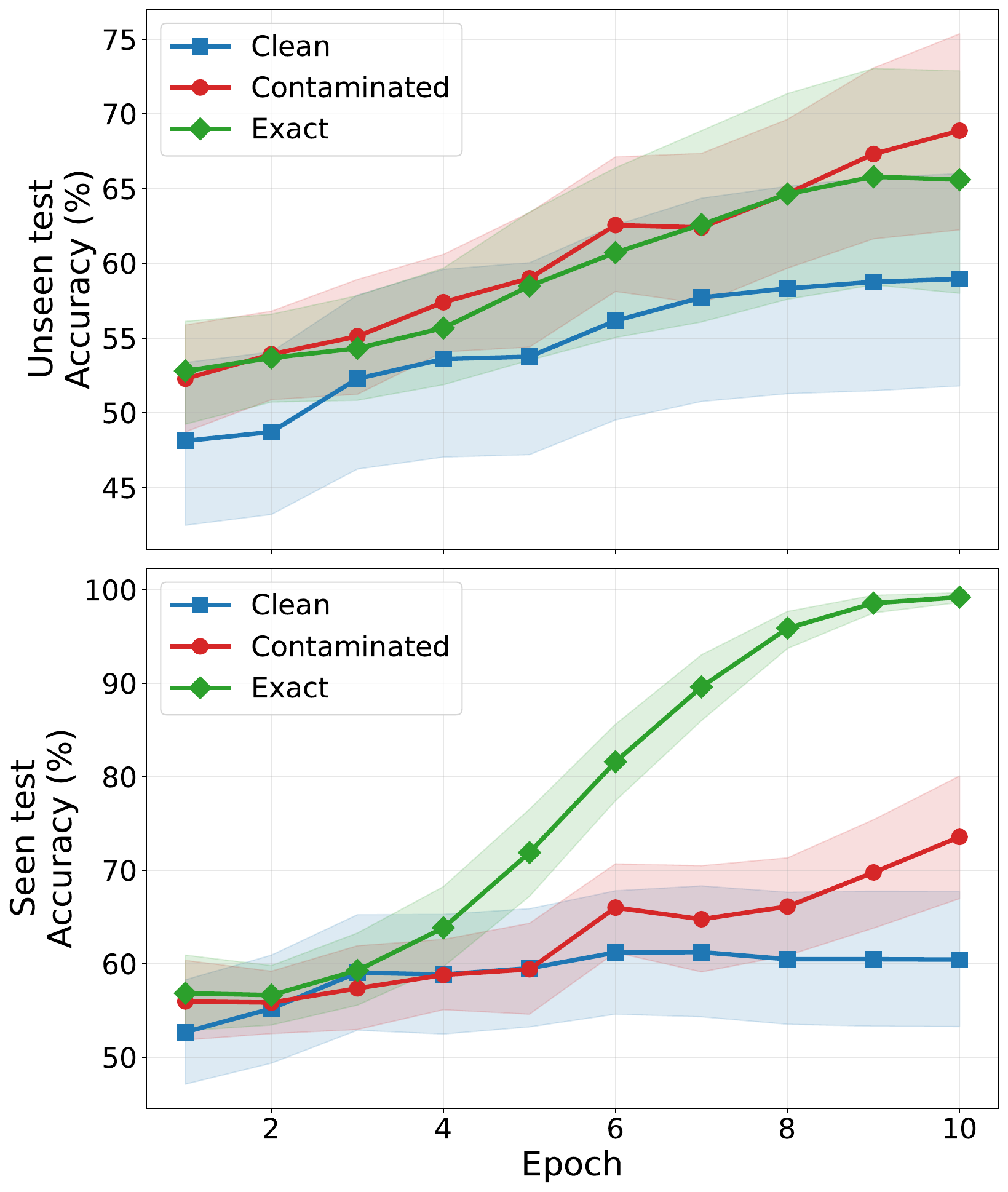}
    \caption{Qwen3}
    \label{fig:qwen3-accuracy-curves}
  \end{subfigure}
  \hfill
  \begin{subfigure}[b]{0.32\columnwidth}
    \includegraphics[width=\linewidth]{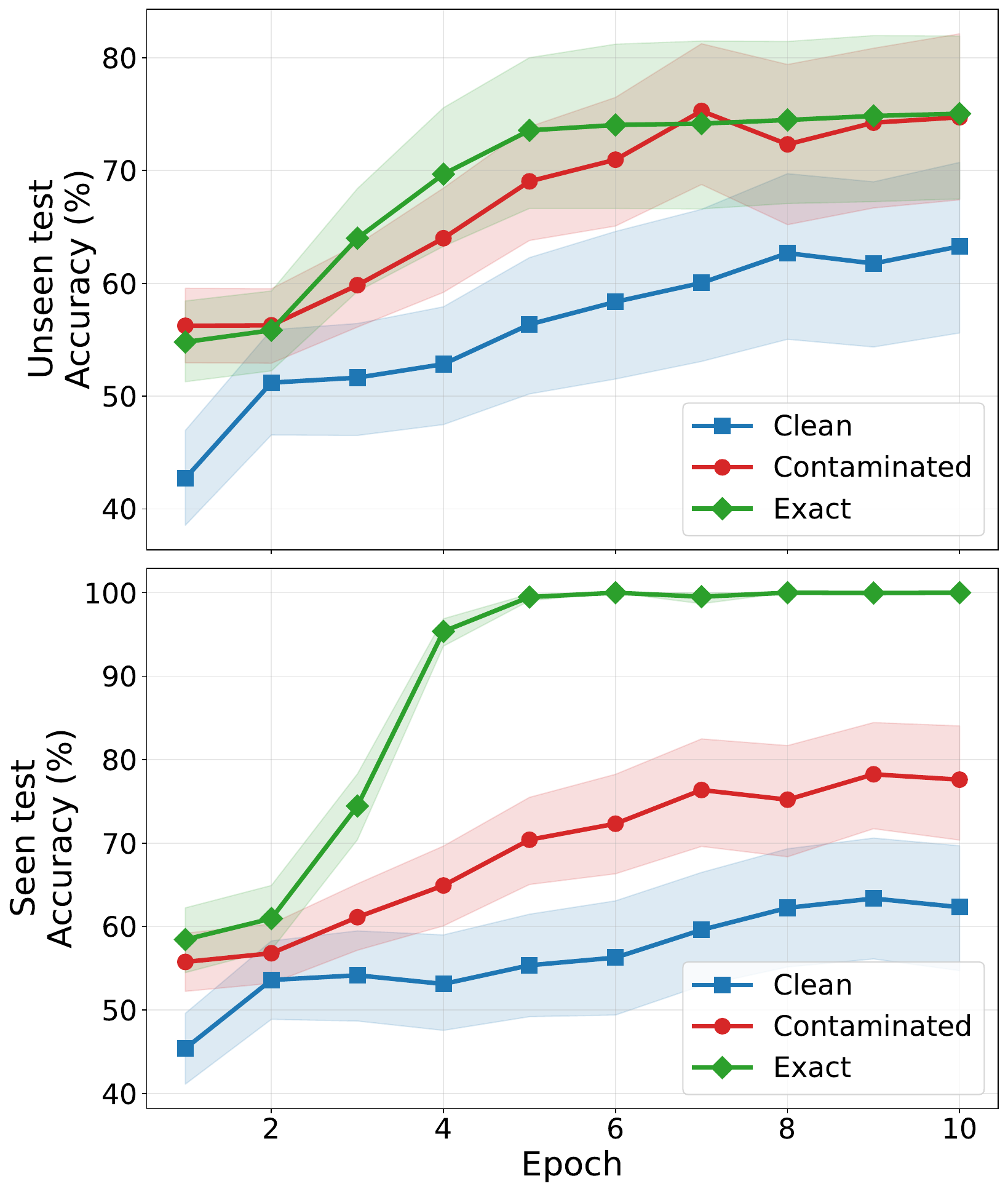}
    \caption{Qwen3.5}
    \label{fig:qwen35-accuracy-curves}
  \end{subfigure}
  \caption{\textbf{Ecologically valid finetuning on MuSR.} Accuracy on unseen items (top row) and seen items (bottom row) across finetuning epochs, under three conditions: no duplicates (clean, blue squares), exact duplicates (exact, green diamonds) and semantic duplicates (contaminated, red circles).
  For all three models, semantic duplicates significantly raise accuracy on unseen items relative to the clean run at epoch 10 ($p < 0.05$); for the Qwen models the difference is already significant at epoch 5.
  Table~\ref{tab:contamination_summary} gives means and $p$-values at epochs 5 and 10. Each run took about 5--8 hours on one GH200.}
  \label{fig:accuracy-curves}
\end{figure}


\textbf{Overall, we observe shallow generalization.}

When duplicates raise accuracy on seen items, they mostly raise it on unseen items too: the gap between the halves is at most 3pp, except for exact duplicates of MBPP (15pp) and ZebraLogic (5pp).
This suggests within-benchmark generalization, as predicted. On MuSR and ZebraLogic, the control benchmarks stay flat, thus completing the pattern we call shallow generalization (subject to the caveats in Section~\ref{sec:finetune-dupes}). On MBPP, semantic duplicates also improve the HumanEval control. Finetuning on high-cosine-similarity non-duplicates does not improve benchmark performance.

\subsection{Ecologically Valid Finetuning}\label{sec:ecological}

\textbf{Realistic Contamination Dosage.}\, We simulate SFT-stage exposure using our observed contamination rates. For MBPP, 40\% of the top 0.1\% cosine-similarity matches are true semantic duplicates. This suggests that about 4 in 10,000 training examples are semantic duplicates of any given MBPP test item; given our sampling density and the judge's false negatives, this is conservative.
We apply this per-item rate to MuSR (which has no natural semantic duplicates in the Olmo 3 corpus).

\textbf{Finetuning contaminated and clean models.}\,
We finetune Olmo-3-7B-base, Qwen3-8B-base \citep{yang2025qwen3} and Qwen3.5-9B-base on Olmo 3 SFT data mixed with synthetic MuSR semantic duplicates. The Olmo 3 post-training data contains no natural semantic duplicates of MuSR (Section~\ref{sec:find-sem-dupes}), so the only MuSR duplicates in the finetuning set are the ones we add. We cannot inspect the Qwen models' pretraining data, but any prior exposure affects their clean and contaminated runs alike.
We split MuSR into two halves of 125 items, \textit{seen} and \textit{unseen}, and generate four semantic duplicates of each seen item (two `$1$-branch' and two `$n$-branch'), 500 in total.
We run three finetuning conditions: 1) \textit{clean}: 10,000 decontaminated SFT examples; 2) \textit{contaminated}: the same set with 500 of the clean examples (5\%) replaced by the 500 semantic duplicates; and 3) \textit{exact}: the same set with 500 clean examples replaced by 500 exact duplicates (four copies of each seen item). Each seen item therefore has 4 duplicates per 10,000 training examples, matching our estimated natural rate; the aggregate rate of 5\% arises because all 125 seen items are duplicated at once.
We evaluate every model at each epoch, sampling 20 answers per question (Appendix~\ref{app:ecological-results}). Figure~\ref{fig:accuracy-curves} shows the accuracy curves and Table~\ref{tab:contamination_summary} reports means and $p$-values.

\textbf{Results on Olmo-3-7B.}\,
At epoch 10, the contaminated model scores 16.9pp higher than the clean model on seen items ($p = 0.001$) and 11.4pp higher on unseen items ($p = 0.037$). The effect appears late: at epoch 5 the two runs are indistinguishable on unseen items ($p = 0.85$), and part of the epoch-10 gap comes from the clean run's unseen accuracy falling from 57.4\% to 51.0\%. On unseen items, exact duplicates give essentially no gain over the clean run (51.2\% vs.\ 51.0\%), but the difference between semantic and exact contamination is not significant ($p = 0.084$).
Accuracy on the TrueDetective control stays flat across conditions, though Olmo 3 only scores at near chance on it.

\textbf{Results on Qwen3-8B and Qwen3.5-9B.}\,
On Qwen3-8B-base, at epoch 10 the contaminated model scores 13.2pp higher than the clean model on seen items ($p < 0.001$) and 9.9pp higher on unseen items ($p < 0.001$). On unseen items, semantic and exact contamination are statistically indistinguishable (68.9\% vs.\ 65.6\%, $p = 0.334$). Unlike with Olmo 3, the clean Qwen3 run also gains from finetuning relative to baseline (+5.0pp seen, +7.8pp unseen).
The contaminated Qwen3 run also improves modestly on the TrueDetective control.
On Qwen3.5-9B-base, the effect is largest early on: at epoch 5 the contaminated run already leads by 15.0pp on seen and 12.6pp on unseen items (both $p < 0.001$). At epoch 10 it leads by 15.3pp on seen items ($p < 0.001$) and 11.4pp on unseen items ($p = 0.009$). Again, semantic and exact contamination are statistically indistinguishable on unseen items (74.7\% vs.\ 75.0\%, $p = 0.754$).
As with Qwen3, the clean run also shows gains relative to its baseline (+6.8pp seen, +8.2pp unseen), concentrated on the random unseen subset (Appendix \ref{app:eco-fine-results}).

\textbf{Natural contamination inflates held-out scores.}
In all three models, semantic duplicates at our estimated natural rate significantly raise accuracy on held-out MuSR items relative to clean finetuning, i.e.\ semantic duplicates produce within-benchmark generalization.

\section{Conclusion}

We found widespread soft contamination of Olmo 3 training data with semantic duplicates, despite strong efforts from the Olmo team to filter out duplicates (Section \ref{sec:results-inthewild}). We also found that finetuning models on an ecologically valid amount of semantic duplicates leads to a statistically significant increase in accuracy on held-out items of the same benchmark (Section~\ref{sec:ecological}). In controlled experiments, semantic duplicates raise MuSR accuracy by up to 22pp while the control benchmark stays flat, the pattern we call shallow generalization; on MBPP, the gains also extend to HumanEval (Section~\ref{sec:finetuning-sem-dupes}). Scores on these benchmarks are therefore confounded by soft contamination.

\subsection{Recommendations}

\textit{At data collection time.} The community could benefit from a small, openly-released model that scores whether two items are semantic duplicates, for use as a corpus filter in place of thresholds on embedding similarity. The classifiers we finetuned on our LLM-judge labels (Appendix~\ref{app:annot}) are a first step. Authors should also report how many duplicates such filters find.

\textit{At evaluation time.} Report a control benchmark from the same domain alongside each target benchmark, chosen so that the model scores well above chance on it. A gain on the target without a gain on the control suggests shallow generalization. 

\textit{At benchmark creation time.} As training corpora grow, static benchmarks become more likely to be duplicated. Benchmarks that are continuously refreshed or that sample items stochastically from a growing pool are harder to duplicate.

\subsection{Limitations}
\label{sec:limits}

We likely underestimate the prevalence of semantic duplicates in frontier training corpora: our limited budget did not allow us to embed the full pretraining corpus in our detection methods. 

Another downward bias is the relative absence of rephrasings in Dolma: synthetic data pipelines now often involve intentionally creating semantic duplicates at scale \citep{wei-zou-2019-eda,wang2023selfinstructaligninglanguagemodels,maini2024rephrasingwebrecipecompute}, and so our estimates of their prevalence are likely lower than the true rate in closed corpora using our methods. 

Similarly, we do not cover SOTA synthetic data provided via RL environments. Our corpus search is limited to models with an open training corpus, a small and potentially unrepresentative set of systems, and for the Qwen models in our finetuning experiments we cannot inspect prior exposure. 

Our control benchmarks are imperfect (Section~\ref{sec:finetune-dupes}), so the scores of our control benchmark being flat is only weak evidence against transfer. Our controlled experiments use LoRA, and absolute effects may differ under full finetuning \citep{biderman2024lora}.

Contamination dynamics in our SFT-stage simulation may also differ from those in the much larger pretraining runs at frontier labs. Frontier training corpora are much larger than Dolma; some Llama 4 runs used 30T tokens \citep{meta2025llama4}. These corpora probably contain more semantic duplicates, and their rate of semantic duplication may differ from Dolma's.

A fundamental objection to our project is that out-of-distribution (OOD) generalization is no longer the (only) goal of AI development: an alternative is to instead bring all common tasks in-distribution (\citeauthor{chollet2024llmreasoning} \citeyear{chollet2024llmreasoning}, \citeauthor{patel2025sutton} \citeyear{patel2025sutton}, \citeauthor{leech2024questionablepracticesmachinelearning} \citeyear{leech2024questionablepracticesmachinelearning} \S 5.3.2). 
Perhaps the real-world utility of LLMs shows that our concerns about OOD generalization are not practically important, since they work even if their generalization were largely shallow. 

This is valid, but 1) then the deviation from the assumptions of empirical risk minimization should be explicitly noted, and 2) even perfect hidden interpolation may not be practically equivalent to OOD generalization \citep{hasson2020direct,balestriero2021learning}.
\clearpage
\section*{Contributions}
Spiesberger: Experimental design and execution of corpus search, model diffusion, embeddings, and ecological finetuning experiments; managed compute infrastructure and code repository, statistical arguments and analysis. Assisted with writing and editing. 

Vazquez: experimental design for generation and validation of semantic duplicates; comparison of duplicate detection methods; designed and executed annotation experiments; data spot-checking; wrote parts of methodology, results, and appendices; designed figures.

Pochinkov: generated MuSR teacher examples; finetuned and evaluated Olmo 3 models on MuSR and ZebraLogic; data sanity checks.

Gavenčiak: generated synthetic semantic duplicates; contributed to methodology, annotation pipeline design, finetuning experiments, and data analysis; provided feedback on the manuscript.

Grietzer: wrote the introduction and related work sections; edited the manuscript.

Leech: original research question; experiment design; wrote abstract, limitations, future work, and impact statement; funding.

Schoots: research and project management; led experiment design; coordinated writing efforts and wrote several parts of the manuscript.

\section*{Acknowledgments}
We thank the Olmo team for making this work possible and for responding with grace to criticism. We thank Noah Perez for his writing help.
We thank David Latshaw II, Max Shen, and Mary Putt for reading the paper and providing comments. We also thank Owain Evans, John Burden, Tom Davidson, Yuxi Liu, Teortaxes, and Martin Vlach for comments on an earlier draft.
Finally, we thank Lennis Schoots uit de Bos and Jasper for emotional support.

\section*{Impact Statement}

We aim to advance understanding of what LLM benchmark scores say about general capabilities on new inputs. Our findings have implications for how the AI research community, policymakers, and the public interpret reported progress on reasoning benchmarks. 

More-accurately measuring AI capabilities helps to calibrate decisions about AI deployment, regulation, and research. If benchmark gains partly reflect interpolation from a growing corpus rather than more general capability improvements, then recognizing this could help prevent overconfidence in model generalization to novel tasks.

We do not believe this work poses significant risks. While our methods could inform more sophisticated benchmark gaming, the contamination we study is likely accidental rather than adversarial, and our detection methods are likely more useful for auditing than for evasion. 

Our finding of contamination in Olmo3 is only possible because of the unusual level of transparency of its model development process. It would be unfair for readers to thereby assume that the degree of contamination in Olmo is unusual, and worse, a perverse incentive against transparency. 

Our work could easily be misread as `debunking' LLM capabilities and so spur complacency about near-term AI impacts. Our results suggest that benchmark gains are confounded (and so partially shallow), not that they are illusory.

\bibliography{references}

@inproceedings{DBLP:conf/iclr/SpragueYBCD24,
  author       = {Zayne Sprague and
                  Xi Ye and
                  Kaj Bostrom and
                  Swarat Chaudhuri and
                  Greg Durrett},
  title        = {MuSR: Testing the Limits of Chain-of-thought with Multistep Soft Reasoning},
  booktitle    = {The Twelfth International Conference on Learning Representations,
                  {ICLR} 2024, Vienna, Austria, May 7-11, 2024},
  publisher    = {OpenReview.net},
  year         = {2024},
  url          = {https://openreview.net/forum?id=jenyYQzue1},
  bibsource    = {dblp computer science bibliography, https://dblp.org}
}

@misc{meta2025llama4,
  author       = {{Meta AI}},
  title        = {The {Llama} 4 herd: The beginning of a new era of natively multimodal {AI} innovation},
  year         = {2025},
  month        = apr,
  day          = {5},
  howpublished = {\url{https://ai.meta.com/blog/llama-4-multimodal-intelligence/}},
  note         = {Accessed: 2026-01-26}
}

@misc{anthropic2025opus45systemcard,
  title        = {System Card: Claude Opus 4.5},
  author       = {{Anthropic}},
  year         = {2025},
  month        = nov,
  howpublished = {\url{https://www.anthropic.com/claude-opus-4-5-system-card}},
  note         = {Accessed 2026-01-26}
}

@inproceedings{brown2020language,
  title     = {Language Models are Few-Shot Learners},
  author    = {Brown, Tom B. and Mann, Benjamin and Ryder, Nick and Subbiah, Melanie and Kaplan, Jared and Dhariwal, Prafulla and others},
  booktitle = {Advances in Neural Information Processing Systems},
  volume    = {33},
  year      = {2020}
}

@misc{openai2024gpt4technicalreport,
      title={GPT-4 Technical Report}, 
      author={OpenAI and Josh Achiam and Steven Adler and Sandhini Agarwal and Lama Ahmad and Ilge Akkaya and Florencia Leoni Aleman and Diogo Almeida and Janko Altenschmidt and Sam Altman and Shyamal Anadkat and Red Avila and Igor Babuschkin and Suchir Balaji and Valerie Balcom and Paul Baltescu and Haiming Bao and Mohammad Bavarian and Jeff Belgum and Irwan Bello and Jake Berdine and Gabriel Bernadett-Shapiro and Christopher Berner and Lenny Bogdonoff and Oleg Boiko and Madelaine Boyd and Anna-Luisa Brakman and Greg Brockman and Tim Brooks and Miles Brundage and Kevin Button and Trevor Cai and Rosie Campbell and Andrew Cann and Brittany Carey and Chelsea Carlson and Rory Carmichael and Brooke Chan and Che Chang and Fotis Chantzis and Derek Chen and Sully Chen and Ruby Chen and Jason Chen and Mark Chen and Ben Chess and Chester Cho and Casey Chu and Hyung Won Chung and Dave Cummings and Jeremiah Currier and Yunxing Dai and Cory Decareaux and Thomas Degry and Noah Deutsch and Damien Deville and Arka Dhar and David Dohan and Steve Dowling and Sheila Dunning and Adrien Ecoffet and Atty Eleti and Tyna Eloundou and David Farhi and Liam Fedus and Niko Felix and Simón Posada Fishman and Juston Forte and Isabella Fulford and Leo Gao and Elie Georges and Christian Gibson and Vik Goel and Tarun Gogineni and Gabriel Goh and Rapha Gontijo-Lopes and Jonathan Gordon and Morgan Grafstein and Scott Gray and Ryan Greene and Joshua Gross and Shixiang Shane Gu and Yufei Guo and Chris Hallacy and Jesse Han and Jeff Harris and Yuchen He and Mike Heaton and Johannes Heidecke and Chris Hesse and Alan Hickey and Wade Hickey and Peter Hoeschele and Brandon Houghton and Kenny Hsu and Shengli Hu and Xin Hu and Joost Huizinga and Shantanu Jain and Shawn Jain and Joanne Jang and Angela Jiang and Roger Jiang and Haozhun Jin and Denny Jin and Shino Jomoto and Billie Jonn and Heewoo Jun and Tomer Kaftan and Łukasz Kaiser and Ali Kamali and Ingmar Kanitscheider and Nitish Shirish Keskar and Tabarak Khan and Logan Kilpatrick and Jong Wook Kim and Christina Kim and Yongjik Kim and Jan Hendrik Kirchner and Jamie Kiros and Matt Knight and Daniel Kokotajlo and Łukasz Kondraciuk and Andrew Kondrich and Aris Konstantinidis and Kyle Kosic and Gretchen Krueger and Vishal Kuo and Michael Lampe and Ikai Lan and Teddy Lee and Jan Leike and Jade Leung and Daniel Levy and Chak Ming Li and Rachel Lim and Molly Lin and Stephanie Lin and Mateusz Litwin and Theresa Lopez and Ryan Lowe and Patricia Lue and Anna Makanju and Kim Malfacini and Sam Manning and Todor Markov and Yaniv Markovski and Bianca Martin and Katie Mayer and Andrew Mayne and Bob McGrew and Scott Mayer McKinney and Christine McLeavey and Paul McMillan and Jake McNeil and David Medina and Aalok Mehta and Jacob Menick and Luke Metz and Andrey Mishchenko and Pamela Mishkin and Vinnie Monaco and Evan Morikawa and Daniel Mossing and Tong Mu and Mira Murati and Oleg Murk and David Mély and Ashvin Nair and Reiichiro Nakano and Rajeev Nayak and Arvind Neelakantan and Richard Ngo and Hyeonwoo Noh and Long Ouyang and Cullen O'Keefe and Jakub Pachocki and Alex Paino and Joe Palermo and Ashley Pantuliano and Giambattista Parascandolo and Joel Parish and Emy Parparita and Alex Passos and Mikhail Pavlov and Andrew Peng and Adam Perelman and Filipe de Avila Belbute Peres and Michael Petrov and Henrique Ponde de Oliveira Pinto and Michael and Pokorny and Michelle Pokrass and Vitchyr H. Pong and Tolly Powell and Alethea Power and Boris Power and Elizabeth Proehl and Raul Puri and Alec Radford and Jack Rae and Aditya Ramesh and Cameron Raymond and Francis Real and Kendra Rimbach and Carl Ross and Bob Rotsted and Henri Roussez and Nick Ryder and Mario Saltarelli and Ted Sanders and Shibani Santurkar and Girish Sastry and Heather Schmidt and David Schnurr and John Schulman and Daniel Selsam and Kyla Sheppard and Toki Sherbakov and Jessica Shieh and Sarah Shoker and Pranav Shyam and Szymon Sidor and Eric Sigler and Maddie Simens and Jordan Sitkin and Katarina Slama and Ian Sohl and Benjamin Sokolowsky and Yang Song and Natalie Staudacher and Felipe Petroski Such and Natalie Summers and Ilya Sutskever and Jie Tang and Nikolas Tezak and Madeleine B. Thompson and Phil Tillet and Amin Tootoonchian and Elizabeth Tseng and Preston Tuggle and Nick Turley and Jerry Tworek and Juan Felipe Cerón Uribe and Andrea Vallone and Arun Vijayvergiya and Chelsea Voss and Carroll Wainwright and Justin Jay Wang and Alvin Wang and Ben Wang and Jonathan Ward and Jason Wei and CJ Weinmann and Akila Welihinda and Peter Welinder and Jiayi Weng and Lilian Weng and Matt Wiethoff and Dave Willner and Clemens Winter and Samuel Wolrich and Hannah Wong and Lauren Workman and Sherwin Wu and Jeff Wu and Michael Wu and Kai Xiao and Tao Xu and Sarah Yoo and Kevin Yu and Qiming Yuan and Wojciech Zaremba and Rowan Zellers and Chong Zhang and Marvin Zhang and Shengjia Zhao and Tianhao Zheng and Juntang Zhuang and William Zhuk and Barret Zoph},
      year={2024},
      eprint={2303.08774},
      archivePrefix={arXiv},
      primaryClass={cs.CL},
      url={https://arxiv.org/abs/2303.08774}, 
}

@misc{patel2025sutton,
  author       = {Patel, Dwarkesh and Sutton, Richard},
  title        = {Richard Sutton -- Father of {RL} thinks {LLMs} are a dead end},
  year         = {2025},
  month        = sep,
  day          = {26},
  howpublished = {Dwarkesh Podcast},
  url          = {https://www.dwarkesh.com/p/richard-sutton},
  note         = {Podcast interview}
}

@inproceedings{dong2024generalization,
  title={Generalization or memorization: Data contamination and trustworthy evaluation for large language models},
  author={Dong, Yihong and Jiang, Xue and Liu, Huanyu and others},
  booktitle={Findings of the Association for Computational Linguistics: ACL 2024},
  pages={12039--12050},
  year={2024}
}

@article{kambhampati2024llms,
  author  = {Subbarao Kambhampati},
  title   = {Can Large Language Models Reason and Plan?},
  journal = {Annals of the New York Academy of Sciences},
  volume  = {1534},
  number  = {1},
  pages   = {15--18},
  year    = {2024}
}

@article{yang2023rethinking,
  title={Rethinking benchmark and contamination for language models with rephrased samples},
  author={Yang, Shuo and Chiang, Wei-Lin and Zheng, Lianmin and others},
  journal={arXiv preprint arXiv:2311.04850},
  year={2023}
}

@misc{meursault2025approximate,
  author = {Vitaly Meursault},
  title  = {{LLMs} are Approximate Retrievers},
  year   = {2025},
  month  = {June},
  howpublished = {\url{https://pleasebringstrangethings.substack.com/p/llms-are-approximate-retrievers}},
  note   = {Substack post. Accessed: 2026-05-06}
}

@misc{chollet2024llmreasoning,
  author       = {Chollet, Fran\c{c}ois},
  title        = {The question of whether {LLMs} can reason is, in many ways, the wrong question [{Tweet}]},
  year         = {2024},
  month        = jul,
  day          = {26},
  howpublished = {X (formerly Twitter)},
  url          = {https://x.com/fchollet/status/1816954290227089656},
  note         = {Accessed: 2026-01-26}
}

@misc{leech2024questionablepracticesmachinelearning,
      title={Questionable practices in machine learning}, 
      author={Gavin Leech and Juan J. Vazquez and Niclas Kupper and Misha Yagudin and Laurence Aitchison},
      year={2024},
      eprint={2407.12220},
      archivePrefix={arXiv},
      primaryClass={cs.LG},
      url={https://arxiv.org/abs/2407.12220}, 
}

@misc{EpochAIModels2025,
  title = {Data on AI Models},
  author = {{Epoch AI}},
  year = {2025},
  month = {07},
  url = {https://epoch.ai/data/ai-models},
  note = {Accessed: 2026-01-26}
}

@misc{maslej2025artificialintelligenceindexreport,
      title={Artificial Intelligence Index Report 2025}, 
      author={Nestor Maslej and Loredana Fattorini and Raymond Perrault and Yolanda Gil and Vanessa Parli and Njenga Kariuki and Emily Capstick and Anka Reuel and Erik Brynjolfsson and John Etchemendy and Katrina Ligett and Terah Lyons and James Manyika and Juan Carlos Niebles and Yoav Shoham and Russell Wald and Toby Walsh and Armin Hamrah and Lapo Santarlasci and Julia Betts Lotufo and Alexandra Rome and Andrew Shi and Sukrut Oak},
      year={2025},
      eprint={2504.07139},
      archivePrefix={arXiv},
      primaryClass={cs.AI},
      url={https://arxiv.org/abs/2504.07139}, 
}

@misc{epoch2025aicapabilitiesprogresshasspedup,
    title={AI capabilities progress has sped up},
    author={Yafah Edelman and Jaeho Lee},
    year={2025},
    url={https://epoch.ai/data-insights/ai-capabilities-progress-has-sped-up},
    note={Accessed: 2026-01-26}
  }

@misc{wang2023selfinstructaligninglanguagemodels,
      title={Self-Instruct: Aligning Language Models with Self-Generated Instructions}, 
      author={Yizhong Wang and Yeganeh Kordi and Swaroop Mishra and Alisa Liu and Noah A. Smith and Daniel Khashabi and Hannaneh Hajishirzi},
      year={2022},
      eprint={2212.10560},
      archivePrefix={arXiv},
      primaryClass={cs.CL},
      url={https://arxiv.org/abs/2212.10560}, 
}

@inproceedings{wei-zou-2019-eda,
    title = "{EDA}: Easy Data Augmentation Techniques for Boosting Performance on Text Classification Tasks",
    author = "Wei, Jason  and
      Zou, Kai",
    editor = "Inui, Kentaro  and
      Jiang, Jing  and
      Ng, Vincent  and
      Wan, Xiaojun",
    booktitle = "Proceedings of the 2019 Conference on Empirical Methods in Natural Language Processing and the 9th International Joint Conference on Natural Language Processing (EMNLP-IJCNLP)",
    month = nov,
    year = "2019",
    address = "Hong Kong, China",
    publisher = "Association for Computational Linguistics",
    url = "https://aclanthology.org/D19-1670/",
    doi = "10.18653/v1/D19-1670",
    pages = "6382--6388"
}

@misc{maini2024rephrasingwebrecipecompute,
      title={Rephrasing the Web: A Recipe for Compute and Data-Efficient Language Modeling}, 
      author={Pratyush Maini and Skyler Seto and He Bai and David Grangier and Yizhe Zhang and Navdeep Jaitly},
      year={2024},
      eprint={2401.16380},
      archivePrefix={arXiv},
      primaryClass={cs.CL},
      url={https://arxiv.org/abs/2401.16380}, 
}

@inproceedings{DBLP:conf/icml/Lin00SPC025,
  author       = {Bill Yuchen Lin and
                  Ronan Le Bras and
                  Kyle Richardson and
                  Ashish Sabharwal and
                  Radha Poovendran and
                  Peter Clark and
                  Yejin Choi},
  title        = {ZebraLogic: On the Scaling Limits of LLMs for Logical Reasoning},
  booktitle    = {Forty-second International Conference on Machine Learning, {ICML}
                  2025, Vancouver, BC, Canada, July 13-19, 2025},
  publisher    = {OpenReview.net},
  year         = {2025},
  url          = {https://openreview.net/forum?id=sTAJ9QyA6l},
  bibsource    = {dblp computer science bibliography, https://dblp.org}
}

@article{DBLP:journals/corr/abs-2108-07732,
  author       = {Jacob Austin and
                  Augustus Odena and
                  Maxwell I. Nye and
                  Maarten Bosma and
                  Henryk Michalewski and
                  David Dohan and
                  Ellen Jiang and
                  Carrie J. Cai and
                  Michael Terry and
                  Quoc V. Le and
                  Charles Sutton},
  title        = {Program Synthesis with Large Language Models},
  journal      = {CoRR},
  volume       = {abs/2108.07732},
  year         = {2021},
  url          = {https://arxiv.org/abs/2108.07732},
  eprinttype    = {arXiv},
  eprint       = {2108.07732},
  bibsource    = {dblp computer science bibliography, https://dblp.org}
}

@inproceedings{riddell-etal-2024-quantifying,
  title = {Quantifying Contamination in Evaluating Code Generation Capabilities of Language Models},
  author = {Riddell, Martin and Ni, Ansong and Cohan, Arman},
  editor = {Ku, Lun-Wei and Martins, Andre and Srikumar, Vivek},
  booktitle = {Proceedings of the 62nd Annual Meeting of the Association for Computational Linguistics (Volume 1: Long Papers)},
  month = aug,
  year = {2024},
  address = {Bangkok, Thailand},
  publisher = {Association for Computational Linguistics},
  url = {https://aclanthology.org/2024.acl-long.761/},
  doi = {10.18653/v1/2024.acl-long.761},
  pages = {14116--14137}
}

@article{vapnik1991principles,
  title={Principles of risk minimization for learning theory},
  author={Vapnik, Vladimir},
  journal={Advances in neural information processing systems},
  volume={4},
  year={1991}
}

@article{dominguez2024training,
  title={Training on the test task confounds evaluation and emergence},
  author={Dominguez-Olmedo, Ricardo and Dorner, Florian E and Hardt, Moritz},
  journal={arXiv preprint arXiv:2407.07890},
  year={2024}
}

@article{balestriero2021learning,
  title={Learning in high dimension always amounts to extrapolation},
  author={Balestriero, Randall and Pesenti, Jerome and LeCun, Yann},
  journal={arXiv preprint arXiv:2110.09485},
  year={2021}
}

@article{hasson2020direct,
  title={Direct fit to nature: an evolutionary perspective on biological and artificial neural networks},
  author={Hasson, Uri and Nastase, Samuel A and Goldstein, Ariel},
  journal={Neuron},
  volume={105},
  number={3},
  pages={416--434},
  year={2020},
  publisher={Elsevier}
}

@inproceedings{xu-etal-2025-ssa,
  title = {{SSA}: Semantic Contamination of {LLM}-Driven Fake News Detection},
  author = {Xu, Cheng and Yan, Nan and Guan, Shuhao and Mei, Yuke and Kechadi, Tahar},
  editor = {Christodoulopoulos, Christos and Chakraborty, Tanmoy and Rose, Carolyn and Peng, Violet},
  booktitle = {Proceedings of the 2025 Conference on Empirical Methods in Natural Language Processing},
  month = nov,
  year = {2025},
  address = {Suzhou, China},
  publisher = {Association for Computational Linguistics},
  url = {https://aclanthology.org/2025.emnlp-main.744/},
  doi = {10.18653/v1/2025.emnlp-main.744},
  pages = {14737--14751}
}

@InProceedings{pmlr-v267-kocyigit25a,
  title = {Overestimation in {LLM} Evaluation: A Controlled Large-Scale Study on Data Contamination’s Impact on Machine Translation},
  author = {Kocyigit, Muhammed Yusuf and Briakou, Eleftheria and Deutsch, Daniel and Luo, Jiaming and Cherry, Colin and Freitag, Markus},
  booktitle = {Proceedings of the 42nd International Conference on Machine Learning},
  pages = {31105--31132},
  year = {2025},
  volume = {267},
  series = {Proceedings of Machine Learning Research},
  publisher = {PMLR},
  url = {https://proceedings.mlr.press/v267/kocyigit25a.html}
}

@inproceedings{magar2022data,
  author    = {Inbal Magar and Roy Schwartz},
  title     = {Data Contamination: From Memorization to Exploitation},
  booktitle = {Proceedings of the 60th Annual Meeting of the Association for Computational Linguistics (Volume 2: Short Papers)},
  pages     = {157--165},
  year      = {2022},
  address   = {Dublin, Ireland},
  publisher = {Association for Computational Linguistics},
  doi       = {10.18653/v1/2022.acl-short.18}
}

@misc{kambhampati2023tweet,
  author       = {Subbarao Kambhampati},
  title        = {[Tweet on {LLMs} as approximate retrievers]},
  year         = {2023},
  month        = {December},
  howpublished = {X (formerly Twitter)},
  url          = {https://x.com/rao2z/status/1740692722099630237},
  note         = {Accessed: 2026-05-06}
}

@article{biderman2024lora,
  title={{LoRA} Learns Less and Forgets Less},
  author={Biderman, Dan and Portes, Jacob and Gonzalez Ortiz, Jose Javier and Paul, Mansheej and Greengard, Philip and Jennings, Connor and King, Daniel and Havens, Sam and Chiley, Vitaliy and Frankle, Jonathan and Blakeney, Cody and Cunningham, John Patrick},
  journal={Transactions on Machine Learning Research},
  year={2024},
  url={https://openreview.net/forum?id=aloEru2qCG}
}

@inproceedings{elazar2024whats,
  author    = {Yanai Elazar and Akshita Bhagia and Ian Helgi Magnusson and Abhilasha Ravichander and Dustin Schwenk and Alane Suhr and Evan Pete Walsh and Dirk Groeneveld and Luca Soldaini and Sameer Singh and Hannaneh Hajishirzi and Noah A. Smith and Jesse Dodge},
  title     = {What's In My Big Data?},
  booktitle = {The Twelfth International Conference on Learning Representations},
  year      = {2024},
  url       = {https://openreview.net/forum?id=RvfPnOkPV4}
}

@misc{jiang2024investigating,
  title = {Investigating Data Contamination for Pre-training Language Models},
  author = {Jiang, Minhao and Liu, Ken Ziyu and Zhong, Ming and Schaeffer, Rylan and Ouyang, Siru and Han, Jiawei and Koyejo, Sanmi},
  year = {2024},
  eprint = {2401.06059},
  archivePrefix = {arXiv},
  primaryClass = {cs.CL},
  doi = {10.48550/arXiv.2401.06059},
  url = {https://arxiv.org/abs/2401.06059}
}

@misc{zhou2025lessleak,
  title = {LessLeak-Bench: A First Investigation of Data Leakage in {LLM}s Across 83 Software Engineering Benchmarks},
  author = {Zhou, Xin and Weyssow, Martin and Widyasari, Ratnadira and Zhang, Ting and He, Junda and Lyu, Yunbo and Chang, Jianming and Zhang, Beiqi and Huang, Dan and Lo, David},
  year = {2025},
  eprint = {2502.06215},
  archivePrefix = {arXiv},
  primaryClass = {cs.SE},
  doi = {10.48550/arXiv.2502.06215},
  url = {https://arxiv.org/abs/2502.06215}
}

@misc{shilov2025mosaic,
  title = {The Mosaic Memory of Large Language Models},
  author = {Shilov, Igor and Meeus, Matthieu and de Montjoye, Yves-Alexandre},
  year = {2025},
  note = {arXiv:2405.15523v2},
  eprint = {2405.15523},
  archivePrefix = {arXiv},
  primaryClass = {cs.CL},
  doi = {10.48550/arXiv.2405.15523},
  url = {https://arxiv.org/abs/2405.15523}
}

@misc{shi2023detecting,
  title = {Detecting Pretraining Data from Large Language Models},
  author = {Shi, Weijia and Ajith, Anirudh and Xia, Mengzhou and Huang, Yangsibo and Liu, Daogao and Blevins, Terra and Chen, Danqi and Zettlemoyer, Luke},
  year = {2023},
  eprint = {2310.16789},
  archivePrefix = {arXiv},
  primaryClass = {cs.CL},
  url = {https://arxiv.org/abs/2310.16789}
}

@misc{chen2021evaluating,
  title = {Evaluating Large Language Models Trained on Code},
  author = {Chen, Mark and Tworek, Jerry and Jun, Heewoo and Yuan, Qiming and Pinto, Henrique Ponde de Oliveira and Kaplan, Jared and Edwards, Harri and Burda, Yuri and Joseph, Nicholas and Brockman, Greg and Ray, Alex and Puri, Raul and Krueger, Gretchen and Petrov, Michael and Khlaaf, Heidy and Sastry, Girish and Mishkin, Pamela and Chan, Brooke and Gray, Scott and Ryder, Nick and Pavlov, Mikhail and Power, Alethea and Kaiser, Lukasz and Bavarian, Mohammad and Winter, Clemens and Tillet, Philippe and Such, Felipe Petroski and Cummings, Dave and Plappert, Matthias and Chantzis, Fotios and Barnes, Elizabeth and Herbert-Voss, Ariel and Guss, William Hebgen and Nichol, Alex and Paino, Alex and Tezak, Nikolas and Tang, Jie and Babuschkin, Igor and Balaji, Suchir and Jain, Shantanu and Saunders, William and Hesse, Christopher and Carr, Andrew N. and Leike, Jan and Achiam, Josh and Misra, Vedant and Morikawa, Evan and Radford, Alec and Knight, Matthew and Brundage, Miles and Murati, Mira and Mayer, Katie and Welinder, Peter and McGrew, Bob and Amodei, Dario and McCandlish, Sam and Sutskever, Ilya and Zaremba, Wojciech},
  year = {2021},
  eprint = {2107.03374},
  archivePrefix = {arXiv},
  primaryClass = {cs.LG},
  url = {https://arxiv.org/abs/2107.03374}
}

@misc{gao2020pile,
  title = {The Pile: An 800GB Dataset of Diverse Text for Language Modeling},
  author = {Gao, Leo and Biderman, Stella and Black, Sid and Golding, Laurence and Hoppe, Travis and Foster, Charles and Phang, Jason and He, Horace and Thite, Anish and Nabeshima, Noa and Presser, Shawn and Leahy, Connor},
  year = {2020},
  eprint = {2101.00027},
  archivePrefix = {arXiv},
  primaryClass = {cs.CL},
  url = {https://arxiv.org/abs/2101.00027}
}

@misc{li2023starcoder,
  title = {{StarCoder}: may the source be with you!},
  author = {Li, Raymond and Ben Allal, Loubna and Zi, Yangtian and Muennighoff, Niklas and Kocetkov, Denis and Mou, Chenghao and Marone, Marc and Akiki, Christopher and Li, Jia and Chim, Jenny and Liu, Qian and Zheltonozhskii, Evgenii and Zhuo, Terry Yue and Wang, Thomas and Dehaene, Olivier and Davaadorj, Mishig and Lamy-Poirier, Joel and Monteiro, Jo{\~a}o and Shliazhko, Oleh and Gontier, Nicolas and Meade, Nicholas and Zebaze, Armel and Yee, Ming-Ho and Umapathi, Logesh Kumar and Zhu, Jian and Lipkin, Benjamin and Oblokulov, Muhtasham and Wang, Zhiruo and Murthy, Rudra and Stillerman, Jason and Patel, Siva Sankalp and Abulkhanov, Dmitry and Zocca, Marco and Dey, Manan and Zhang, Zhihan and Fahmy, Nour and Bhattacharyya, Urvashi and Yu, Wenhao and Singh, Swayam and Luccioni, Sasha and Villegas, Paulo and Kunakov, Maxim and Zhdanov, Fedor and Romero, Manuel and Lee, Tony and Timor, Nadav and Ding, Jennifer and Schlesinger, Claire and Schoelkopf, Hailey and Ebert, Jan and Dao, Tri and Mishra, Mayank and Gu, Alex and Robinson, Jennifer and Anderson, Carolyn Jane and Dolan-Gavitt, Brendan and Contractor, Danish and Reddy, Siva and Fried, Daniel and Bahdanau, Dzmitry and Jernite, Yacine and Ferrandis, Carlos Mu{\~n}oz and Hughes, Sean and Wolf, Thomas and Guha, Arjun and von Werra, Leandro and de Vries, Harm},
  year = {2023},
  eprint = {2305.06161},
  archivePrefix = {arXiv},
  primaryClass = {cs.CL},
  doi = {10.48550/arXiv.2305.06161},
  url = {https://arxiv.org/abs/2305.06161}
}

@misc{weber2024redpajama,
  title = {RedPajama: an Open Dataset for Training Large Language Models},
  author = {Weber, Maurice and Fu, Daniel and Anthony, Quentin and Oren, Yonatan and Adams, Shane and Alexandrov, Anton and Lyu, Xiaozhong and Nguyen, Huu and Yao, Xiaozhe and Adams, Virginia and Athiwaratkun, Ben and Chalamala, Rahul and Chen, Kezhen and Ryabinin, Max and Dao, Tri and Liang, Percy and R{\'e}, Christopher and Rish, Irina and Zhang, Ce},
  year = {2024},
  eprint = {2411.12372},
  archivePrefix = {arXiv},
  primaryClass = {cs.CL},
  doi = {10.48550/arXiv.2411.12372},
  url = {https://arxiv.org/abs/2411.12372}
}

@misc{olmo2025olmo3,
      title={Olmo 3}, 
      author={Team Olmo and : and Allyson Ettinger and Amanda Bertsch and Bailey Kuehl and David Graham and David Heineman and Dirk Groeneveld and Faeze Brahman and Finbarr Timbers and Hamish Ivison and Jacob Morrison and Jake Poznanski and Kyle Lo and Luca Soldaini and Matt Jordan and Mayee Chen and Michael Noukhovitch and Nathan Lambert and Pete Walsh and Pradeep Dasigi and Robert Berry and Saumya Malik and Saurabh Shah and Scott Geng and Shane Arora and Shashank Gupta and Taira Anderson and Teng Xiao and Tyler Murray and Tyler Romero and Victoria Graf and Akari Asai and Akshita Bhagia and Alexander Wettig and Alisa Liu and Aman Rangapur and Chloe Anastasiades and Costa Huang and Dustin Schwenk and Harsh Trivedi and Ian Magnusson and Jaron Lochner and Jiacheng Liu and Lester James V. Miranda and Maarten Sap and Malia Morgan and Michael Schmitz and Michal Guerquin and Michael Wilson and Regan Huff and Ronan Le Bras and Rui Xin and Rulin Shao and Sam Skjonsberg and Shannon Zejiang Shen and Shuyue Stella Li and Tucker Wilde and Valentina Pyatkin and Will Merrill and Yapei Chang and Yuling Gu and Zhiyuan Zeng and Ashish Sabharwal and Luke Zettlemoyer and Pang Wei Koh and Ali Farhadi and Noah A. Smith and Hannaneh Hajishirzi},
      year={2025},
      eprint={2512.13961},
      archivePrefix={arXiv},
      primaryClass={cs.CL},
      url={https://arxiv.org/abs/2512.13961}, 
}

@inproceedings{DBLP:conf/iclr/HuSWALWWC22,
  author       = {Edward J. Hu and
                  Yelong Shen and
                  Phillip Wallis and
                  Zeyuan Allen{-}Zhu and
                  Yuanzhi Li and
                  Shean Wang and
                  Lu Wang and
                  Weizhu Chen},
  title        = {LoRA: Low-Rank Adaptation of Large Language Models},
  booktitle    = {The Tenth International Conference on Learning Representations, {ICLR}
                  2022, Virtual Event, April 25-29, 2022},
  publisher    = {OpenReview.net},
  year         = {2022},
  url          = {https://openreview.net/forum?id=nZeVKeeFYf9},
  bibsource    = {dblp computer science bibliography, https://dblp.org}
}

@misc{penedo2025codeforces,
      title={CodeForces}, 
      author={Guilherme Penedo and Anton Lozhkov and Hynek Kydlíček and Loubna Ben Allal and Edward Beeching and Agustín Piqueres Lajarín and Quentin Gallouédec and Nathan Habib and Lewis Tunstall and Leandro von Werra},
      year={2025},
      publisher = {Hugging Face},
      journal = {Hugging Face repository},
      howpublished = {\url{https://huggingface.co/datasets/open-r1/codeforces}}
}

@inproceedings{del2023true,
  title={True detective: A deep abductive reasoning benchmark undoable for GPT-3 and challenging for GPT-4},
  author={Del, Maksym and Fishel, Mark},
  booktitle={Proceedings of the 12th Joint Conference on Lexical and Computational Semantics (* SEM 2023)},
  pages={314--322},
  year={2023}
}

@article{hendrycks2020measuring,
  title={Measuring massive multitask language understanding},
  author={Hendrycks, Dan and Burns, Collin and Basart, Steven and Zou, Andy and Mazeika, Mantas and Song, Dawn and Steinhardt, Jacob},
  journal={arXiv preprint arXiv:2009.03300},
  year={2020}
}

@article{cobbe2021training,
  title={Training verifiers to solve math word problems},
  author={Cobbe, Karl and Kosaraju, Vineet and Bavarian, Mohammad and Chen, Mark and Jun, Heewoo and Kaiser, Lukasz and Plappert, Matthias and Tworek, Jerry and Hilton, Jacob and Nakano, Reiichiro and others},
  journal={arXiv preprint arXiv:2110.14168},
  year={2021}
}

@article{clark2018think,
  title={Think you have solved question answering? try arc, the ai2 reasoning challenge},
  author={Clark, Peter and Cowhey, Isaac and Etzioni, Oren and Khot, Tushar and Sabharwal, Ashish and Schoenick, Carissa and Tafjord, Oyvind},
  journal={arXiv preprint arXiv:1803.05457},
  year={2018}
}

@inproceedings{muennighoff2023mteb,
  title={Mteb: Massive text embedding benchmark},
  author={Muennighoff, Niklas and Tazi, Nouamane and Magne, Lo{\"\i}c and Reimers, Nils},
  booktitle={Proceedings of the 17th Conference of the European Chapter of the Association for Computational Linguistics},
  pages={2014--2037},
  year={2023}
}

@article{babakhin2025llama,
  title={Llama-Embed-Nemotron-8B: A Universal Text Embedding Model for Multilingual and Cross-Lingual Tasks},
  author={Babakhin, Yauhen and Osmulski, Radek and Ak, Ronay and Moreira, Gabriel and Xu, Mengyao and Schifferer, Benedikt and Liu, Bo and Oldridge, Even},
  journal={arXiv preprint arXiv:2511.07025},
  year={2025}
}

@misc{google2025gemini3flash,
  title        = {Gemini 3 Flash Model Card},
  author       = {{Google DeepMind}},
  year         = {2025},
  month        = dec,
  howpublished = {Model Card},
  url          = {https://storage.googleapis.com/deepmind-media/Model-Cards/Gemini-3-Flash-Model-Card.pdf}
}

@misc{anthropic2025claudehaiku45,
  title        = {System Card: Claude Haiku 4.5},
  author       = {{Anthropic}},
  year         = {2025},
  month        = oct,
  howpublished = {Model Card},
  url          = {https://www-cdn.anthropic.com/7aad69bf12627d42234e01ee7c36305dc2f6a970.pdf}
}

@misc{anthropic2025claudesonnet45,
  title        = {System Card: Claude Sonnet 4.5},
  author       = {{Anthropic}},
  year         = {2025},
  month        = sep,
  howpublished = {Model Card},
  url          = {https://www-cdn.anthropic.com/963373e433e489a87a10c823c52a0a013e9172dd.pdf}
}

@article{wei2022chain,
  title={Chain-of-thought prompting elicits reasoning in large language models},
  author={Wei, Jason and Wang, Xuezhi and Schuurmans, Dale and Bosma, Maarten and Xia, Fei and Chi, Ed and Le, Quoc V and Zhou, Denny and others},
  journal={Advances in neural information processing systems},
  volume={35},
  pages={24824--24837},
  year={2022}
}

@article{clark2019boolq,
  title={Boolq: Exploring the surprising difficulty of natural yes/no questions},
  author={Clark, Christopher and Lee, Kenton and Chang, Ming-Wei and Kwiatkowski, Tom and Collins, Michael and Toutanova, Kristina},
  journal={arXiv preprint arXiv:1905.10044},
  year={2019}
}

@article{zellers2019hellaswag,
  title={Hellaswag: Can a machine really finish your sentence?},
  author={Zellers, Rowan and Holtzman, Ari and Bisk, Yonatan and Farhadi, Ali and Choi, Yejin},
  journal={arXiv preprint arXiv:1905.07830},
  year={2019}
}

@inproceedings{bisk2020piqa,
  title={Piqa: Reasoning about physical commonsense in natural language},
  author={Bisk, Yonatan and Zellers, Rowan and Gao, Jianfeng and Choi, Yejin and others},
  booktitle={Proceedings of the AAAI conference on artificial intelligence},
  volume={34},
  number={05},
  pages={7432--7439},
  year={2020}
}

@article{sakaguchi2021winogrande,
  title={Winogrande: An adversarial winograd schema challenge at scale},
  author={Sakaguchi, Keisuke and Bras, Ronan Le and Bhagavatula, Chandra and Choi, Yejin},
  journal={Communications of the ACM},
  volume={64},
  number={9},
  pages={99--106},
  year={2021},
  publisher={ACM New York, NY, USA}
}

@article{ren2020codebleu,
  title={Codebleu: a method for automatic evaluation of code synthesis},
  author={Ren, Shuo and Guo, Daya and Lu, Shuai and Zhou, Long and Liu, Shujie and Tang, Duyu and Sundaresan, Neel and Zhou, Ming and Blanco, Ambrosio and Ma, Shuai},
  journal={arXiv preprint arXiv:2009.10297},
  year={2020}
}

@inproceedings{guo2022unixcoder,
  title={Unixcoder: Unified cross-modal pre-training for code representation},
  author={Guo, Daya and Lu, Shuai and Duan, Nan and Wang, Yanlin and Zhou, Ming and Yin, Jian},
  booktitle={Proceedings of the 60th Annual Meeting of the Association for Computational Linguistics (Volume 1: Long Papers)},
  pages={7212--7225},
  year={2022}
}

@article{agarwal2025gpt,
  title={gpt-oss-120b \& gpt-oss-20b model card},
  author={Agarwal, Sandhini and Ahmad, Lama and Ai, Jason and Altman, Sam and Applebaum, Andy and Arbus, Edwin and Arora, Rahul K and Bai, Yu and Baker, Bowen and Bao, Haiming and others},
  journal={arXiv preprint arXiv:2508.10925},
  year={2025}
}

@article{yang2025qwen3,
  title={Qwen3 technical report},
  author={Yang, An and Li, Anfeng and Yang, Baosong and Zhang, Beichen and Hui, Binyuan and Zheng, Bo and Yu, Bowen and Gao, Chang and Huang, Chengen and Lv, Chenxu and others},
  journal={arXiv preprint arXiv:2505.09388},
  year={2025}
}

@article{dekoninck2024constat,
  title={Constat: Performance-based contamination detection in large language models},
  author={Dekoninck, Jasper and M{\"u}ller, Mark N and Vechev, Martin},
  journal={Advances in Neural Information Processing Systems},
  volume={37},
  pages={92420--92464},
  year={2024}
}

@article{grattafiori2024llama,
  title={The llama 3 herd of models},
  author={Grattafiori, Aaron and Dubey, Abhimanyu and Jauhri, Abhinav and Pandey, Abhinav and Kadian, Abhishek and Al-Dahle, Ahmad and Letman, Aiesha and Mathur, Akhil and Schelten, Alan and Vaughan, Alex and others},
  journal={arXiv preprint arXiv:2407.21783},
  year={2024}
}

@article{yang2024qwen2,
  title={Qwen2. 5-math technical report: Toward mathematical expert model via self-improvement},
  author={Yang, An and Zhang, Beichen and Hui, Binyuan and Gao, Bofei and Yu, Bowen and Li, Chengpeng and Liu, Dayiheng and Tu, Jianhong and Zhou, Jingren and Lin, Junyang and others},
  journal={arXiv preprint arXiv:2409.12122},
  year={2024}
}

@article{achiam2023gpt,
  title={GPT-4 technical report},
  author={Achiam, J},
  journal={arXiv preprint arXiv:2303.08774},
  year={2023}
}

@article{comanici2025gemini,
  title={Gemini 2.5: Pushing the frontier with advanced reasoning, multimodality, long context, and next generation agentic capabilities},
  author={Comanici, Gheorghe and Bieber, Eric and Schaekermann, Mike and Pasupat, Ice and Sachdeva, Noveen and Dhillon, Inderjit and Blistein, Marcel and Ram, Ori and Zhang, Dan and Rosen, Evan and others},
  journal={arXiv preprint arXiv:2507.06261},
  year={2025}
}

@article{guo2025deepseek,
  title={Deepseek-r1: Incentivizing reasoning capability in llms via reinforcement learning},
  author={Guo, Daya and Yang, Dejian and Zhang, Haowei and Song, Junxiao and Wang, Peiyi and Zhu, Qihao and Xu, Runxin and Zhang, Ruoyu and Ma, Shirong and Bi, Xiao and others},
  journal={arXiv preprint arXiv:2501.12948},
  year={2025}
}

@article{bai2023qwen,
  title={Qwen technical report},
  author={Bai, Jinze and Bai, Shuai and Chu, Yunfei and Cui, Zeyu and Dang, Kai and Deng, Xiaodong and Fan, Yang and Ge, Wenbin and Han, Yu and Huang, Fei and others},
  journal={arXiv preprint arXiv:2309.16609},
  year={2023}
}
\bibliographystyle{plainnat}

\newpage
\appendix
\onecolumn

\section{Detailed Methodology}

\subsection{Olmo3 Instruct Training Datasets}\label{app:olmo3-data}

We embed 1\% of the data Olmo3 Base was trained on (`Dolma3\_6T-mix-1025'); and 1\% of Dolmino-mix-1025 (100B tokens, 2.7 GB), a high quality dataset Olmo3-base is trained on. We excluded long-context data.
We also embed the following three high quality datasets used to finetune Olmo3 Base into Olmo3 Instruct: Dolci3 SFT (2M input-output prompts, 3.08 GB), Dolci3 Instruct DPO (260k preference pairs, 811 MB), and Dolci3 Instruct RL (169k prompts, 483 MB).

\begin{table}[h]
\centering
\small
\caption{Datasets processed for contamination analysis. All text chunks are filtered to 50--2,048 tokens. Pretraining data is sampled at 1\% using stratified reservoir sampling to preserve source distribution topology.}
\label{tab:dataset_methodology}
\begin{tabularx}{\linewidth}{X r r l}
\toprule
\textbf{Dataset} & \textbf{Orig. Size} & \textbf{Sample} & \textbf{Sampling Method} \\
\midrule
\multicolumn{4}{l}{\textit{Pretraining Data}} \\
Dolma3\_6T-mix & 6.0 TB & 1.0\% & Stratified Reservoir (Hierarchical) \\
Dolmino-mix & 250 GB & 1.0\% & Stratified Reservoir (Hierarchical) \\
\midrule
\multicolumn{4}{l}{\textit{Instruction Tuning Data}} \\
Dolci3 SFT & 3.08 GB & 100.0\% & Full Ingestion \\
Dolci3 DPO & 811 MB & 100.0\% & Full Ingestion \\
Dolci3 RL & 483 MB & 100.0\% & Full Ingestion \\
\bottomrule
\end{tabularx}
\end{table}

\subsection{Exact duplicates of ZebraLogic test data in Olmo corpus}
\label{app:exact_zebra}
\citet{olmo2025olmo3} reports a ZebraLogic benchmark score. Surprisingly, we found that the Olmo Instruct RL dataset contains exact duplicates of 70\% of ZebraLogic problems-and-solutions.

\begin{figure}[ht]
  \vskip 0.2in
  \begin{center}
    \centerline{\includegraphics[width=\columnwidth]{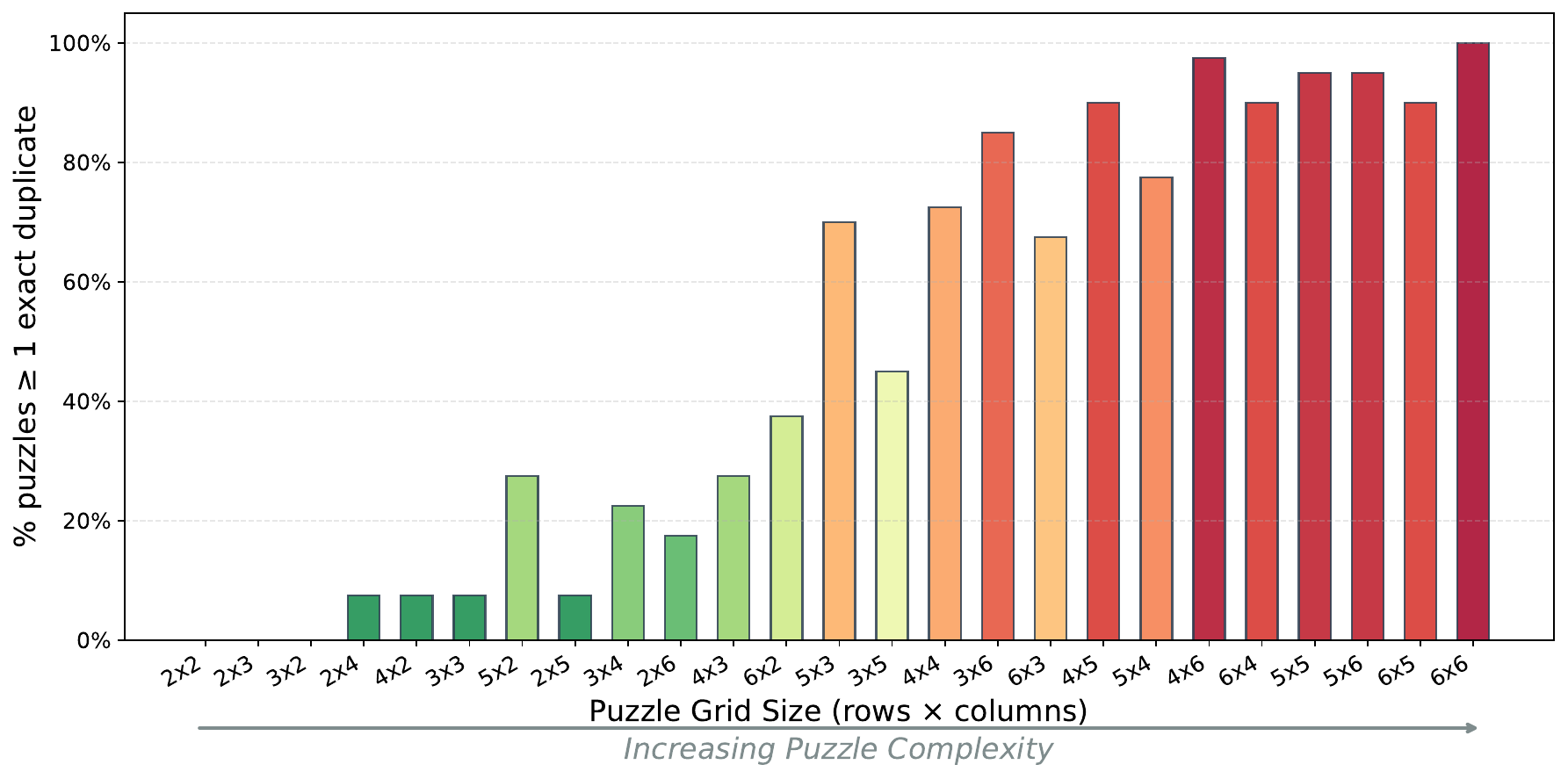}}
    \caption{On the y-axis we plot the following statistic: for each ZebraLogic benchmark datapoint we check among the top 10 highest cosine similarity training datapoints if any of those samples is an exact duplicate, we then calculate the proportion of benchmark datapoints (of a given grid size) that have at least one exact duplicate. On the x-axis we plot puzzle grid size.}
    \label{fig:zebralogic-exact-dupe}
  \end{center}
\end{figure}

In Figure \ref{fig:zebralogic-exact-dupe} we see that for puzzles of $4 \times 4$ or larger, for ~70\% or more problems, there exists at least one exact duplicate in the Olmo3 Instruct RL dataset (\texttt{Dolci-Instruct-RL}). In total, we find corpus duplicates for at least 49.5\% of dataset tasks.

The model card of Olmo3 benchmarks the performance of Olmo3 Instruct after SFT training as 18\%, after DPO training as 28.4\%, and after RL training as 32.9\%. The model is improved in that order. 
We do not find exact or semantic duplicates of ZebraLogic in the DPO dataset, so we ascribe the improvement between SFT training and DPO training to general logical reasoning improvement. 
However, the increase from 28.4\% to 32.9\% (possibly on harder problems) is likely due to directly training on ZebraLogic data.

We also briefly investigated the existence of semantic duplicates of ZebraLogic, but did not find substantial amounts.

\subsection{Cosine Similarity discussion}\label{app:cos-sim}

We embed both Olmo3 corpus data and benchmark data, and calculate the cosine similarity between data points in MBPP and CodeForces. This consisted of checking the 10 highest cosine similarity matches for each test point (adding up to 10,000 annotations for the entire benchmark).

When comparing with pretraining data sets, we split MBPP into inputs and outputs. 
When comparing MBPP with instruct data, we join MBPP inputs and outputs to match the prompt response format in the instruct data. 
We here present cosine similarity matches between corpus data and MuSR and ZebraLogic.
We take the cosine similarity of the benchmark point embedding with the corpus text embeddings. 

\begin{figure}[th]
    \centering
    \includegraphics[width=\linewidth]{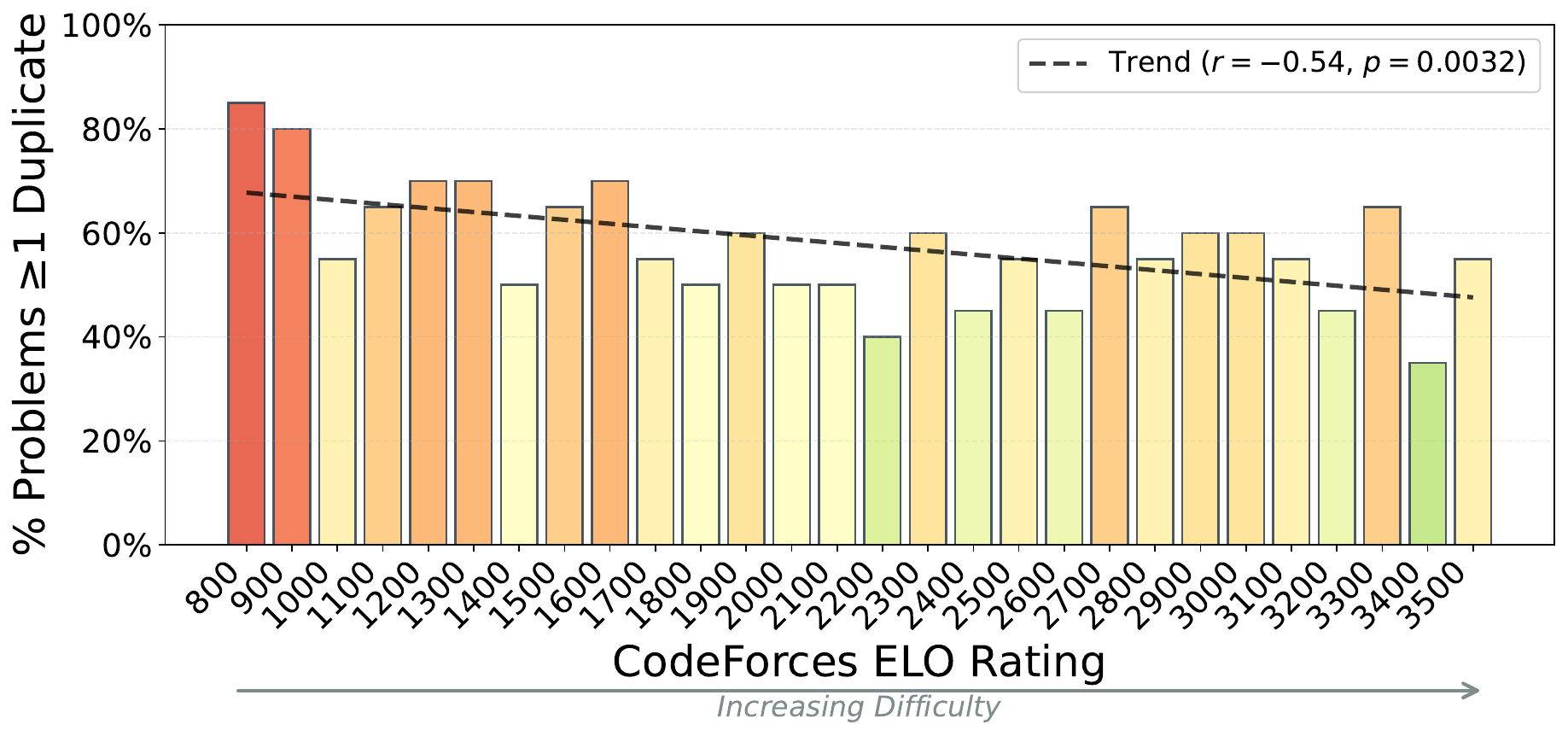}
    \caption{Occurrence of semantic duplicates by problem difficulty (elo). For each benchmark datapoint we check among the top 100 cosine similarity training datapoints if any of those samples is a semantic duplicate and calculate the proportion of all benchmark datapoints that have $\geq 1$ semantic duplicate. We plot the CodeForces difficulty (elo score) on $x$.
    }
    \label{fig:elo-vs-sem-dupe}
\end{figure}

\begin{figure}[ht]
  \centering
  \begin{subfigure}[b]{0.48\columnwidth}
    \includegraphics[width=\linewidth]{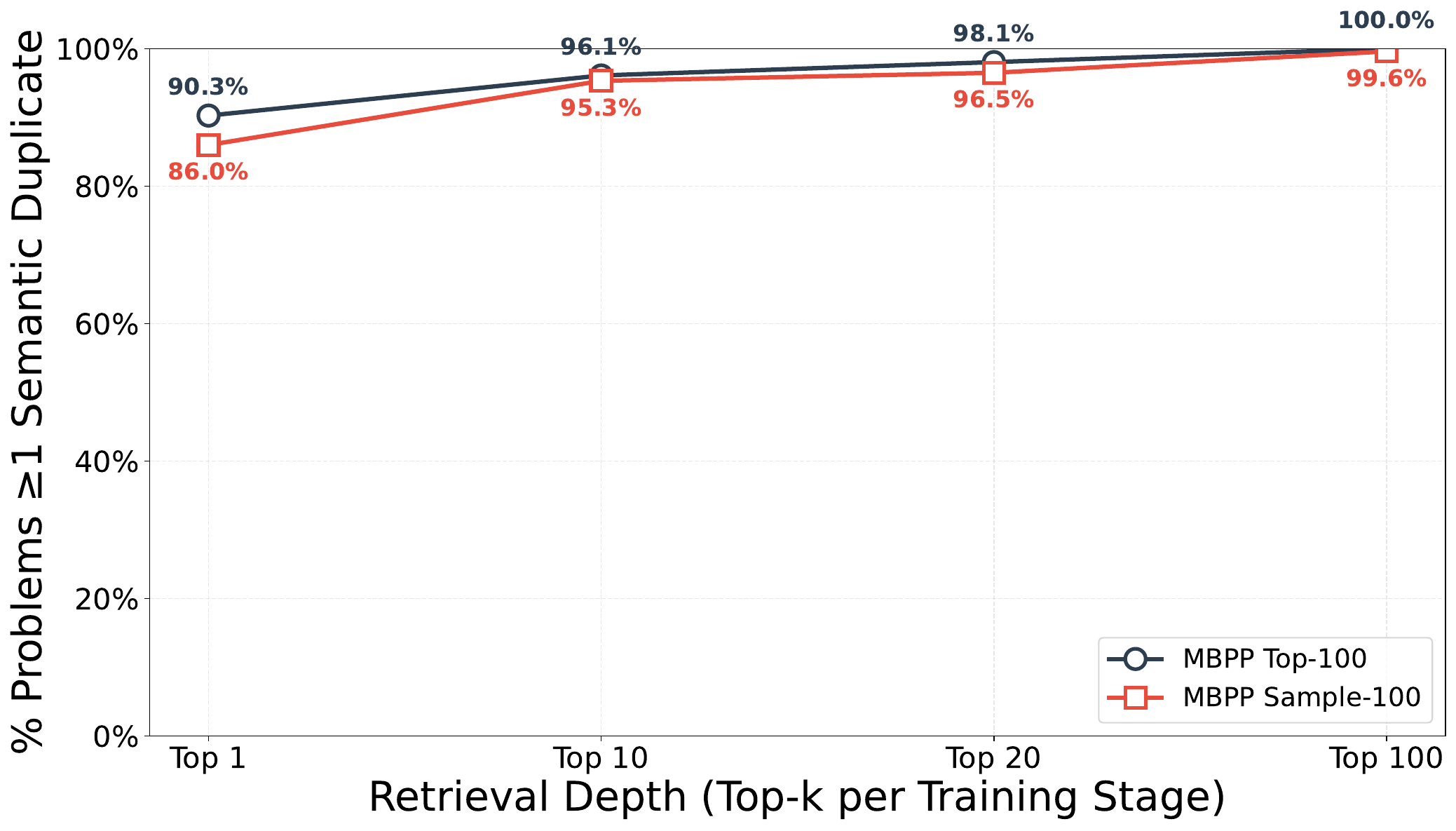}
    \caption{MBPP}
    \label{fig:}
  \end{subfigure}
  \hfill 
  \begin{subfigure}[b]{0.48\columnwidth}
  \includegraphics[width=\linewidth]{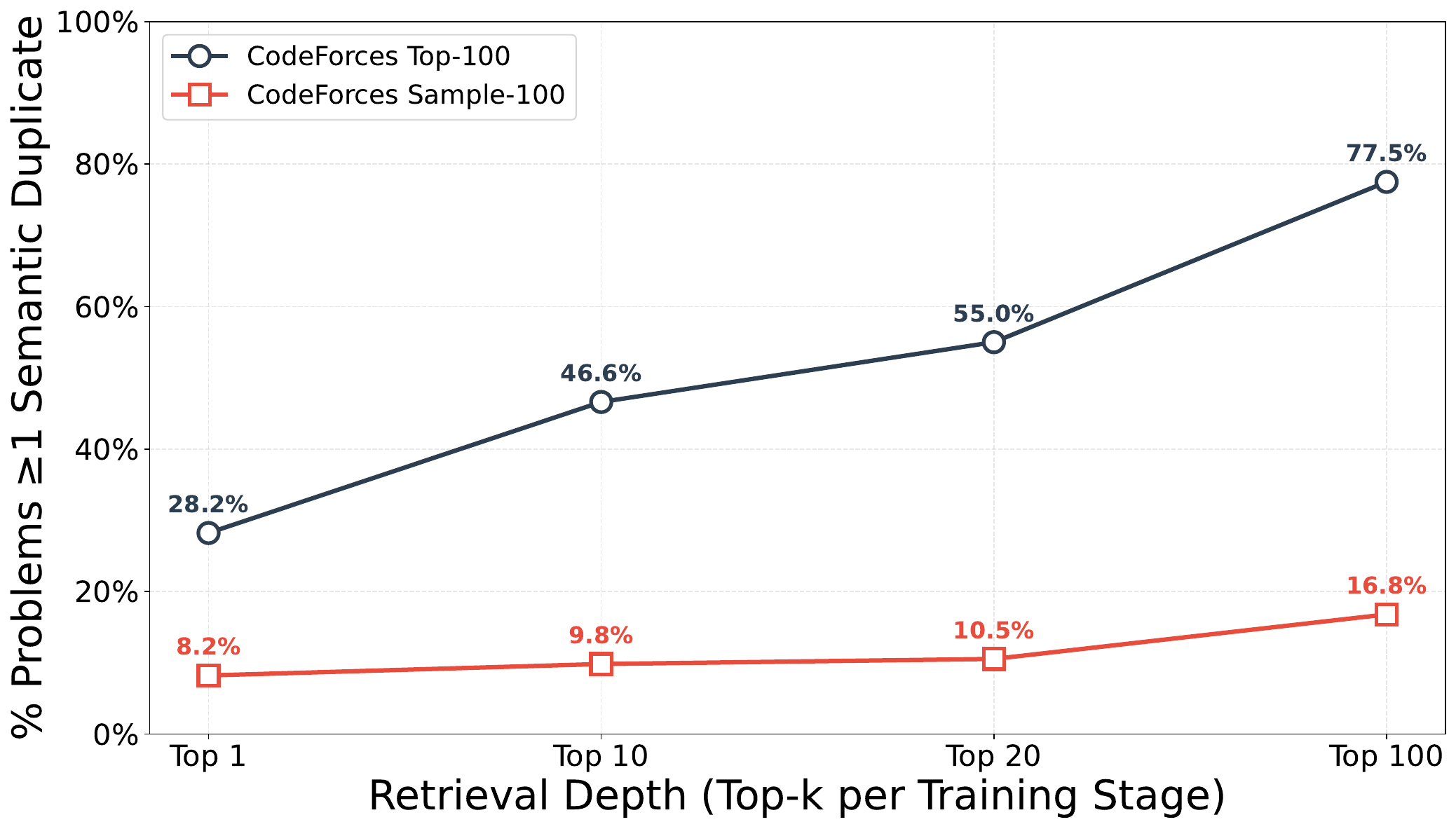}
    \caption{CodeForces}
    \label{fig:}
  \end{subfigure}
  \caption{Occurrence by number of cosine similarity matches investigated. We take the number of semantic duplicates at top-$n$ at each of our dataset comparisons.} 
  \label{fig:scaling-law-sem-dupe}
\end{figure}

Figure~\ref{fig:scaling-law-sem-dupe} shows that semantic duplicate coverage of benchmark test samples increases with retrieval depth across MBPP and CodeForces. MBPP already reaches saturation at Top-1 with 90\% coverage, while CodeForces is 28\% at the same level, increasing up to 77.5\% at Top-100.


\begin{figure*}[!htbp]
    \centering

    \begin{subfigure}{0.19\textwidth}
        \centering
        \includegraphics[width=\linewidth, trim=0 0 0 2, clip]{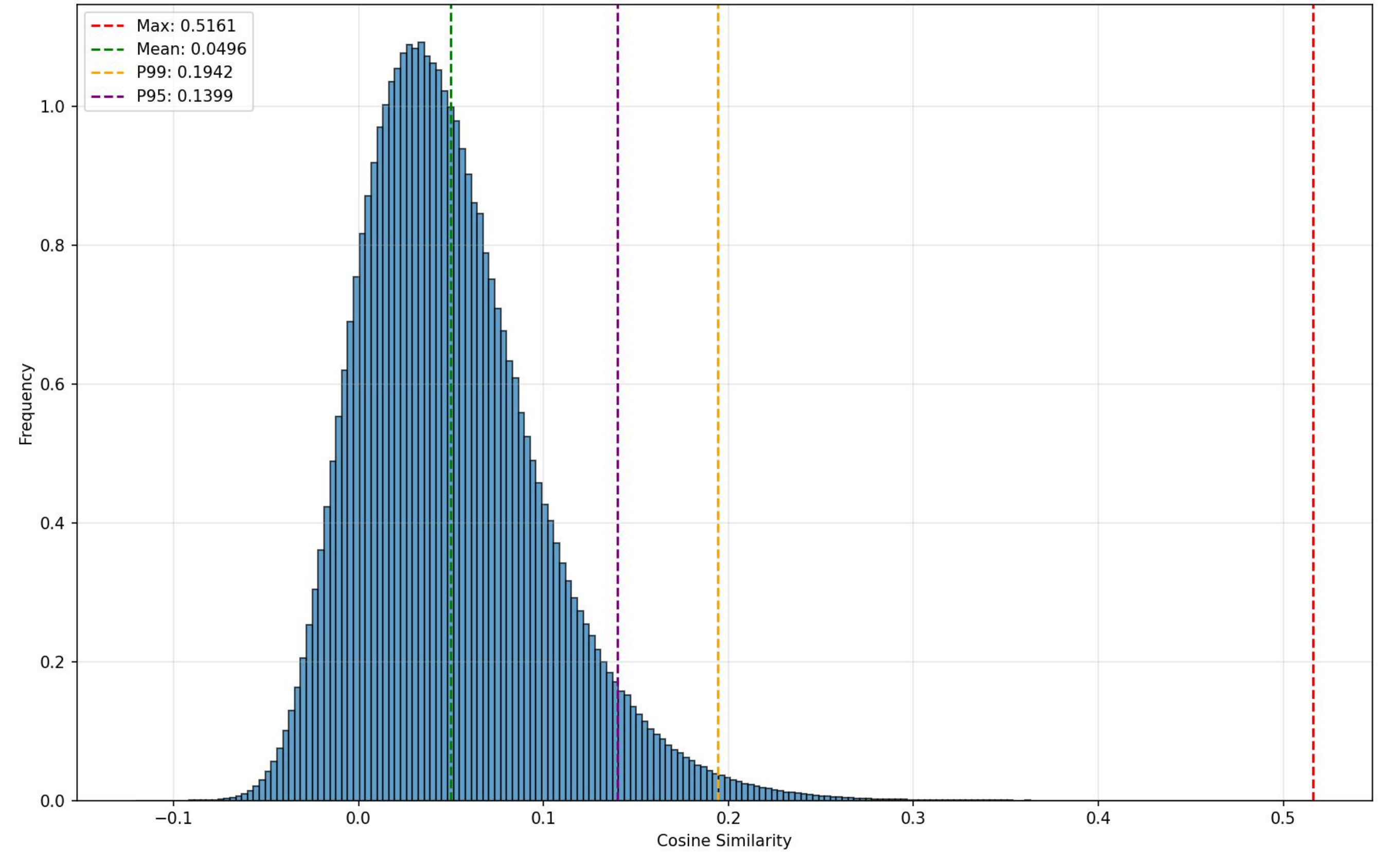}
        \caption{MuSR,Dolma}
    \end{subfigure}
    \hfill
    \begin{subfigure}{0.19\textwidth}
        \centering
        \includegraphics[width=\linewidth, trim=0 0 0 2, clip]{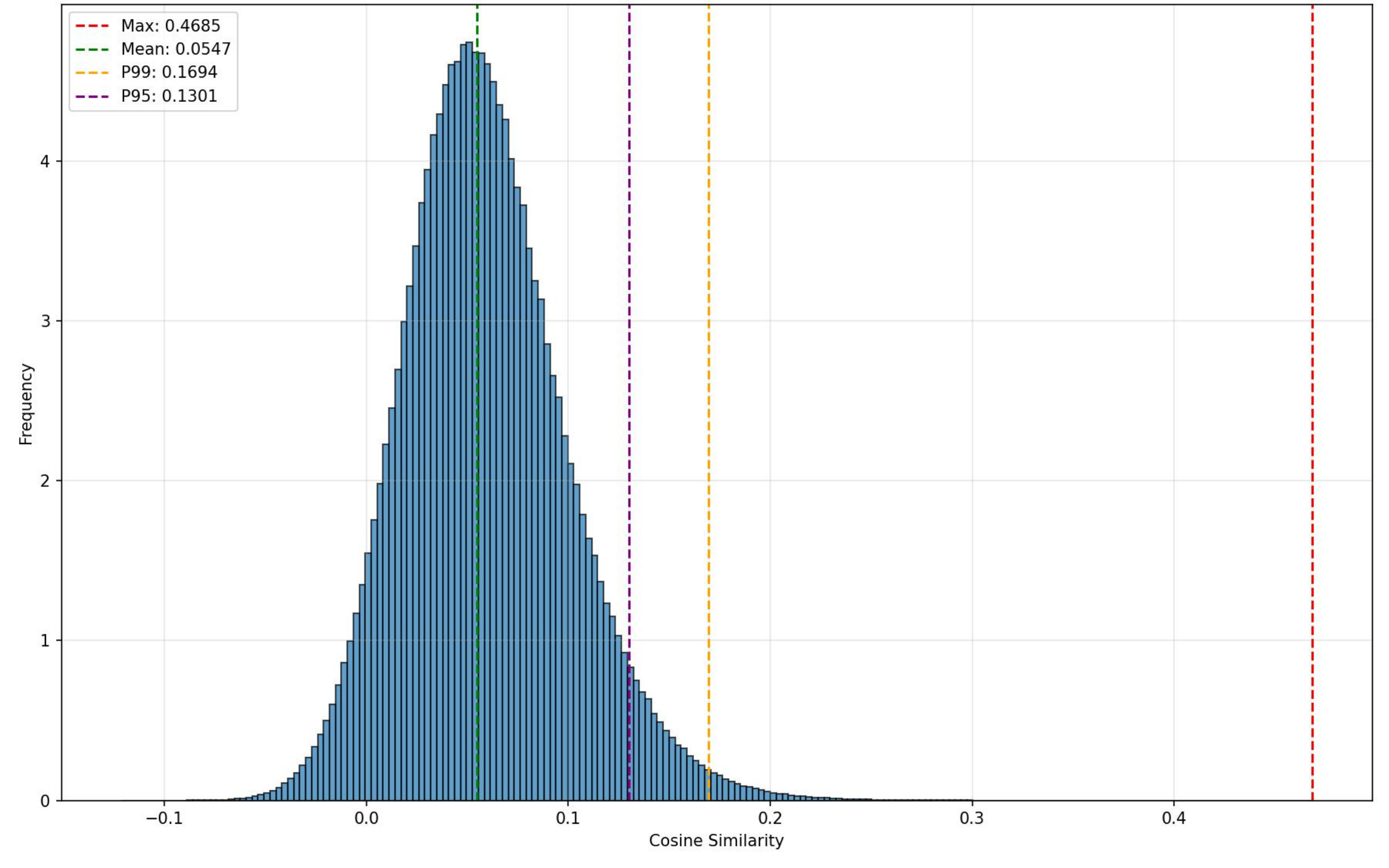}
        \caption{MuSR,Dolmino}
    \end{subfigure}
    \hfill
    \begin{subfigure}{0.19\textwidth}
        \centering
        \includegraphics[width=\linewidth, trim=0 0 0 5, clip]{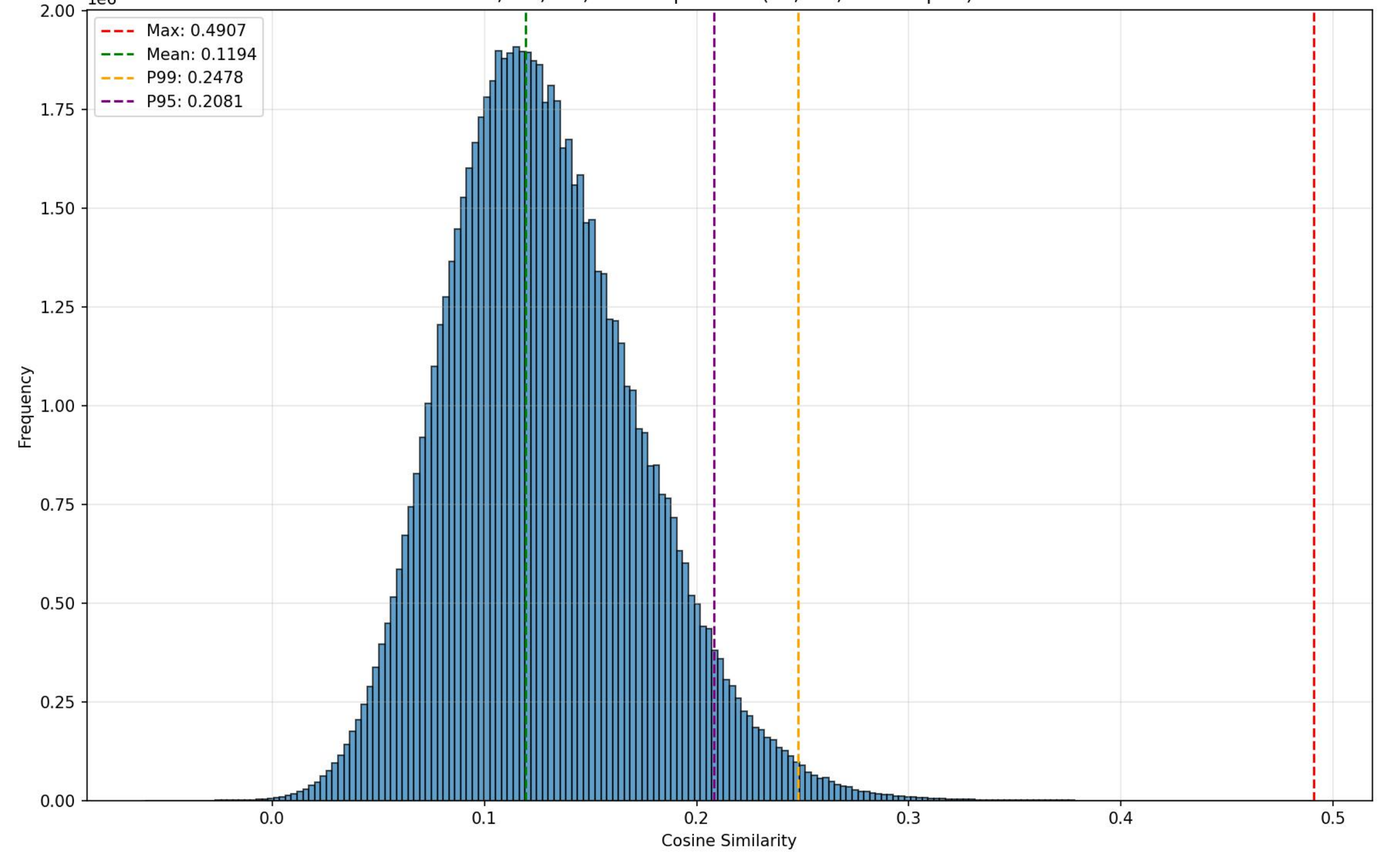}
        \caption{MuSR,DolciSFT}
    \end{subfigure}
    \hfill
    \begin{subfigure}{0.19\textwidth}
        \centering
        \includegraphics[width=\linewidth, trim=0 0 0 8, clip]{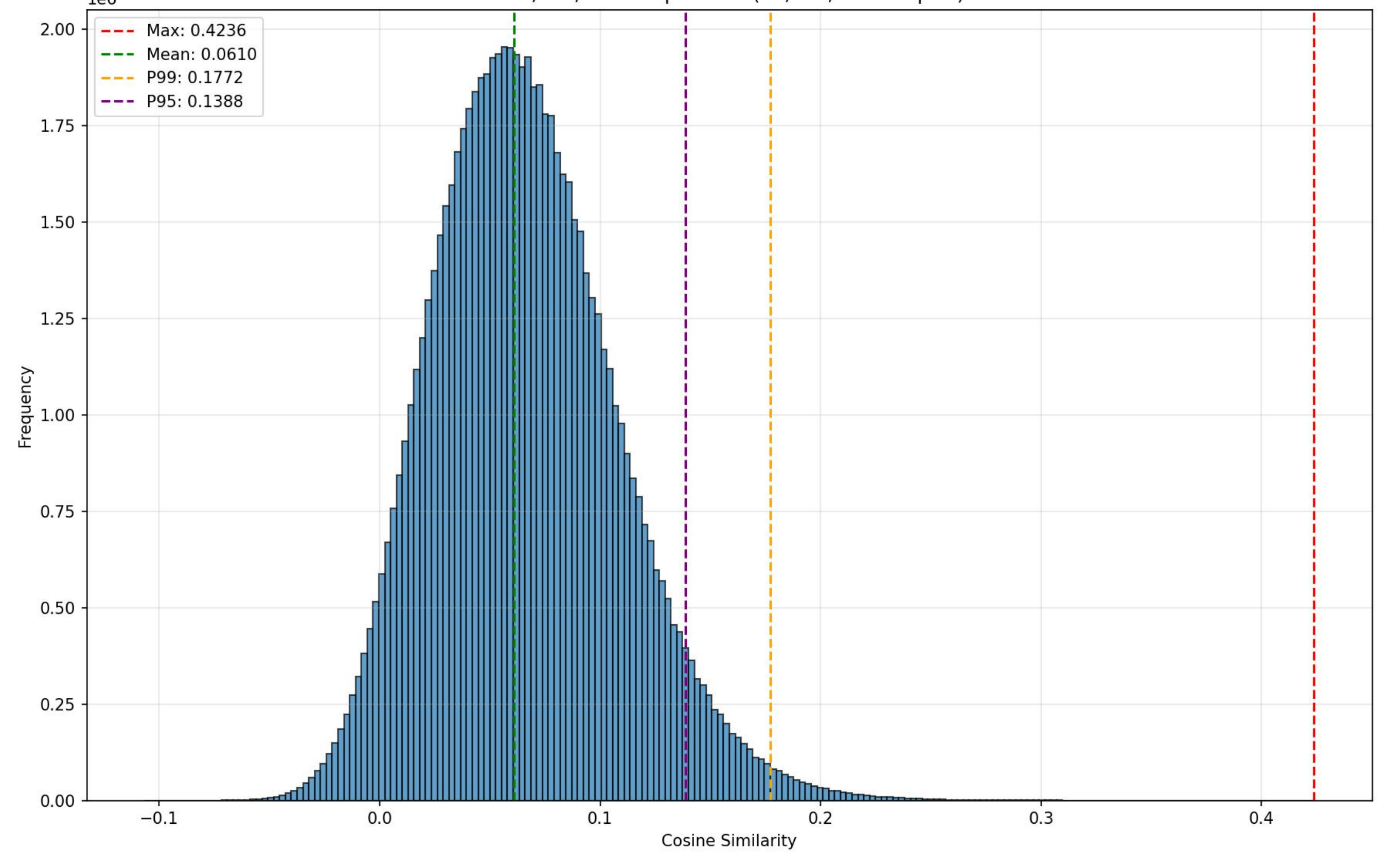}
        \caption{MuSR,DolciDPO}
    \end{subfigure}
    \hfill
    \begin{subfigure}{0.19\textwidth}
        \centering
        \includegraphics[width=\linewidth, trim=0 0 0 8, clip]{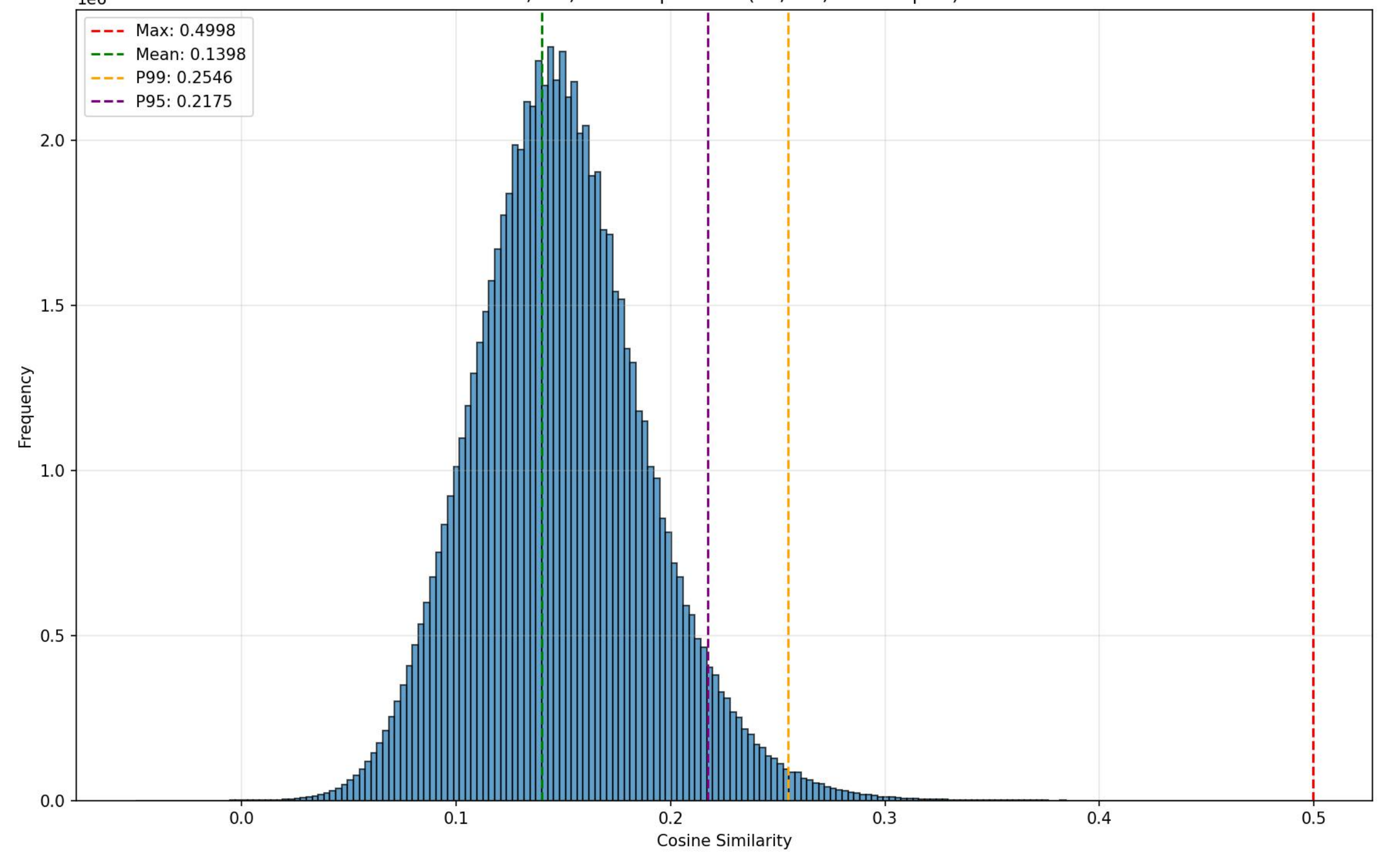}
        \caption{MuSR,DolciRL}
    \end{subfigure}

    \vspace{0.5em}
    
    \begin{subfigure}{0.19\textwidth}
        \centering
        \includegraphics[width=\linewidth, trim=0 0 0 4, clip]{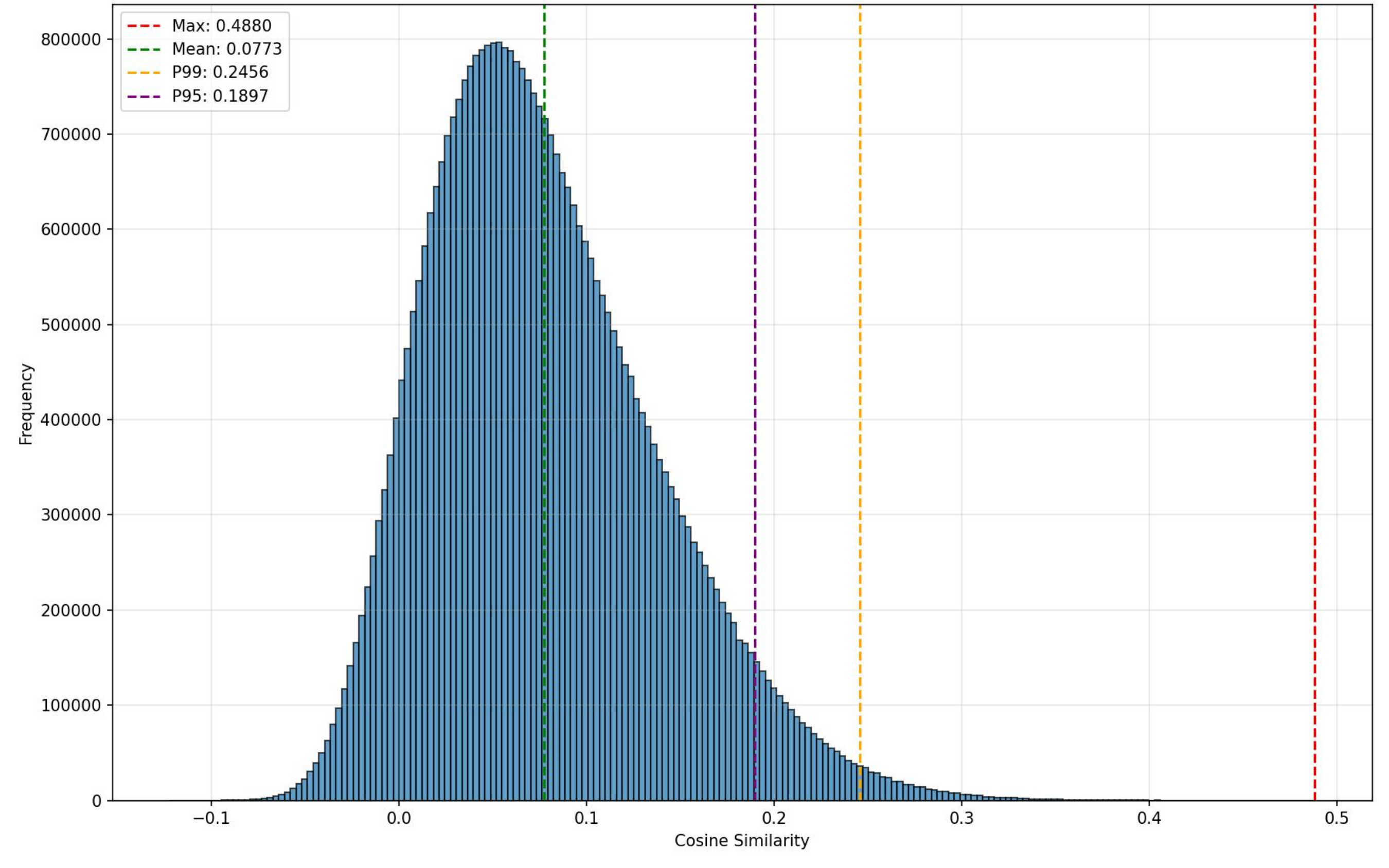}
        \caption{Zebra, Dolma}
    \end{subfigure}
    \hfill
    \begin{subfigure}{0.19\textwidth}
        \centering
        \includegraphics[width=\linewidth, trim=0 0 0 4, clip]{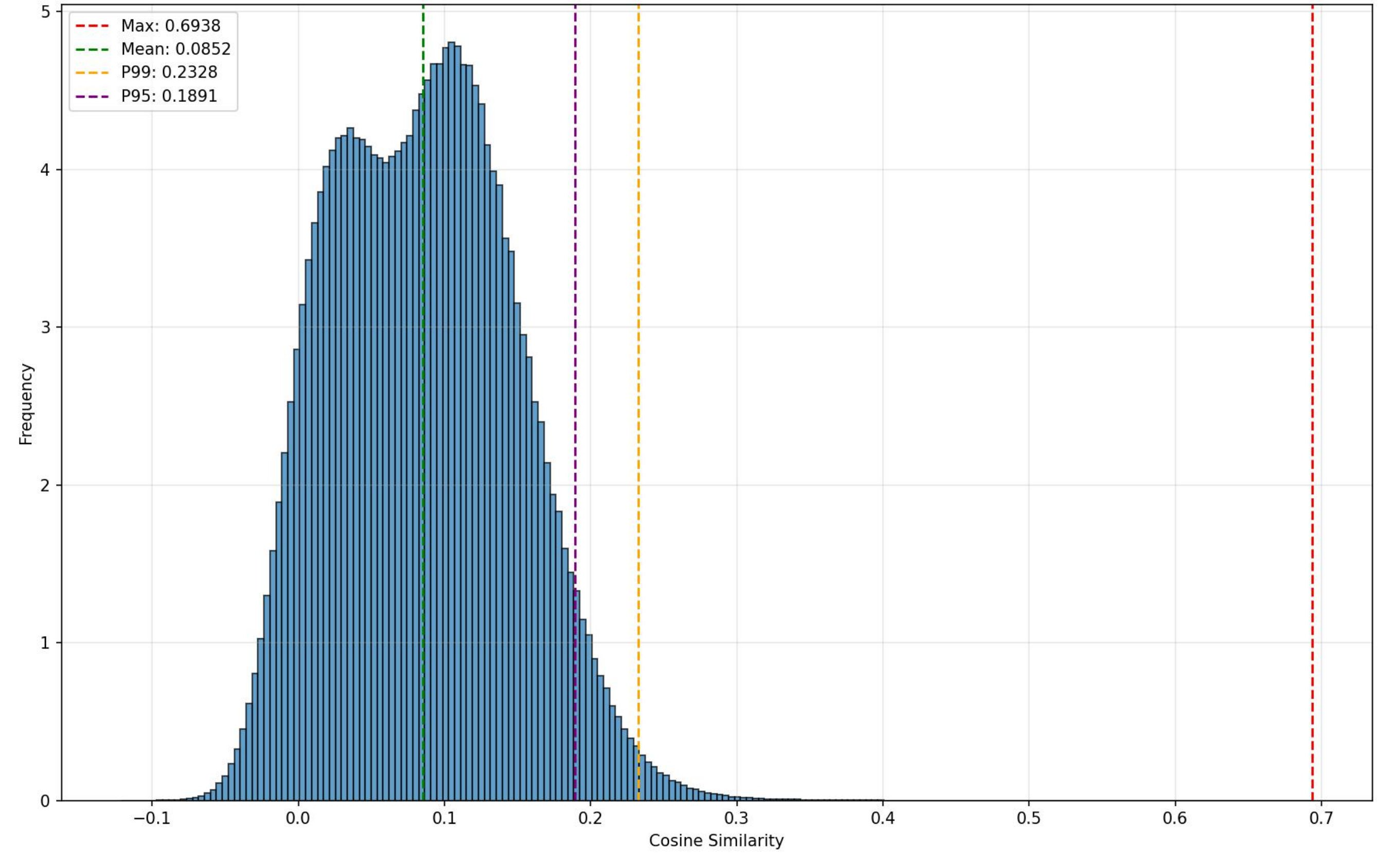}
        \caption{Zebra, Dolmino}
    \end{subfigure}
    \hfill
    \begin{subfigure}{0.19\textwidth}
        \centering
        \includegraphics[width=\linewidth, trim=0 0 0 8, clip]{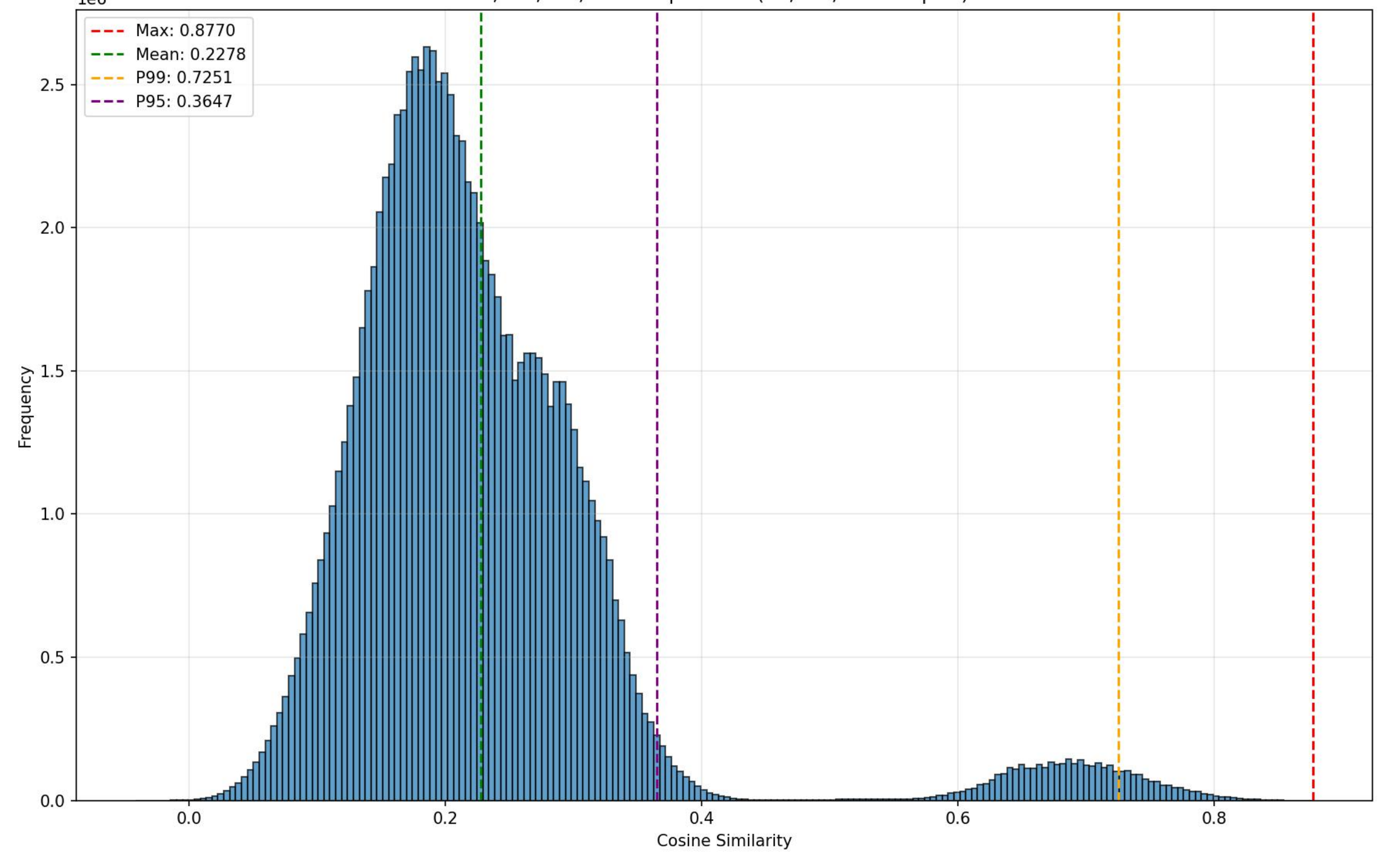}
        \caption{Zebra, DolciSFT}
    \end{subfigure}
    \hfill
    \begin{subfigure}{0.19\textwidth}
        \centering
        \includegraphics[width=\linewidth, trim=0 0 0 8, clip]{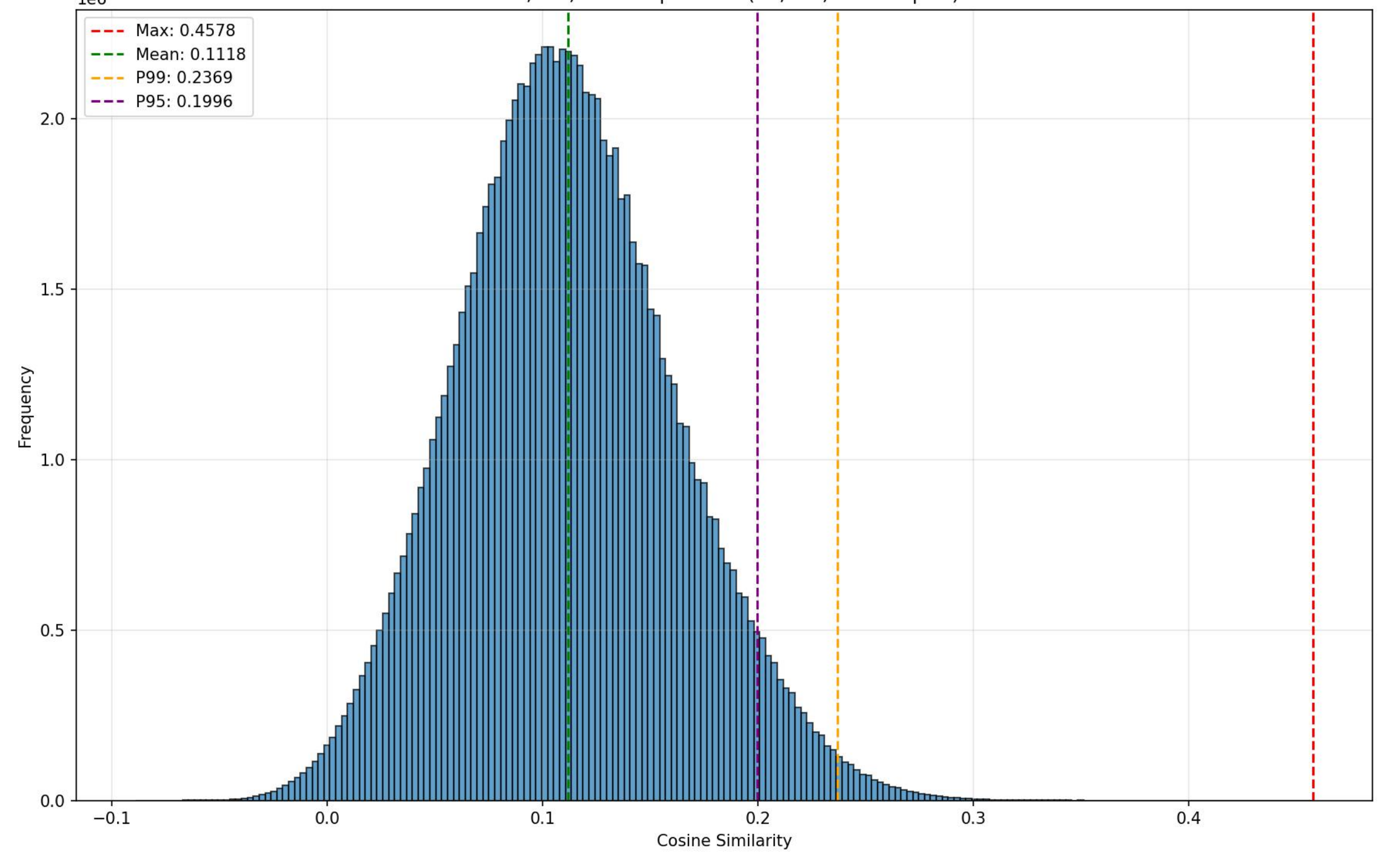}
        \caption{Zebra, DolciDPO}
    \end{subfigure}
    \hfill
    \begin{subfigure}{0.19\textwidth}
        \centering
        \includegraphics[width=\linewidth, trim=0 0 0 5, clip]{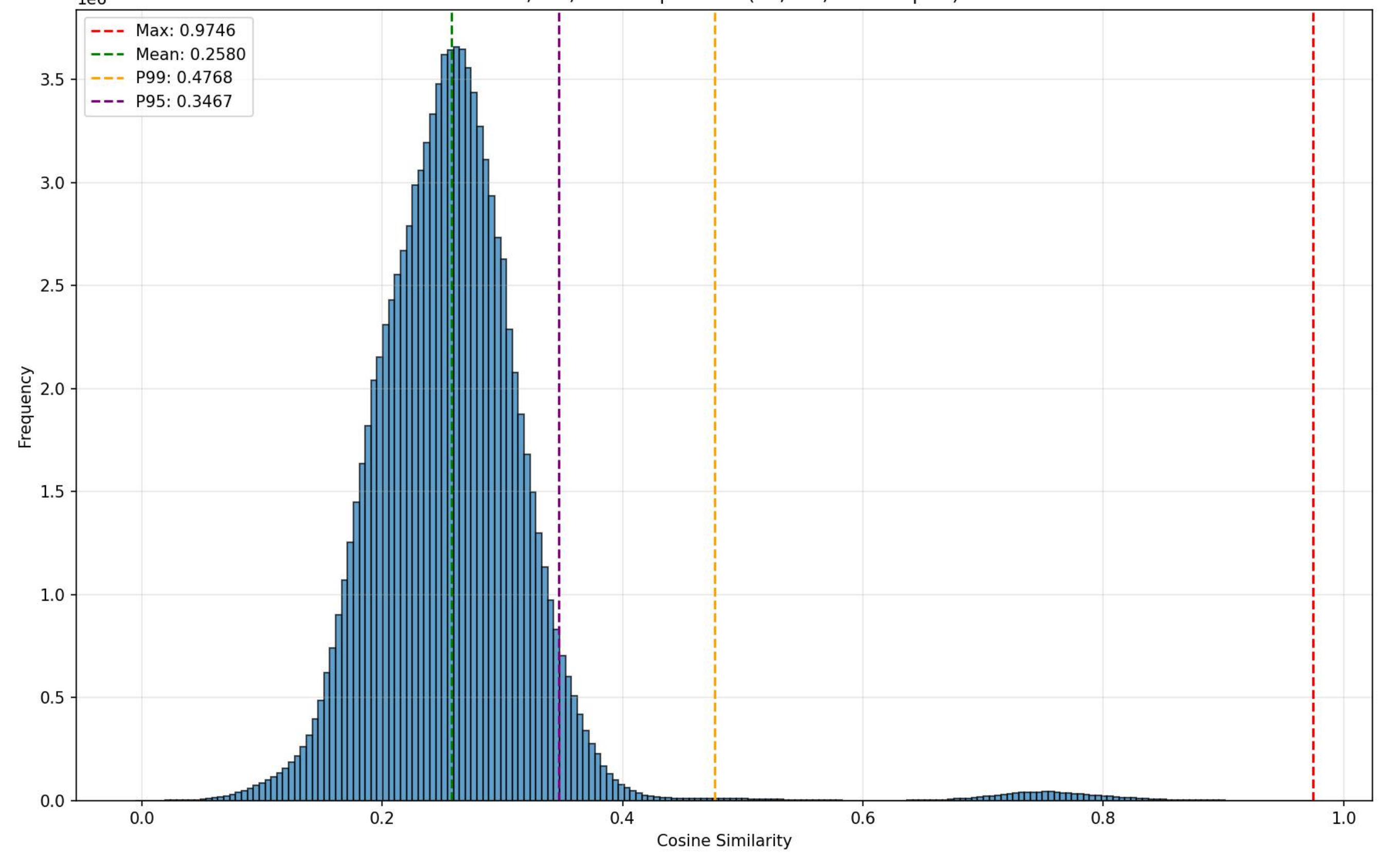}
        \caption{Zebra, DolciRL}
    \end{subfigure}

    \begin{subfigure}{0.19\textwidth}
        \centering
        \includegraphics[width=\linewidth]{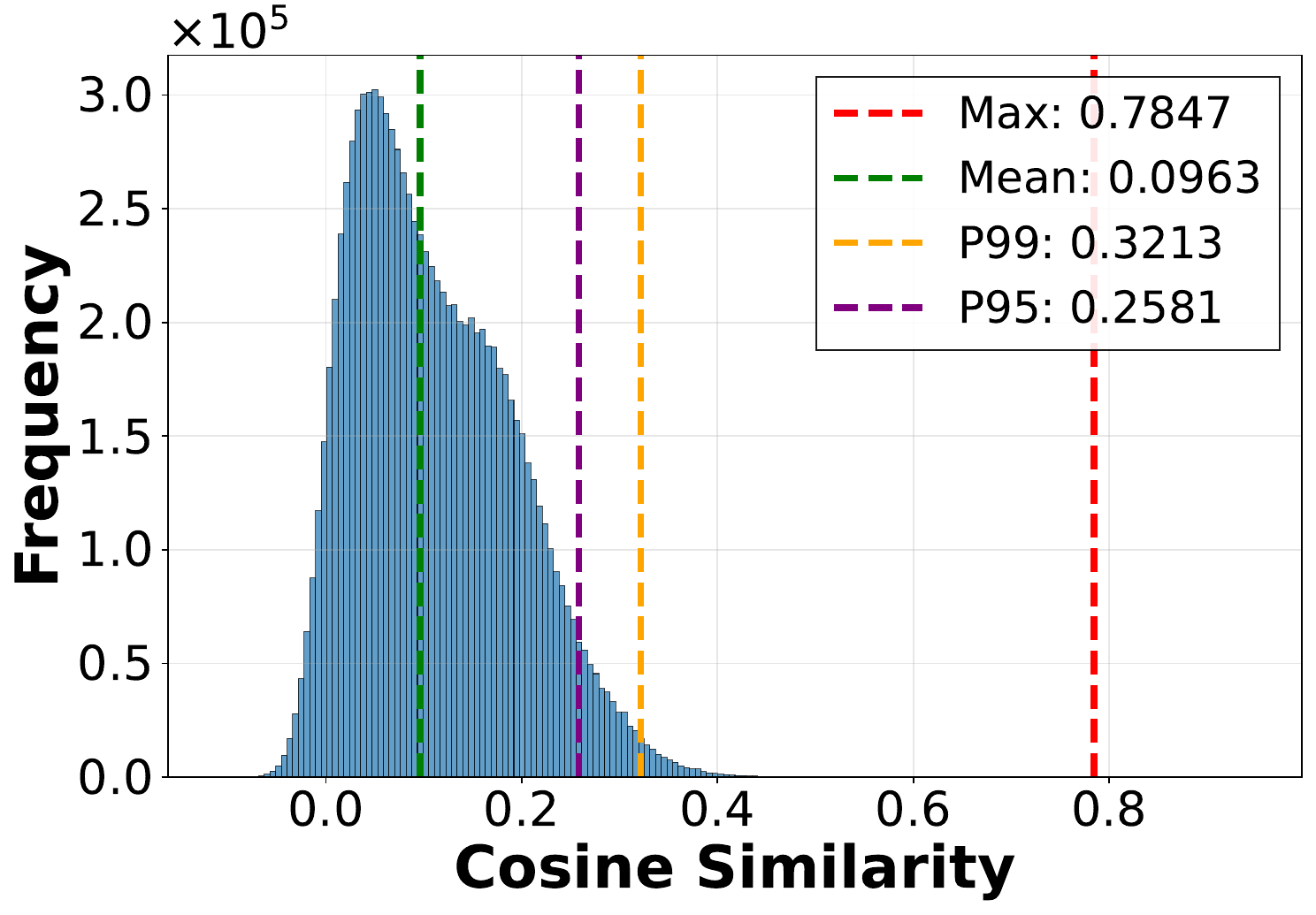}
        \caption{MBPP,Dolma}
    \end{subfigure}
    \hfill
    \begin{subfigure}{0.19\textwidth}
        \centering
        \includegraphics[width=\linewidth]{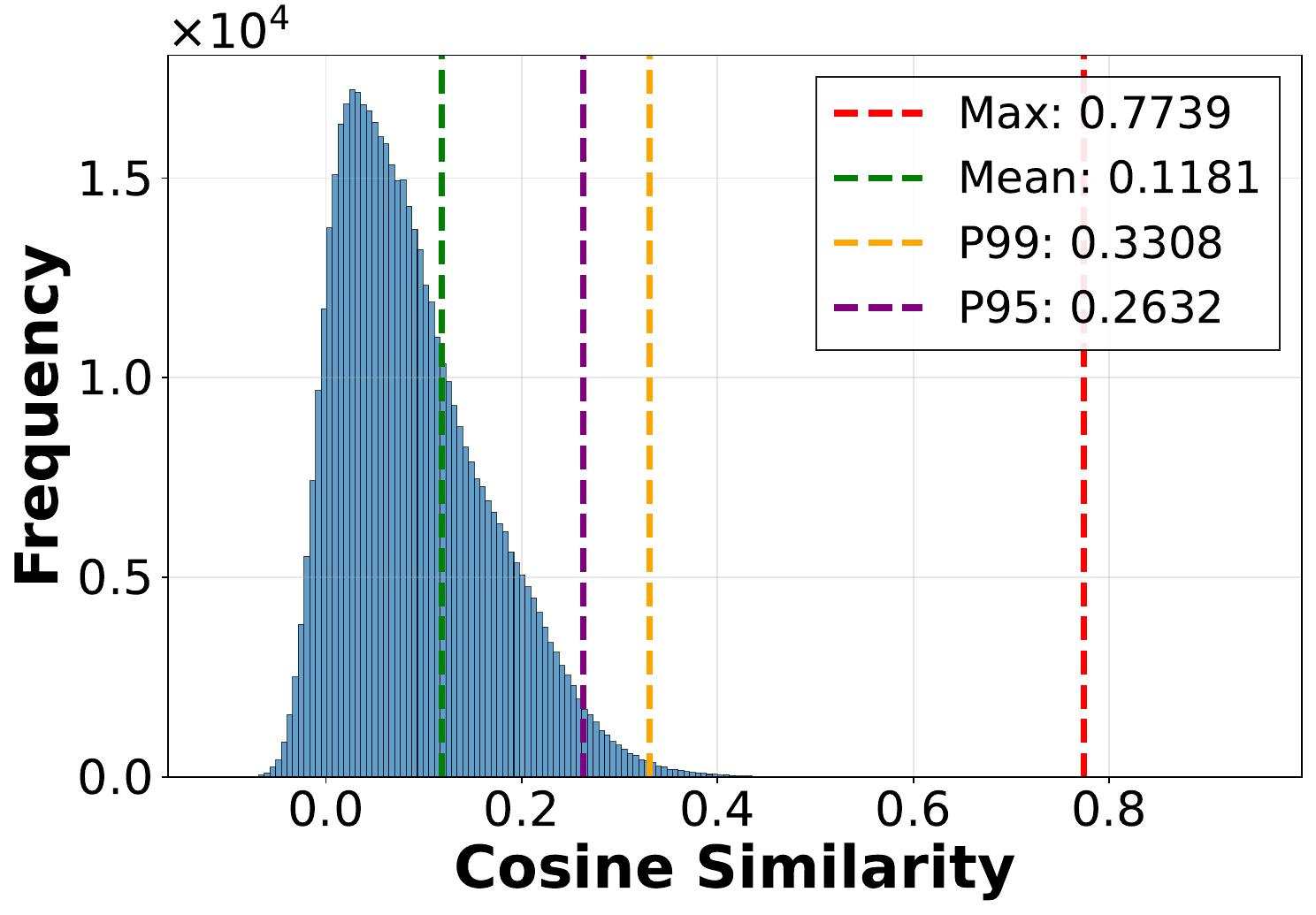}
        \caption{MBPP,Dolmino}
    \end{subfigure}
    \hfill
    \begin{subfigure}{0.19\textwidth}
        \centering
        \includegraphics[width=\linewidth]{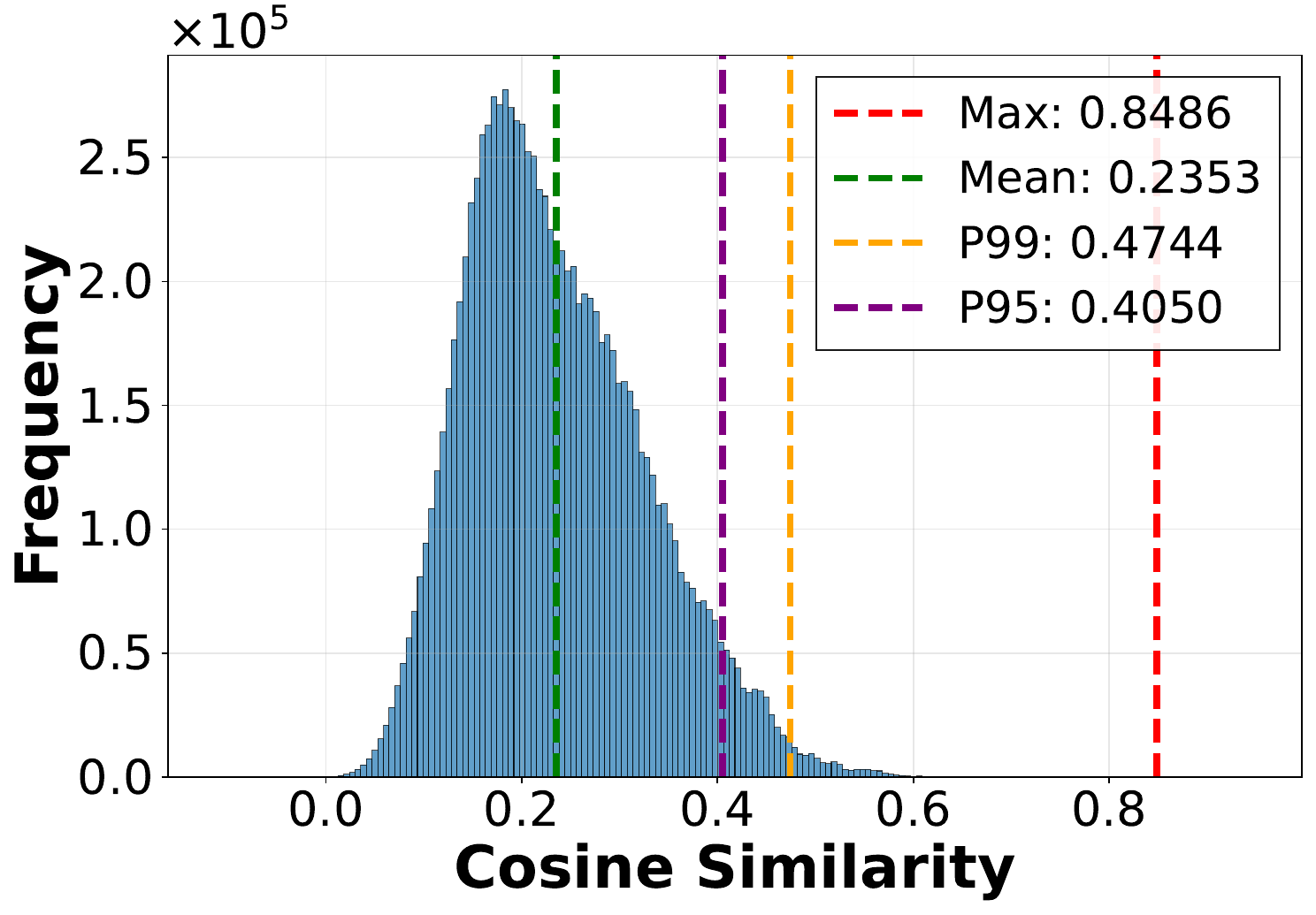}
        \caption{MBPP,DolciSFT}
    \end{subfigure}
    \hfill
    \begin{subfigure}{0.19\textwidth}
        \centering
        \includegraphics[width=\linewidth]{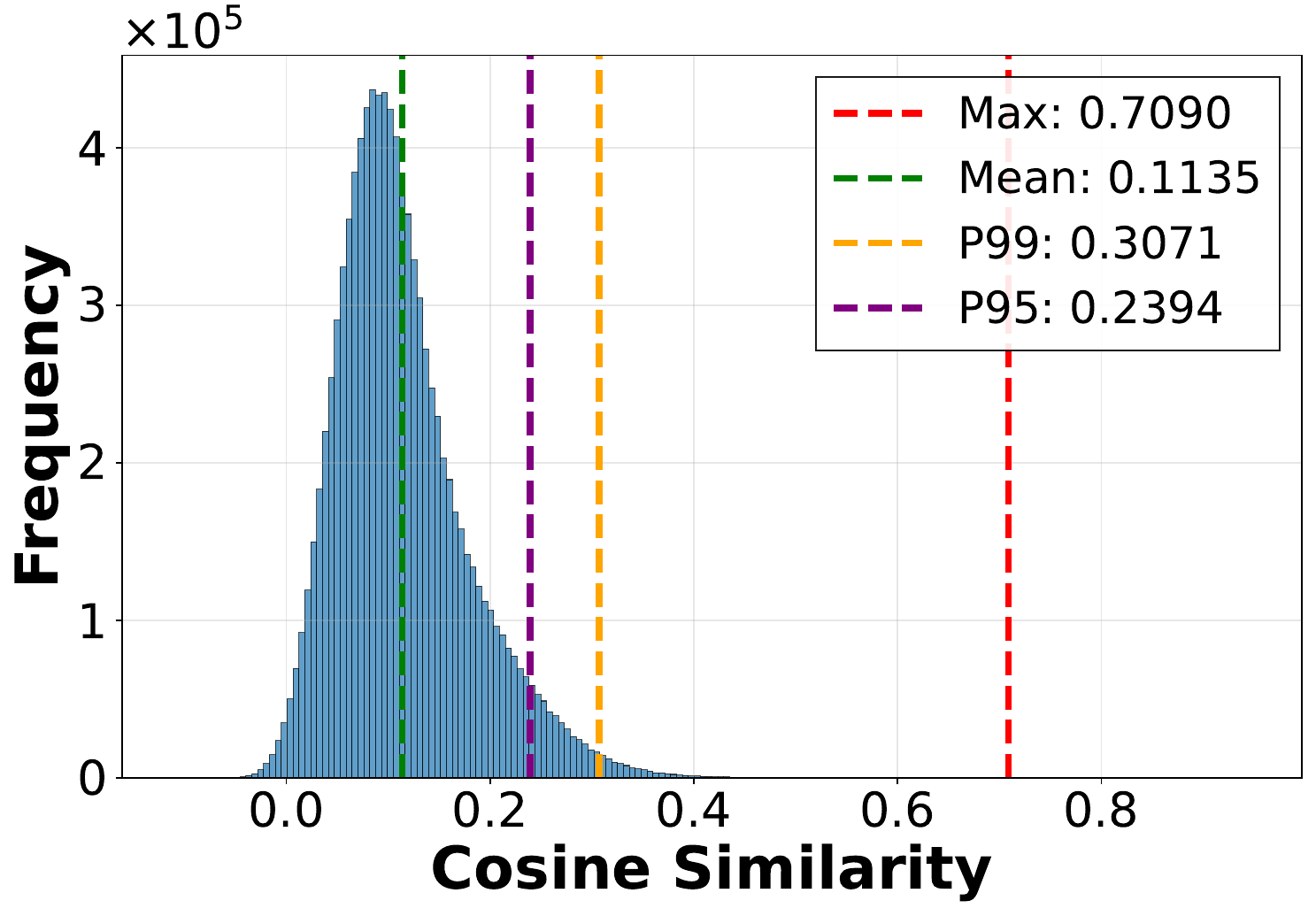}
        \caption{MBPP,DolciDPO}
    \end{subfigure}
    \hfill
    \begin{subfigure}{0.19\textwidth}
        \centering
        \includegraphics[width=\linewidth]{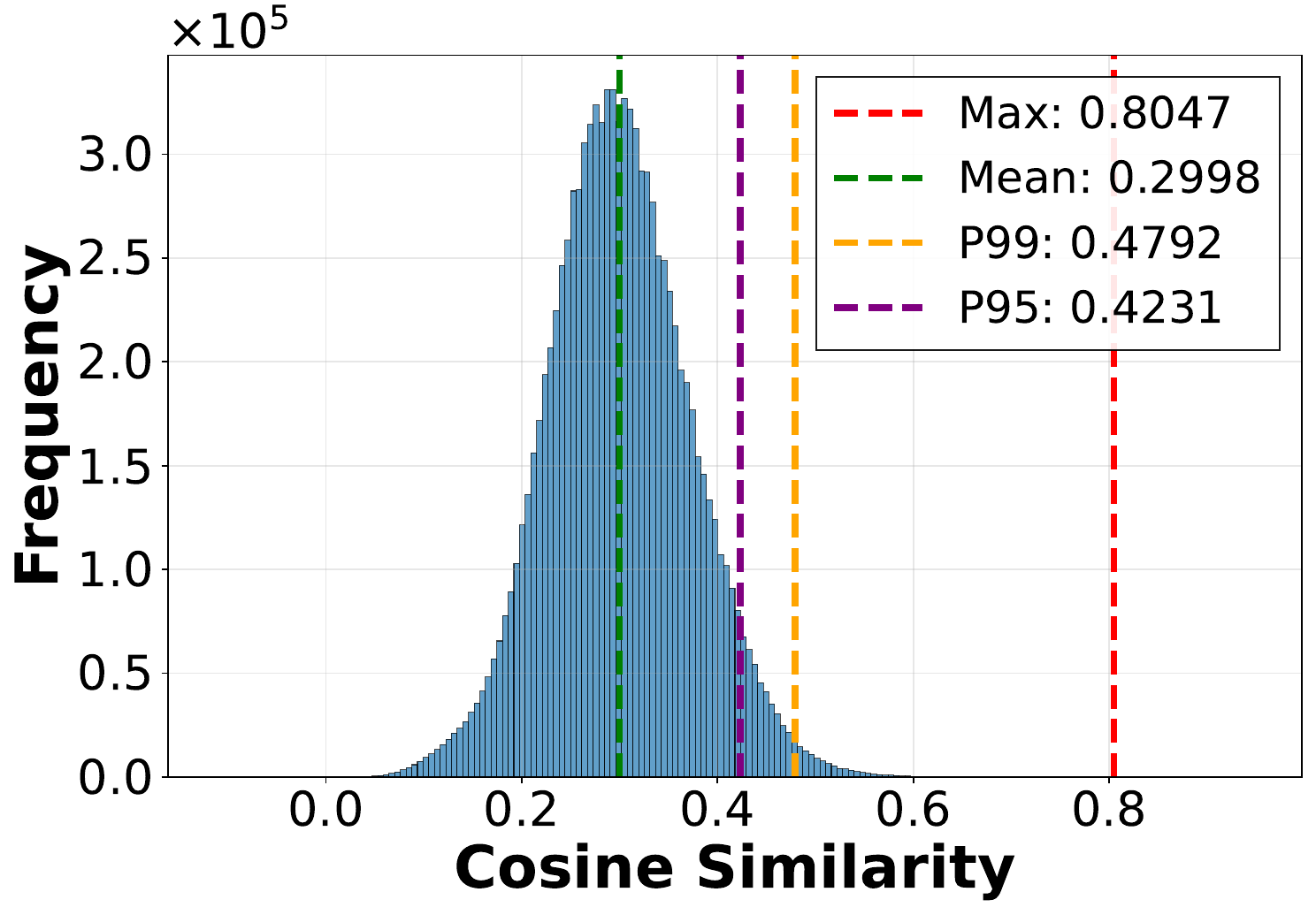}
        \caption{MBPP,DolciRL}
    \end{subfigure}

    \vspace{0.5em}

    \begin{subfigure}{0.19\textwidth}
        \centering
        \includegraphics[width=\linewidth]{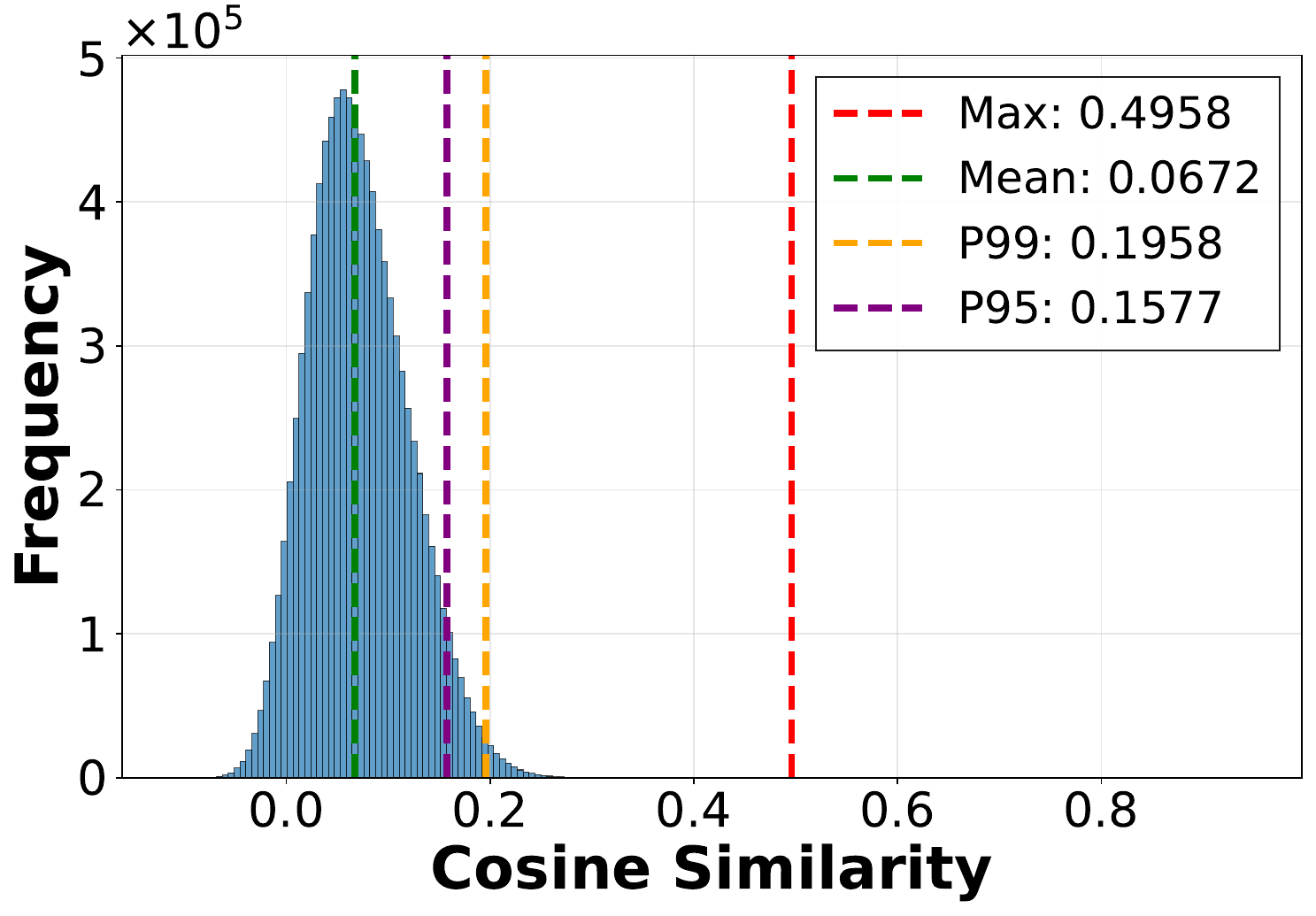}
        \caption{CodeF,Dolma}
    \end{subfigure}
    \hfill
    \begin{subfigure}{0.19\textwidth}
        \centering
        \includegraphics[width=\linewidth]{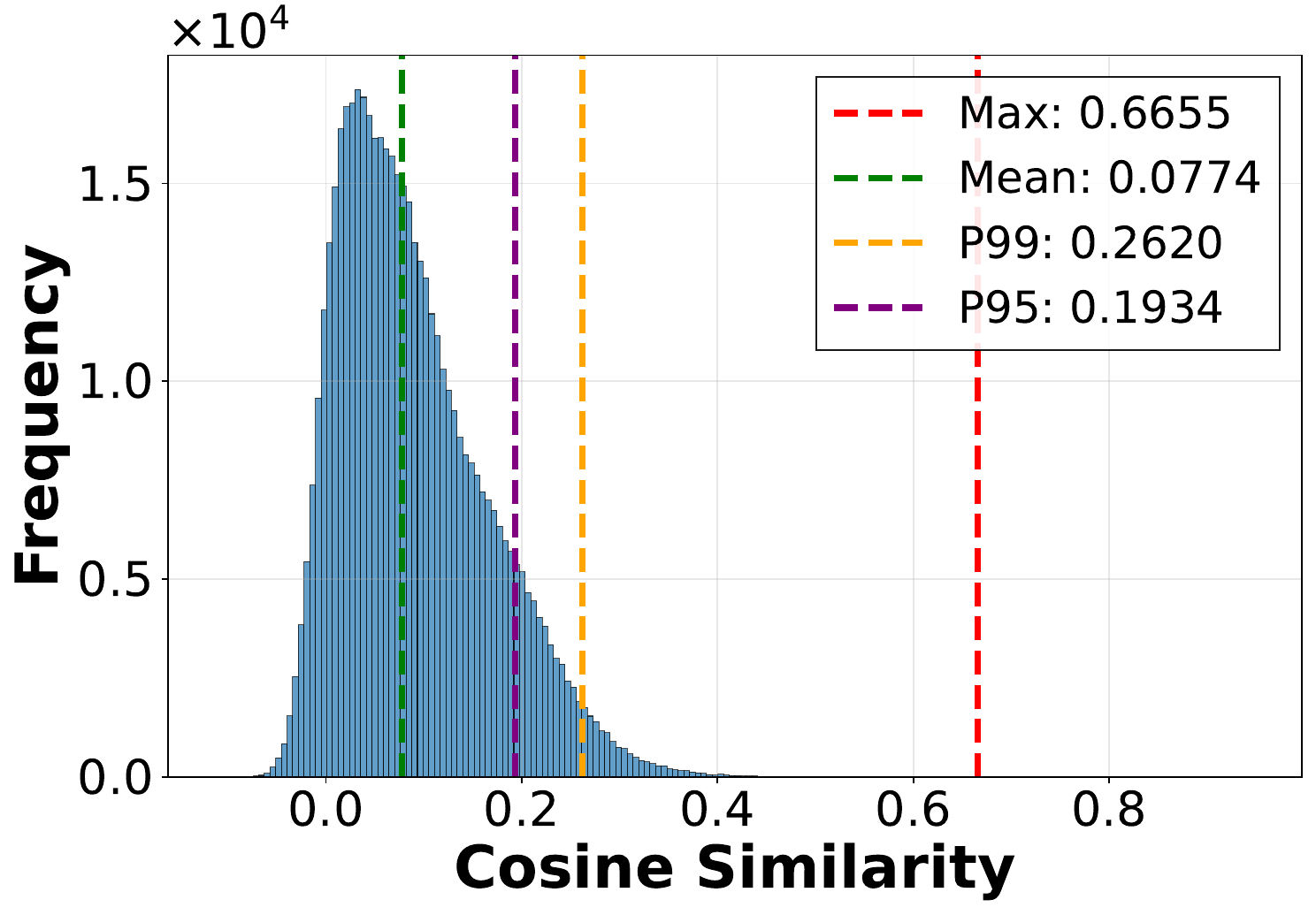}
        \caption{CodeF,Dolmino}
    \end{subfigure}
    \hfill
    \begin{subfigure}{0.19\textwidth}
        \centering
        \includegraphics[width=\linewidth]{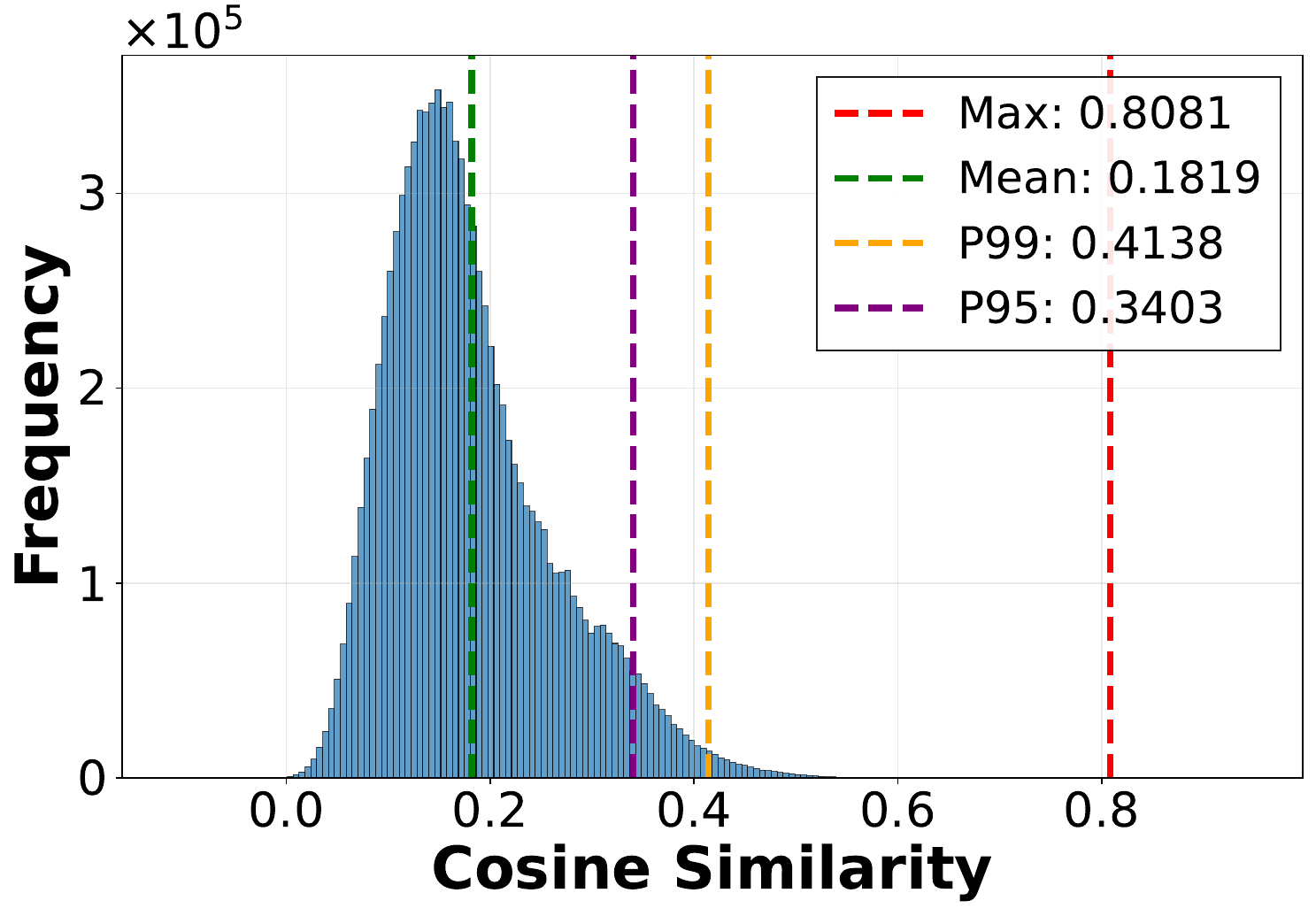}
        \caption{CodeF,DolciSFT}
    \end{subfigure}
    \hfill
    \begin{subfigure}{0.19\textwidth}
        \centering
        \includegraphics[width=\linewidth]{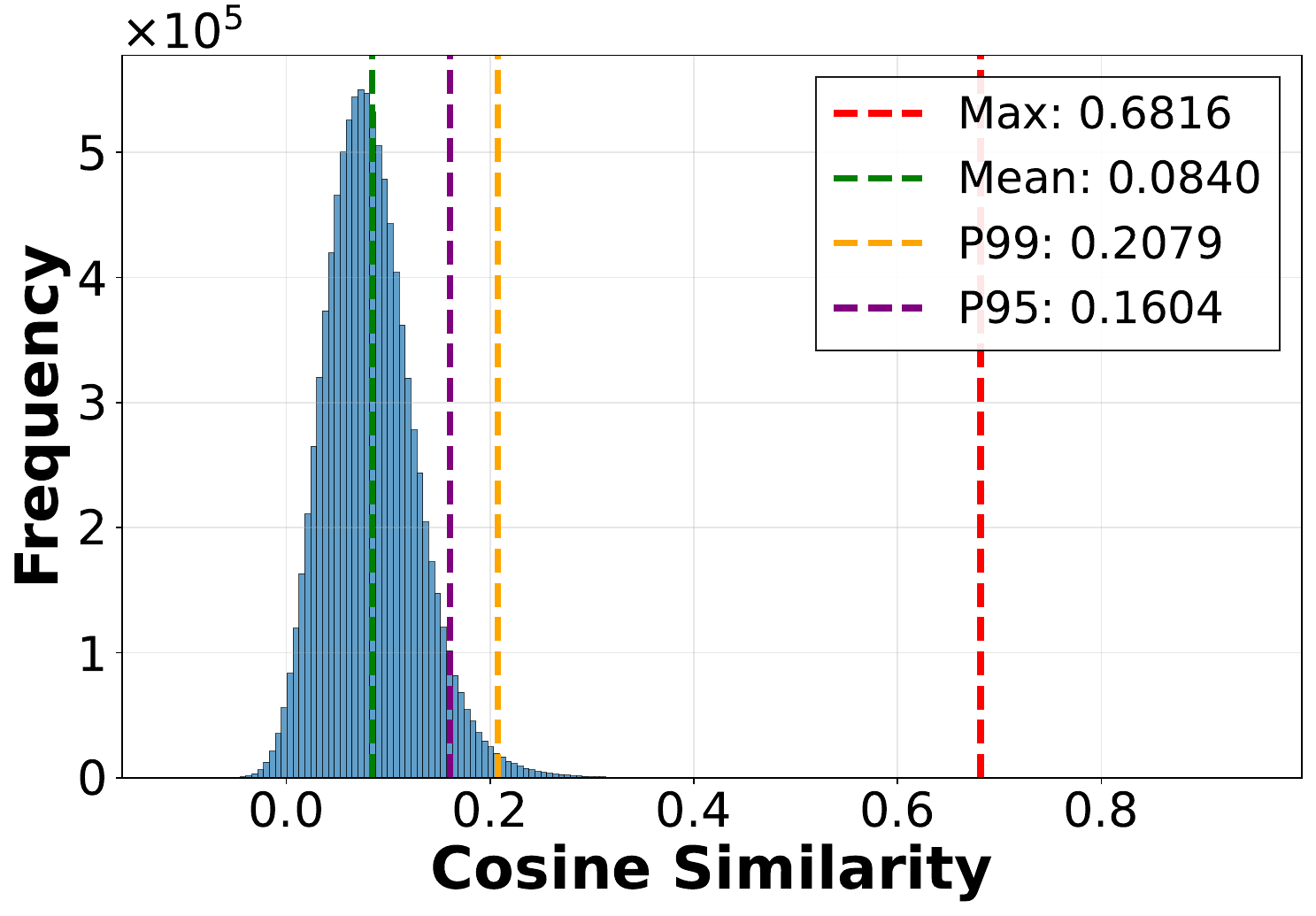}
        \caption{CodeF,DolciDPO}
    \end{subfigure}
    \hfill
    \begin{subfigure}{0.19\textwidth}
        \centering
        \includegraphics[width=\linewidth]{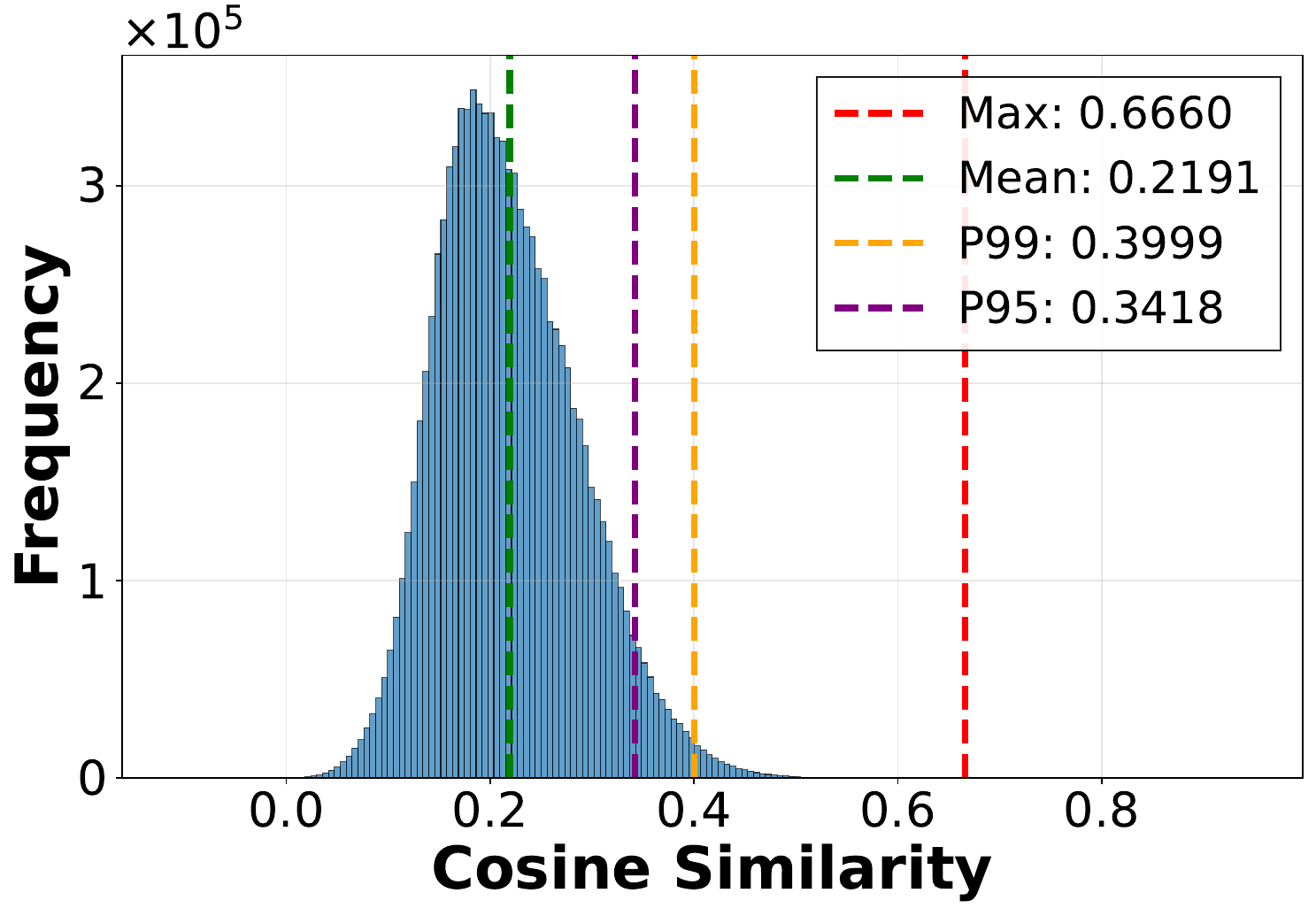}
        \caption{CodeF,DolciRL}
    \end{subfigure}

    \vspace{0.5em}

    \caption{Cosine similarity distributions between benchmark data and training corpus data. Rows from top to bottom: MuSR, ZebraLogic, MBPP, CodeForces. Columns from left to right: Dolma, Dolmino, Dolci SFT, Dolci DPO, and Dolci RL.}
    \label{fig:cos-sim-main-all-plots}
\end{figure*}

Figure~\ref{fig:cos-sim-main-all-plots} shows that high cosine similarity matches are sparse.
We also see in Figure \ref{fig:corr-cos-sim-vs-sem-dupe} that a substantial portion of the very high (over 0.8) cosine similarity matches is a semantic duplicate.

\subsection{Statistical Tests}

In reference to Table~\ref{tab:big_deltas}, we report standard errors on accuracy as $\text{SE}(\hat{p}) = \sqrt{\hat{p}(1-\hat{p})/n}$, the standard error of a binomial proportion. With a minimum evaluation size of $n = 125$, this estimate is well-behaved away from the boundaries $p \in \{0, 1\}$, which is the regime of all reported accuracies.

For Figure~\ref{fig:accuracy-curves} and Table~\ref{tab:contamination_summary} we report Wilcoxon signed-rank test statistics and p-values. The test is paired by sample id: each evaluation problem appears under both the clean and contaminated-trained models, so pairing absorbs the substantial between-problem variance (some problems are near-uniformly easy or hard for every variant) and isolates the contamination effect. We use the rank-based test rather than a paired t-test because the per-problem pass-rates are bounded and concentrate near 0\% and 100\%, so the distribution of paired differences is heavy-tailed and decidedly non-Gaussian; Wilcoxon makes no parametric assumption on the difference distribution and is robust to the resulting outliers. It is the simplest test under these conditions that yields valid inference.

\subsection{Finetuning benchmarks}
\label{app:benchmarks}

We select benchmarks based on: 1) ability to generate new synthetic samples from the same distribution, 2) tractability of creating semantic duplicates by modifying the original samples, 3) the benchmark's general prominence, and 4) benchmarks on which Olmo3 is not saturated, to enable us to track any performance improvement from finetuning. The following are the \textit{only} benchmarks we tested:

\textbf{MBPP -- \citep{DBLP:journals/corr/abs-2108-07732}}. 
$n=427$ programming questions with validated solutions in Python and test cases to check correctness. 
Since they are solvable by entry-level programmers, it is simple to generate correct alternative python solutions or translations in other programming languages.
We expected that the training corpora would contain semantic duplicates of MBPP test data. Unlike for other benchmarks, \textit{LLM-assisted} annotation to detect such duplicates may be feasible without requiring very complex reasoning, but would still be undetectable by typical deduplication methods.
We use the sanitized test set which contains 257 tasks.

\textbf{CodeForces -- \citep{penedo2025CodeForces}}. 
Competitive programming tasks with larger text inputs and problem context than the average code benchmark, such as MBPP. We expected equivalent tasks with less context to appear in training corpora which would be difficult to identify without manual (or LLM) annotation.
We use the 468 problems in the default test set in our experiments.

\textbf{MuSR -- \citep{DBLP:conf/iclr/SpragueYBCD24}}. 
Long narratives generated from a ground-truth logic tree. These trees are provided for each original sample by the authors, which allows the creation of semantic duplicates by regenerating story context from the same tree, and slightly different narratives by modifying non-critical tree branches. We use `murder mysteries', $n=250$ problems.

\textbf{ZebraLogic -- \citep{DBLP:conf/icml/Lin00SPC025}}. 
$n=1000$ logic grid puzzles of varying complexity levels used to evaluate logical reasoning capabilities.
We chose ZebraLogic because Olmo 3 Instruct is not saturated on it (having 32.9\% accuracy).
Additionally, as it is a non-coding benchmark it increases the variety of our suite of benchmarks.

\subsection{Synthetic Data Generation}\label{app:synthetic-data}

\subsubsection{Overview}

\textbf{MBPP}\quad

We generate 5 semantic duplicates for both inputs (questions), and outputs (responses).
For the text inputs, we generate paraphrasings simultaneously to maximize difference between the texts.
For code outputs, we generate alternative python implementations and validate them against provided test cases. 


\textbf{MuSR}\quad
We adapt the original MuSR sample generation logic to generate new samples from existing reasoning trees of the public dataset problems. For the Murder Mysteries and Team Allocation tasks, we create three levels of semantic duplicates differing by how much the underlying logic trees differ from the original while maintaining the answer and the necessary reasoning to solve the problem the same: \textit{regen}) tree is kept the same and story context is regenerated; \textit{$1$-branch}) One branch that does not affect problem outcome or its complexity is changed; \textit{$n$-branches}) All branches of the previous type are changed.

For each level, we generate 2 semantic duplicates for each of the 250 original samples, adding up to 1500 semantic duplicates per task.
We also generate a new test set of 250 samples using the original code.

\textbf{ZebraLogic}\quad
To generate semantic duplicates of ZebraLogic datapoints we use a pipeline of transformations, most of which are LLM-based. 

We apply the following transformation methods:
1) use category mappings that e.g. substitute ``color" by ``shape" and ``red" by ``square'', using Claude 4.5 Haiku \citep{anthropic2025claudehaiku45};
2) shuffle conditions or clues in the prompt using a Python function; and
3) paraphrasing text while preserving all values exactly, using Claude 4.5 Sonnet \cite{anthropic2025claudesonnet45}.
These strategies are applied to the first 500 samples in the benchmark, generating semantic duplicates for paraphrasing alone, and the following combinations of the above: 1) and 2); 2) and 3); and 1), 2) and 3).

\subsubsection{Details of Finetuning on Duplicates}\label{app:finetuning-on-dupes}



\begin{table}[h]
\centering
\caption{Finetuning hyperparameters for each benchmark. All experiments use LoRA Rank 16, Alpha 32, and Dropout 0.05.}
\label{tab:finetuning-hyperparams}
\begin{tabular}{lccc}
\toprule
\textbf{Hyperparameter} & \textbf{MBPP} & \textbf{MuSR} & \textbf{ZebraLogic} \\
\midrule
Learning rate & $1.5e{-}4$         & $2e{-}4$   & $2e{-}4$ \\
KL penalty    & $0.02$     & --         & -- \\
Epochs        & 10         & 6          & 10 \\
\bottomrule
\end{tabular}
\end{table}

\subsubsection{Choice of LoRA as the finetuning method}
\label{app:lora-defense}
We employ LoRA for computational tractability, which introduces a low-rank bottleneck. A potential critique is that this bottleneck restricts the model from learning general reasoning, forcing it to rely on shallow templates. However, recent analyses demonstrate that LoRA at sufficient ranks can closely mirror the performance of full finetuning on reasoning tasks \citep{biderman2024lora}.

Crucially, any capacity limits induced by LoRA apply uniformly across all of our experimental treatment conditions (clean, exact, semantic, and cosine-similar). Because our primary finding of shallow generalization is defined by the \textit{comparative difference} in performance across seen, unseen, and out-of-distribution control splits within the same treatment condition, our comparative claims regarding data leakage and inflation remain structurally robust regardless of LoRA's absolute capacity limits.

\subsubsection{MuSR}\label{app:musr-data}
The generation pipeline uses the same few-shot examples and prompt templates as the original benchmark generation set up. The following model parameters are used with \texttt{gpt-4-0613}:
\begin{itemize}
    \item Temperature: 1.0
    \item Top P: 1.0
    \item Max tokens: 2400
\end{itemize}

Specific details for each MuSR task:

\textbf{Murder Mysteries}\quad
In this task, the model needs to figure out which of the two suspects is the murderer based on a long narrative generated from a logic tree. We make the following modifications to the logic trees before the story generation step:
\begin{itemize}
    \item \textit{regen}: Story context regenerated from unchanged original tree.
    \item \textit{$1$-branch}: One suspicious fact branches changed.
    \item \textit{$n$-branches}: All branches (suspicious, means, motive, opportunity) belonging to the innocent suspect changed.
\end{itemize}

\textbf{Team Allocation}\quad
In this task, the model needs to determine how to best allocate the 3 individuals mentioned in the long narrative to perform two tasks.
We make the following modifications to the logic trees before the story generation step:
\begin{itemize}
    \item \textit{regen}: Story context regenerated from unchanged original tree.
    \item \textit{$1$-branch}: One randomly selected skill branch is changed while keeping skill level unchanged.
    \item \textit{$n$-branches}: All branches swapped while skill and cooperation levels unchanged
\end{itemize}

\textbf{Object Placement}\quad
The data available for the original samples is not enough to generate semantic duplicates. While it is possible to attempt to extract most of the required data with LLMs to attempt sample regeneration, format and style is lost, affecting subsequent generation steps in the multi-stage process. So we choose to omit this MuSR category due to difficulty in creating semantic duplicates of similar complexity and rigor.

\subsubsection{MBPP}\label{app:mbpp-data}
We used \texttt{claude-opus-4-5-20251101} for all generation tasks with parameters:
\begin{itemize}
    \item \texttt{max tokens}: $1024$
    \item \texttt{temperature}: 1.0
\end{itemize}
We generate semantic duplicates for the full MBPP sanitized dataset, that is, 427 tasks. In the process, we find that the following tasks contain bugs (either in the function, or in the test cases): $229$, $438$, $461$, $579$, $769$, $802$.

\textbf{Inputs}\quad We batch generate \textit{paraphrasings} to ensure clear differences between duplicates. 

\textbf{Outputs}\quad We generate \textit{Python semantic duplicates} for each sample sequentially, allowing the model to see the original and previously generated ones to increase the uniqueness of new solutions. We validate solutions at each step and ensure $<0.85$ \texttt{difflib.SequenceMatcher} similarity score between solutions.


\textbf{Prompts}\quad

Prompt for paraphrased question text duplicates:
\begin{lstlisting}[language=Python]
PARAPHRASE_BATCH_PROMPT = """You are an expert at paraphrasing programming task descriptions.

ORIGINAL TASK:
{text}

YOUR TASK:
Generate exactly 5 DISTINCT paraphrases of this programming task. Each paraphrase must:
1. Have COMPLETELY DIFFERENT wording from the others
2. Preserve the EXACT same meaning and requirements
3. Maintain the same level of technical detail and clarity
4. Keep any mentioned function names UNCHANGED

CRITICAL: Each paraphrase must be noticeably different from the others. Vary:
- Sentence structure (active vs passive, questions vs statements)
- Vocabulary choices (synonyms, different technical terms)
- Order of information presented
- Level of formality

AVOID:
- Starting multiple paraphrases the same way (e.g., don't start 3 with "Write a...")
- Simply swapping one or two words while keeping structure identical
- Adding or removing requirements not in the original
- Changing the programming language if one is specified
- Making the task ambiguous or less precise

Output EXACTLY in this JSON format (no extra text, no markdown):
{{
  "para1": "first paraphrase here",
  "para2": "second paraphrase here",
  "para3": "third paraphrase here",
  "para4": "fourth paraphrase here",
  "para5": "fifth paraphrase here"
}}

Generate the 5 diverse paraphrases now:"""
\end{lstlisting}

Prompts for alternative Python solution implementations:
\begin{lstlisting}[language=Python]
# Python semantic duplicate generation
def get_generation_prompt(
    task_description: str,
    original_code: str,
    test_list: list[str],
    previous_solutions: list[str],
    previous_error: Optional[str] = None,
    require_more_different: bool = False
) -> str:
    """Generate the prompt for creating a Python semantic duplicate."""
    
    test_context = "\n".join(test_list[:3])
    func_name = extract_function_name(original_code)
    
    base_prompt = f"""You are an expert Python programmer. Your task is to write a DIFFERENT Python solution for the following problem.

TASK DESCRIPTION:
{task_description}

ORIGINAL PYTHON SOLUTION (for reference - DO NOT COPY):
```python
{original_code}
```

PYTHON TEST EXAMPLES (your solution must pass these):
```python
{test_context}
```

"""
    
    # Add previous solutions if any
    if previous_solutions:
        base_prompt += "PREVIOUS SOLUTIONS YOU'VE ALREADY WRITTEN (your new solution must be STRUCTURALLY DIFFERENT from ALL of these):\n"
        for i, sol in enumerate(previous_solutions, 1):
            base_prompt += f"\n--- Solution {i} ---\n```python\n{sol}\n```\n"
        base_prompt += "\n"
    
    # Add error feedback if retry
    if previous_error:
        base_prompt += f"""YOUR PREVIOUS ATTEMPT HAD ERRORS:
{previous_error}

Please fix these errors in your new solution.

"""
    
    # Add stronger differentiation request if needed
    if require_more_different:
        base_prompt += """IMPORTANT: Your previous solution was TOO SIMILAR to existing ones!
You MUST use a significantly DIFFERENT algorithmic approach. Consider:
- Using different data structures (list vs set vs dict vs deque)
- Using different iteration patterns (for vs while vs recursion vs comprehensions)
- Using different built-in functions or libraries
- Restructuring the logic flow completely

"""
    
    base_prompt += f"""REQUIREMENTS:
1. Write a COMPLETE Python solution that passes all tests
2. The function MUST be named EXACTLY: {func_name}
3. Use a DIFFERENT algorithmic approach or implementation style than the solutions shown above
4. ADD A COMMENT ON THE LINE ABOVE EACH LINE OF CODE explaining what it does
5. Comments should be ORIGINAL, CONCISE, and INSIGHTFUL - not generic
6. Make sure every substantive line has a comment above it
7. The comments should help distinguish this solution semantically from others

COMMENT STYLE EXAMPLE:
```python
# Initialize counter for tracking element frequency
count = 0
# Iterate through each item in the input sequence
for item in items:
    # Increment counter when condition is met
    if condition:
        count += 1
# Return the final accumulated count
return count
```

OUTPUT ONLY THE PYTHON CODE with comments. No markdown code blocks, no explanations outside the code."""

    return base_prompt
\end{lstlisting}

\subsubsection{ZebraLogic}\label{app:zebra-data}
Below we discuss the different methods and transformations:

\textbf{Paraphrasing}. With \texttt{claude-4.5-sonnet} we use the following prompts using the original sample
\begin{lstlisting}
# SYSTEM PROMPT
You are an expert editor tasked with rewriting logic grid puzzles while exactly preserving the logical structure and semantics.

# USER PROMPT
Rewrite the following logic puzzle to express the exact same conditions in different words or with different word order etc. while exactly preserving the logical structure and semantics.

Original Puzzle:
{puzzle}

REQUIREMENTS:
1. Reformulate both the task description and every numbered condition.
2. You may change word order, use synonyms, and alter sentence structure.
3. PRESERVE the strict logical meaning. For example, "A is next to B" must remain logically equivalent (e.g., "B is adjacent to A").
4. PRESERVE all entity names, values, numbers, and categories EXACTLY. Do not change "Red" to "Crimson" or "John" to "Jon". The specific terms used for the puzzle items MUST remain identical to match the solution exactly.
5. The output must be natural, clear, and readable. Avoid contrived or unnatural constructions.
6. Maintain the formatting of the puzzle, including the format and numbering of the list of clues.
7. Do not start your response with a header or a preamble. Start with a naturally flowing puzzle statement in a very similar style and format as the original.

Output ONLY the rewritten puzzle text.
\end{lstlisting}

\textbf{Category substitution}. With \texttt{claude-4.5-haiku} we follow a two step process: (1) generate a substitution plan for each puzzle; (2) apply the substitution with LLMs for improved text cohesiveness; (3) transform the solution programmatically.
\begin{lstlisting}
# STEP 1: SYSTEM PROMPT
You are a helpful assistant that creates substitution plans for logic puzzles.
Your goal is to transform the puzzle by changing BOTH the categories and their values to new domains.

# STEP 1: USER PROMPT
Create a substitution plan to transform this logic grid puzzle.
1. Identify all categories (e.g., Color, Drink, Pet).
2. Assign a NEW category to each (e.g., Color -> Shape, Drink -> Snack, Pet -> Book).
3. Map every existing value to a new value appropriate for the new category.

Original Puzzle:
{puzzle}

Original Solution:
{solution_json}

REQUIREMENTS:
1. Change the categories to natural, distinct alternatives (e.g., colors -> shapes, flowers -> animals).
2. Keep the new categories and values DISTINCT from all of the original ones. Avoid number categories (to avoid confusion with the numbering of the puzzle).
3. Ensure 1-to-1 mapping for all values.
4. Do NOT use obscure or unusual categories. Stick to common categories like colors, animals, shapes, countries, etc. Choose natural categories and values within the flow of the puzzle wording.

Output ONLY a JSON object with this structure:
{
  "substitution_plan": {
    "OriginalCategoryName": {
      "new_category": "NewCategoryName",
      "values": {
        "OldValue1": "NewValue1",
        "OldValue2": "NewValue2"
      }
    },
    ...
  }
}

# STEP 2: SYSTEM PROMPT
You are a helpful assistant that rewrites logic puzzles based on a substitution plan.
You must replace categories and values exactly according to the plan while PRESERVING the puzzle structure, logic, and clues EXACTLY.

# STEP 2: USER PROMPT
Rewrite this logic puzzle by applying the following substitution plan.
Replace ALL occurrences of the old categories and values with their corresponding new ones.

Substitution Plan:
{plan_json}

Original Puzzle:
{puzzle}

CRITICAL INSTRUCTIONS:
1. Replace old categories (e.g., "Color") with new categories (e.g., "Shape").
2. Replace old values (e.g., "Red") with new values (e.g., "Square").
3. Do NOT change the logic, clues, or structure.
4. Keep the puzzle wording identical as much as possible, only make minor syntactic adjustments where necessary to preserve the flow and meaning of the puzzle wording.
5. Keep the numbering and formatting identical.
6. Output ONLY the rewritten puzzle text.
\end{lstlisting}

\textbf{Shuffling}. Puzzles clues are parsed, randomly reordered, and renumbered sequentially. The solution of the resulting semantically equivalent puzzle remains the same.

\textbf{Composite transformations}. With respect to the order of transformations in composite methods, shuffling is always performed first, and paraphrasing is performed last.

\subsubsection{Similarity distributions of synthetic semantic duplicates}
\begin{figure}[ht]
  \centering
  \begin{subfigure}[b]{0.48\columnwidth}
    \includegraphics[width=\linewidth]{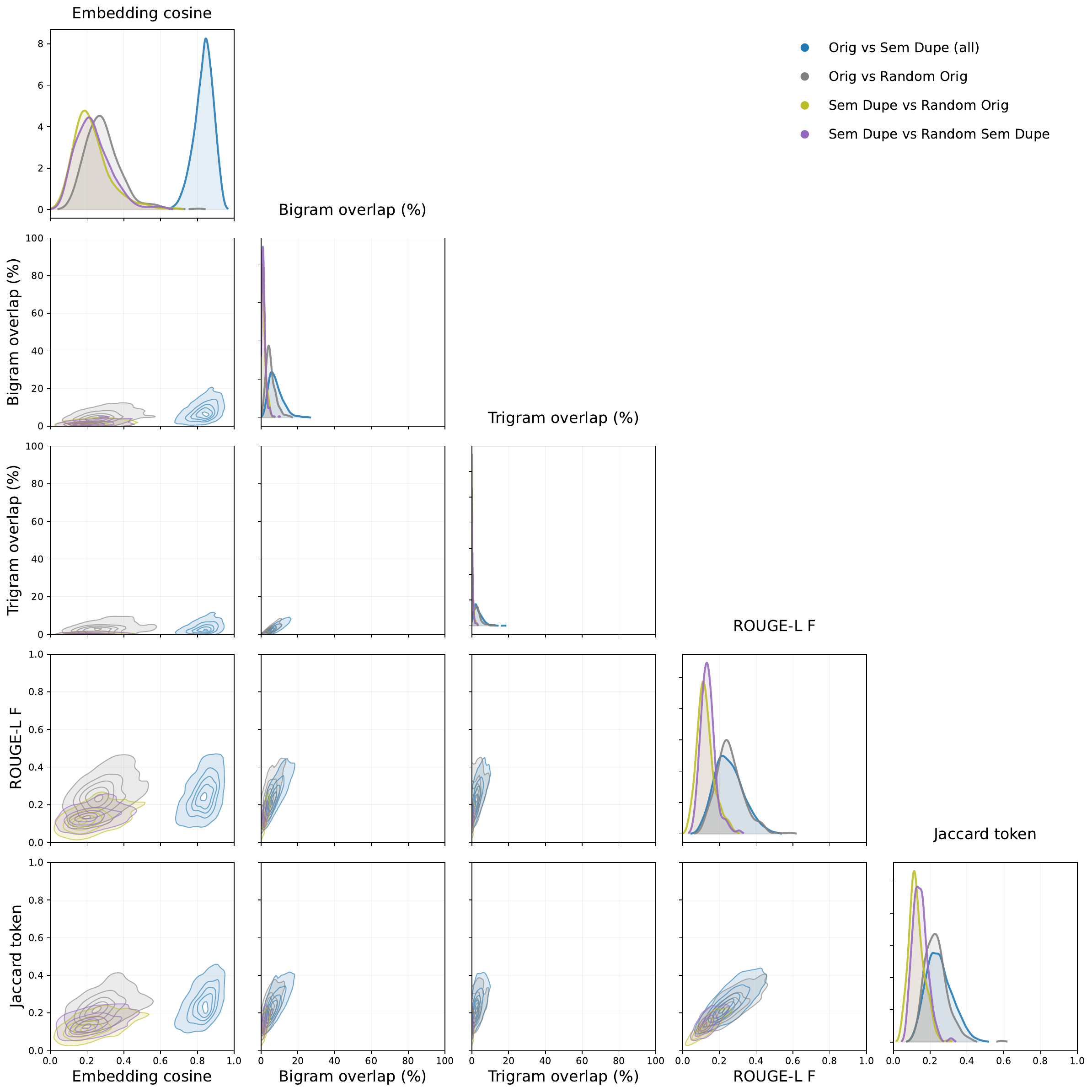}
    \caption{MBPP sanitized test set ($n=257$).}
    \label{fig:mbpp-distributions}
  \end{subfigure}
  \hfill
  \begin{subfigure}[b]{0.48\columnwidth}
    \includegraphics[width=\linewidth]{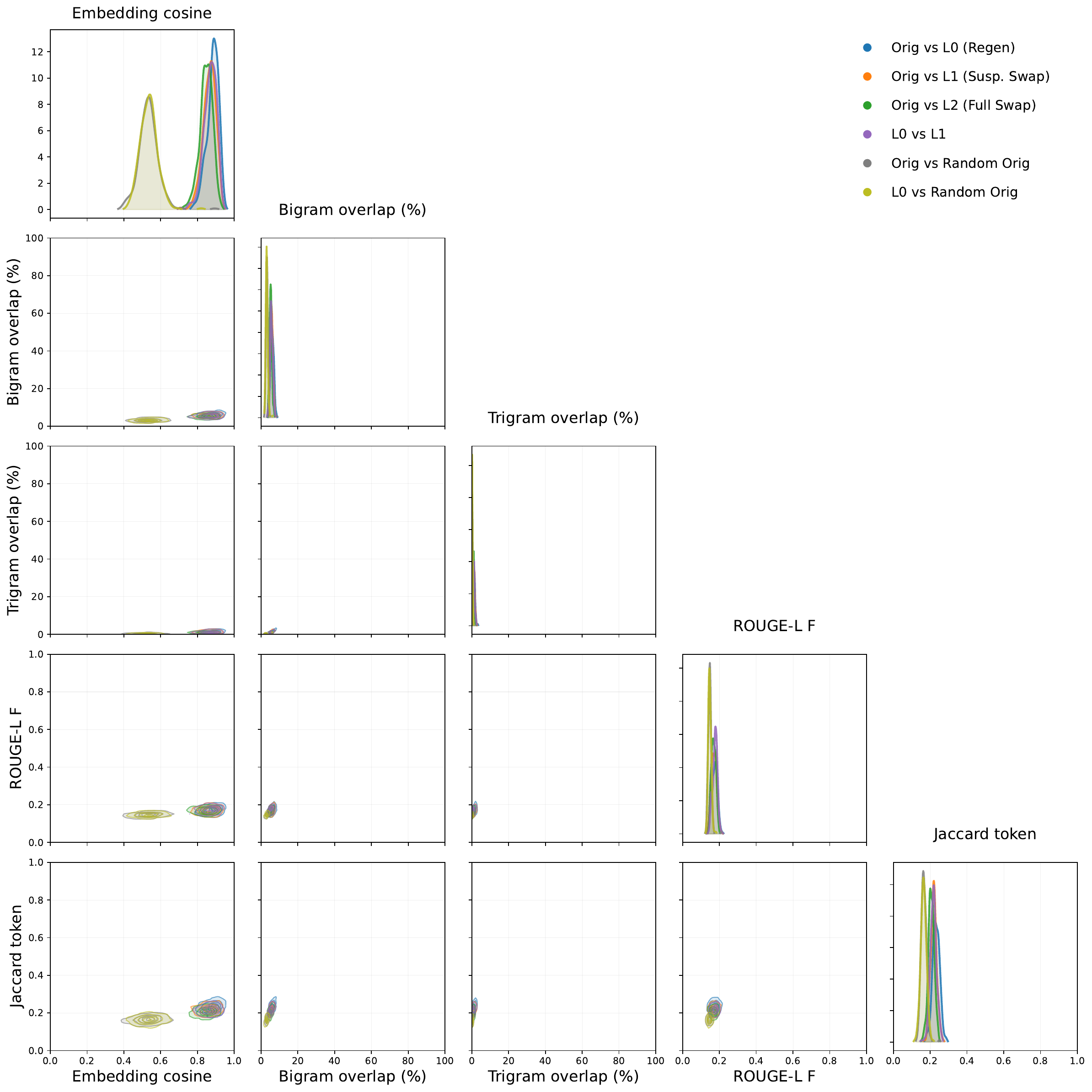}
    \caption{MuSR Murder Mystery task split ($n=250$).}
    \label{fig:musr-distributions}
  \end{subfigure}

  \vspace{0.5em}

  \begin{subfigure}[b]{0.48\columnwidth}
    \centering
    \includegraphics[width=\linewidth]{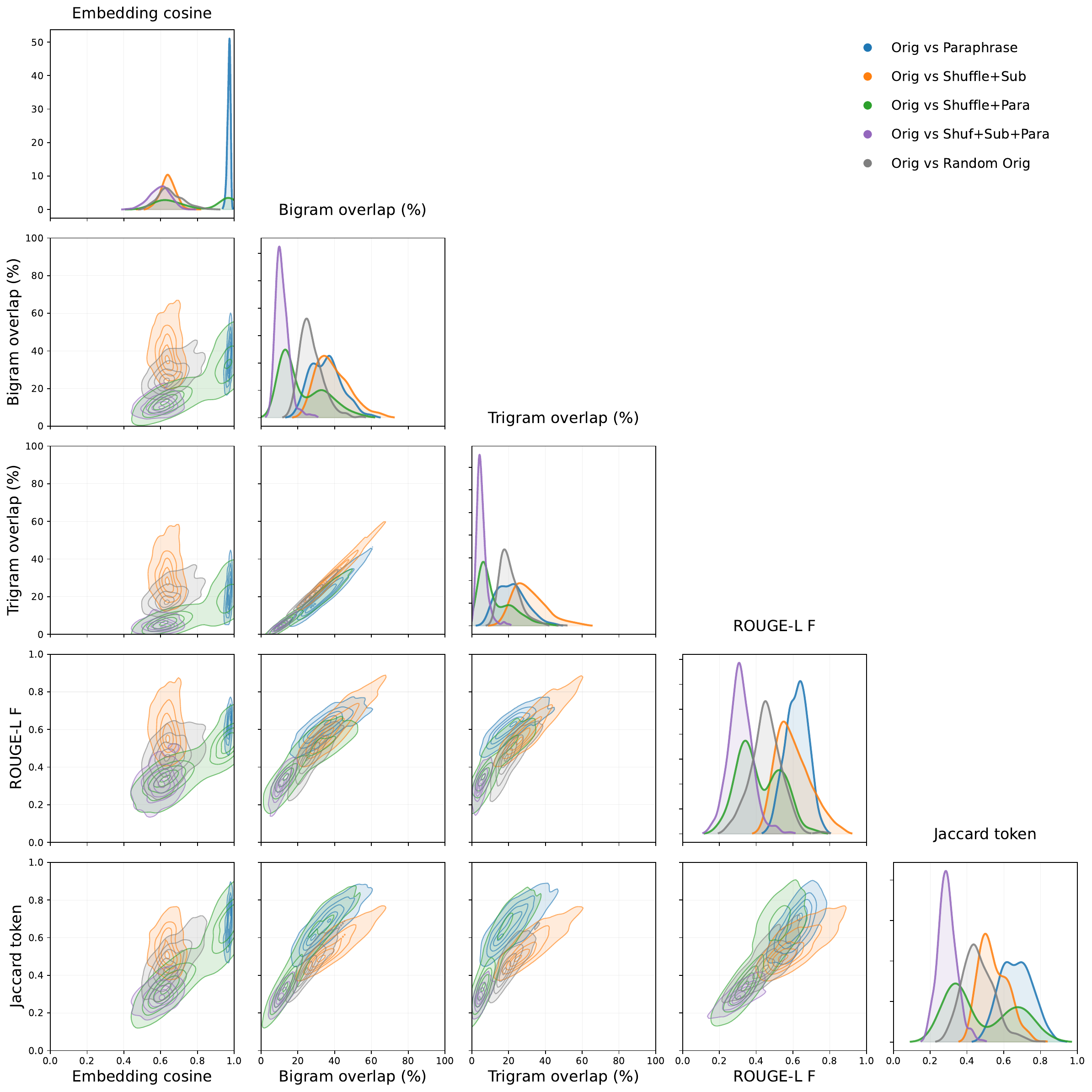}
    \caption{ZebraLogic dataset ($n=1000$).}
    \label{fig:zebra-distributions}
  \end{subfigure}

  \caption{Similarity distributions using several deduplication metrics for each benchmark.}
  \label{fig:benchmark-dist}
\end{figure}

We compare the generated semantic duplicates against the original samples for MBPP, MuSR and CodeForces, and show the analysis for Cosine similarity and several common metrics used in deduplication: n-gram overlap (2 and 3 grams), ROUGE-L F, and Jaccard token. See Figure~\ref{fig:benchmark-dist}.

Finding that in most cases, cosine similarity is better at separating semantic duplicate pairs than the other metrics. This motivates, in part, the use of embedding similarity for the other experiments.




\subsection{Annotation schemes for high cosine similarity matches}\label{app:annot}

We use \texttt{gemini-3-flash-preview} with the following parameters:
\begin{itemize}
    \item \texttt{thinking\_level}: \texttt{MEDIUM} (we use \texttt{HIGH} for CodeForces due to the length of the problems.)
    \item \texttt{temperature}: $1.0$
    \item \texttt{max\_output\_tokens}: $8192$
    \item \texttt{response\_format}: JSON, with structured output conforming to the schema below.
\end{itemize}

For each annotated pair, we use a schema to collect the following:
\begin{itemize}
    \item \texttt{is\_sd}: A boolean indicating whether the pair constitutes a semantic duplicate (true if the corpus task is the same as or subsumes the test task.)
    \item \texttt{confidence}: A confidence score in $[0, 1]$, where $1.0$ indicates certainty and $0.0$ represents a 50-50 guess.
    \item \texttt{reasoning}: A free-text explanation of the judgment, including key similarities and differences observed.
    \item \texttt{match\_type}: A categorical label describing the relationship: \textit{exact} (nearly identical), \textit{equivalent} (same task, different wording), \textit{subset} (test is a subset of corpus), \textit{superset} (corpus is a subset of test), or \textit{unrelated} (different tasks.)
\end{itemize}

The annotations are then used to finetune \texttt{gpt-oss-20b} \citep{agarwal2025gpt} as a classifier for CodeForces semantic duplicate detection, and \texttt{Qwen/Qwen3-30B-A3B-Instruct-2507} for MBPP. 

\subsubsection{Classifier finetuning results}

\textbf{MBPP.}\quad We finetune \texttt{Qwen/Qwen3-30B-A3B-Instruct-2507} with LoRA (rank 32) on $13266$ samples ($9\%$ semantic duplicates), and achieve $96.8\%$ accuracy on a held-out validation set (n=$560$, $25\%$ semantic duplicates) with $97.7\%$ precision and $93.3\%$ F1 on the semantic duplicate class.

\textbf{CodeForces.}\quad We also finetune \texttt{gpt-oss-20b} with LoRA (rank 32) on $13398$ samples ($9\%$ semantic duplicates), and achieve $95.3\%$ accuracy on the held-out validation set (n=$512$, $25\%$ semantic duplicates) with $97\%$ precision and $90\%$ F1 on the semantic duplicate class.

In Table~\ref{tab:eco-classifier-performance} we show all details of the finetuning set up for the two classifiers, and the final metrics that validate their performance.

\begin{table}[t]
\centering
\caption{Summary of classifier performance}
\label{tab:eco-classifier-performance}
\begin{tabular}{l cc}
\toprule
\textbf{Feature} & \textbf{Codeforces} & \textbf{MBPP} \\
 & \textbf{(GPT-OSS-20B)} & \textbf{(Qwen3-30B-A3B-Instruct)} \\
\midrule
\multicolumn{3}{l}{\textit{Training Configuration}} \\
\cmidrule(l){1-3}
Base Model          & openai/gpt-oss-20b & Qwen3-30B-A3B-Instruct \\
LoRA Rank           & 32        & 32        \\
Epochs              & 10        & 5         \\
Batch Size          & 128       & 128       \\
Learning Rate       & 0.0001    & 0.0005    \\
Max Seq Length      & 4096      & 4096      \\
\midrule
\multicolumn{3}{l}{\textit{Dataset Configuration}} \\
\cmidrule(l){1-3}
Train Size          & 13{,}398  & 13{,}266  \\
Train Duplicates    & 1{,}218   & 1{,}206   \\
Eval Size           & 512       & 560       \\
Eval Duplicates     & 128       & 140       \\
\midrule
\multicolumn{3}{l}{\textit{Final Metrics (Eval Set)}} \\
\cmidrule(l){1-3}
Accuracy            & 95.31\%   & 96.79\%   \\
Balanced Accuracy   & 91.41\%   & 94.29\%   \\
Macro F1            & 0.9343    & 0.9559    \\
Duplicate Precision & 97.27\%   & 97.66\%   \\
Duplicate Recall    & 83.59\%   & 89.29\%   \\
Duplicate F1        & 0.8992    & 0.9328    \\
Non-Dup Precision   & 94.78\%   & 96.53\%   \\
Non-Dup Recall      & 99.22\%   & 99.29\%   \\
TP / FN / FP / TN   & 107 / 21 / 3 / 381 & 125 / 15 / 3 / 417 \\
\bottomrule
\end{tabular}
\end{table}

\subsubsection{Annotation prompt for MBPP matches}

\begin{lstlisting}
MBPP_PROMPT_TEMPLATE = """You are an expert programmer analyzing potential semantic duplicates between coding tasks.

## Task
Determine if the following two coding tasks are semantic duplicates - meaning they describe the same programming task, just potentially phrased differently.

## Test Task (from benchmark):
{test_text}

## Corpus Task (from training data):
{corpus_text}

## Guidelines:
1. **Focus on the TASK, not the solution** - ignore any code or solutions that may be present
2. **Mathematical equivalence counts as duplicate** - e.g., "sum 1 to n" and "sum n, n-1, ..., 1" are equivalent
3. **Corpus subsumes test = duplicate** - if the corpus task is strictly harder (asks for more), but solving it would trivially solve the test task, mark as duplicate
4. **Be calibrated** - use confidence primarily for ambiguous cases, tricky phrasing, or when you're uncertain

## Match Types:
- "exact": Nearly identical wording
- "equivalent": Different phrasing, same underlying task
- "subset": Test task is a subset of corpus task (corpus is harder but solves test)
- "superset": Corpus task is a subset of test task (test is harder) - NOT a duplicate
- "unrelated": Different tasks entirely

Analyze the tasks and provide your structured judgment."""
\end{lstlisting}

\subsubsection{Annotation prompt for CodeForces matches}

\begin{lstlisting}[language=Python]
CODEFORCES_PROMPT_TEMPLATE = """You are an expert competitive programmer analyzing potential semantic duplicates between programming problems.

## Task
Determine if the following two competitive programming problems are semantically related - meaning exposure to the corpus problem during training could help solve the test problem.

## Test Problem (from benchmark):
{test_text}

## Corpus Problem (from training data):
{corpus_text}

## Analysis Steps:
1. **Check data quality first**: Is the corpus text a complete problem statement? If it's empty, fragmentary, or contains only code without a problem description, mark as "unrelated".
2. **Check for exact text match**: If the corpus text appears VERBATIM (word-for-word) within the test text (e.g., corpus contains just the problem statement while test contains problem + examples), this counts as "exact" match.
3. **Extract the core problem**: Strip away story/narrative framing. What is the actual computational task?
4. **Identify the key insight**: What algorithmic technique or observation is needed?
5. **Compare**: Is there meaningful overlap in what's being asked or how to solve it?

## Match Types:
- "exact": Nearly identical problem statements, OR corpus text is a verbatim substring/subsection of test text (exact text match even if corpus is shorter)
- "equivalent": Different framing but identical algorithmic core
- "subset": Test is a special case of corpus (test asks for less than corpus)
- "superset": Corpus asks for something simpler than test, but NOT a verbatim text match
- "related": Corpus covers a component or shares key insight with test
- "unrelated": Different problems, or corpus data is unusable

## IMPORTANT: Exact Match Clarification
If the corpus text is an exact substring of the test text (the corpus text appears word-for-word inside the test text, just without some sections like examples or input/output format), mark this as "exact" NOT "superset". The key distinction:
- "exact": Corpus text IS CONTAINED VERBATIM in test text
- "superset": Corpus asks a DIFFERENT (simpler) question than test

## What counts as semantically related:
- Same computational task (any framing)
- One is a special case of the other
- Shared key insight or trick
- Corpus solves a significant component of test

## What is unrelated:
- Sharing only common techniques (DP, BFS) without structural similarity
- Unusable corpus data (empty, fragmentary, code-only)
- Genuinely different computational questions"""
\end{lstlisting}

\subsubsection{Final grader prompt for selected matches}

\begin{lstlisting}[language=Python]
GRADER_PROMPT_TEMPLATE = """You are an exceptionally talented programmer and mathematician with the goal to determine whether there are semantic duplicates between two chunks of texts.

You will be provided with a benchmark problem and a corpus text problem, as well as a reason that they may or may not be semantic duplicates. The rationale may be wrong.

Please consider how both problems are solved. Your goal here is to determine whether or not it is the case that the corpus problem provides significant algorithmic insight required to solve a major part of the test problem, thus making it a semantic duplicate.

Respond with TRUE if it is a semantic duplicate.
Provide your reason, in detail, as to why it is or is not.
Provide a score on the amount of duplication from 0 to 1.

## Test (benchmark) problem:
{test_text}

## Corpus problem:
{corpus_text}

## Prior reasoning (may be wrong):
{prior_reasoning}"""
\end{lstlisting}

\subsection{Self-consistency results for the annotation system against human validations}
\label{app:human-validation}
In Tables~\ref{tab:consistency-codeforces} and \ref{tab:consistency-mbpp}, we perform a self-consistency analysis on selected annotations which we manually annotate across 5 categories with respect to semantic duplicates:
\begin{itemize}
    \item \textbf{equivalent:} the corpus sample states the same task as the test sample in different words.
    \item \textbf{exact:} the corpus sample is a near-verbatim restatement of the test problem (solutions may differ). This is distinct from the exact duplicates used for finetuning in Section 3.2.
    \item \textbf{subset:} the test sample is a special case or component of the corpus sample, so solving the corpus problem solves the test problem. For MBPP, the corpus program contains a working version of the test program as part of its implementation.
    \item \textbf{superset:} the corpus sample solves only part of the test sample, so it does not supply a full solution.
    \item \textbf{related} (CodeForces only): the samples share a technique or insight, but the corpus does not cover the test problem.
    \item \textbf{unrelated:} no algorithmic overlap; a false match.
\end{itemize}

\begin{table}[t]
\centering
\caption{Self-consistency of the Gemini grader on 23 CodeForces problems per category, run 10 times each. Fleiss's $\kappa = 0.78$ indicates substantial agreement. Equivalent and Exact are frequently confused, but both count as semantic duplicates downstream, so the conflation does not affect our prevalence estimate. The net bias is conservative: 30.4\% of true Subsets leak to Superset, Related, or Unrelated (missed hits), while only 4.3\% of true Supersets leak to Equivalent (false hits). Since true Subsets dominate true Supersets in practice, the matrix biases our reported prevalence downward.}
\label{tab:consistency-codeforces}
\begin{tabular}{l cccccc}
\toprule
 & \multicolumn{6}{c}{\textbf{Predicted}} \\
\cmidrule(lr){2-7}
\textbf{H.\ Annotation} & \textsc{Equiv.} & \textsc{Exact} & \textsc{Subset} & \textsc{Superset} & \textsc{Related} & \textsc{Unrelated} \\
\midrule
Equivalent & \textbf{39.1} & 60.9 &  --  &  --   &  --   &  --   \\
Exact      & 13.0 & \textbf{87.0} &  --  &  --   &  --   &  --   \\
Subset     & 13.0 &  4.3 & \textbf{52.2} &  8.7  & 17.4  &  4.3  \\
Superset   &  4.3 &  --  &  --  & \textbf{52.2} & 39.1  &  4.3  \\
Related    &  --  &  --  &  --  &  --   & \textbf{91.3} &  8.7  \\
Unrelated  &  --  &  --  &  --  &  --   &  --   & \textbf{100.0} \\
\bottomrule

\end{tabular}
\end{table}

\begin{table}[t]
\centering
\caption{Self-consistency of the Gemini grader on MBPP, with 20 duplicates per category labeled 10 times each (Exact has only 1 sample, as it is the only Exact duplicate in the corpus). Fleiss's $\kappa = 0.904$ indicates almost perfect agreement. As with CodeForces, Equivalent and Exact are confounded but both count as semantic duplicates downstream, so the conflation does not affect our prevalence estimate. Subset is occasionally dropped to Unrelated (16.5\%) and Superset to Unrelated (23.5\%), indicating that the matrix biases our reported prevalence downward.}
\label{tab:consistency-mbpp}
\begin{tabular}{l ccccc}
\toprule
 & \multicolumn{5}{c}{\textbf{Predicted}} \\
\cmidrule(lr){2-6}
\textbf{H.\ Annotation} & \textsc{Equiv.} & \textsc{Exact} & \textsc{Subset} & \textsc{Superset} & \textsc{Unrelated} \\
\midrule
Equivalent & \textbf{96.0} &  0.0 &  0.5 &  3.5 &  0.0 \\
Exact      & 99.5 & \textbf{0.5} &  0.0 &  0.0 &  0.0 \\
Subset     &  0.0 &  0.0 & \textbf{83.5} &  0.0 & 16.5 \\
Superset   &  0.0 &  0.0 &  5.5 & \textbf{71.0} & 23.5 \\
Unrelated  &  0.0 &  0.0 &  4.5 &  0.5 & \textbf{95.0} \\
\bottomrule
\end{tabular}
\end{table}

\FloatBarrier
\section{Further Semantic Duplicates in the Wild Results}

\begin{figure}[ht]
  \centering
  \begin{subfigure}[b]{0.48\columnwidth}
    \includegraphics[width=\linewidth]{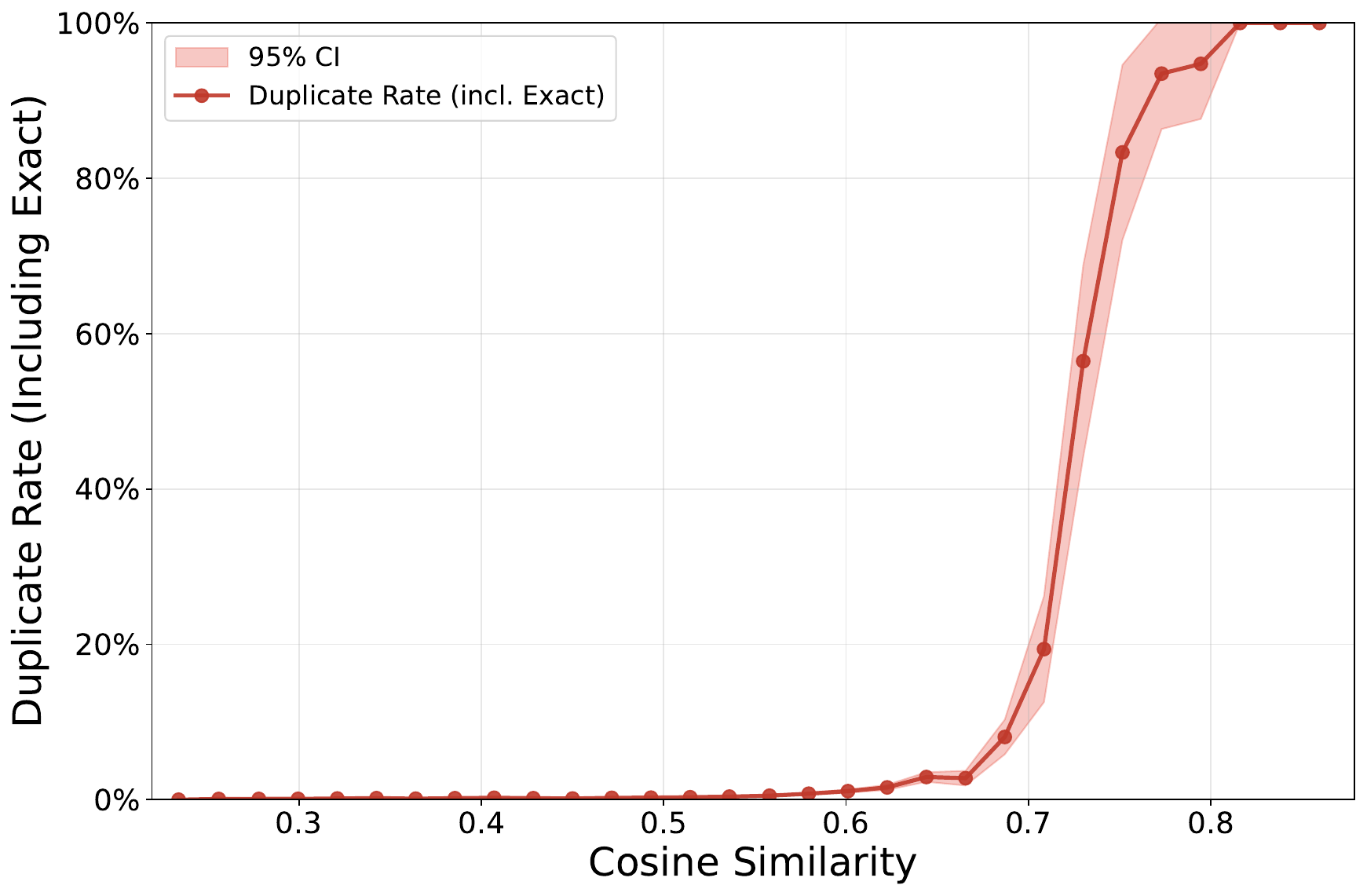}
    \caption{Correlation of Similarity vs. Duplicates (including semantic and exact)}
  \end{subfigure}
  \hfill 
  \begin{subfigure}[b]{0.48\columnwidth}
    \includegraphics[width=\linewidth]{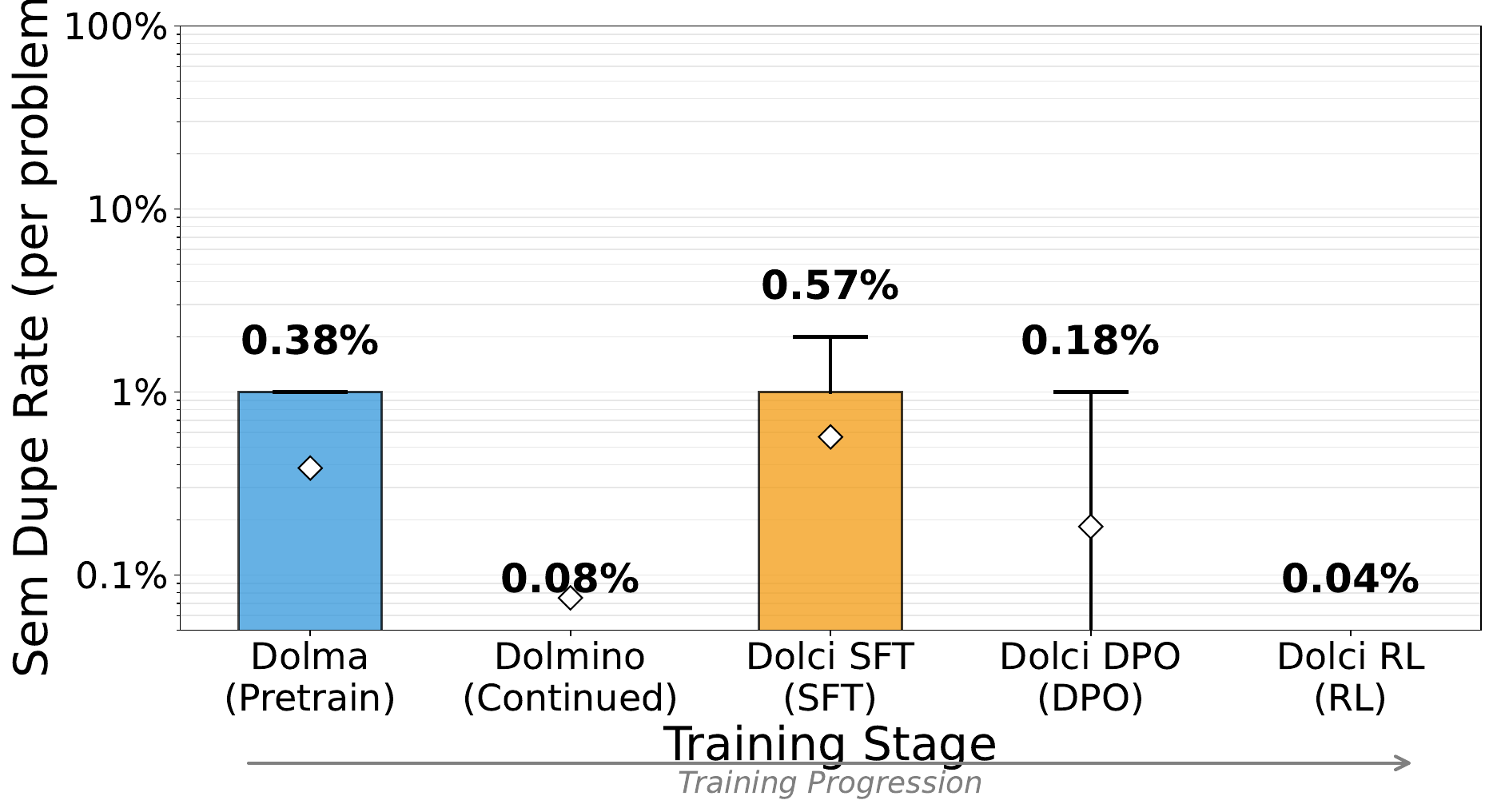}
    \caption{Propensity of semantic duplicates by Training Scheme (excluding exact duplicates)}
  \end{subfigure}
  \caption{\textbf{CodeForces duplicates using the top 100 cosine-similarity matches per problem and training stage} (a) Rate of semantic or exact duplicates as a function of cosine similarity, with 95\% CIs. (b) Mean semantic duplicate rate per problem by training stage, excluding exact duplicates, log scale (Figure~\ref{fig:codeforces-sem-dupe-propensity-training-scheme} reports the share of problems with at least one duplicate).}
  \label{fig:top100-sem-dupe-wild}
\end{figure}


\subsection{Semantic duplicates are hard to detect}\label{app:sem-dupes-hard-to-detect}

Semantic duplicates in the wild are sparse and difficult to find. From the above we notice that semantic duplicates can be found, even for hard CodeForces level problems. We consider, per test point, approximately 350 million texts.  We find in the case of CodeForces on average a few semantic duplicates. For MBPP there are tens to one hundred across our entire dataset. We run our algorithm across around ~250GB of data to find necessary duplicates for a representative population. 

To illustrate this point we demonstrate the probability of a semantic duplicate occurring given a cosine similarity and being in the top 0.1\% of a semantic duplicates test set. 

The relationship between embedding cosine similarity and the probability of semantic duplication, aggregated across all training stages (N=128,408 training-test pairs). Points represent binned similarity scores (30 bins); shaded regions indicate 95\% confidence intervals computed using the normal approximation to the binomial distribution. Sample sizes for each bin are annotated.

Semantic duplicate rate exhibits a nearly monotonically increasing relationship with cosine similarity as we would expect. Below a similarity threshold of approximately 0.35, the duplicate rate is effectively zero (<1\%). The rate increases sharply between 0.4 and 0.7, following an approximately sigmoidal trajectory, and reaches 60–85\% at the highest observed similarities (>0.8). The widening confidence intervals at high similarity values and volatility reflect reduced sample sizes in these bins. This calibration curve suggests that cosine similarity serves as a useful but imperfect proxy for semantic duplication, with a practical decision threshold in the 0.5–0.6 range capturing the inflection point of the relationship.

\subsubsection{Ecologically valid finetuning experiment}\label{app:ecological-results}
\paragraph{Finetuning Hyperparameters}
Hyperparameters mirror the official Olmo 3 hybrid SFT recipe so that the training trajectory is representative of contemporary post-training rather than a setting tuned to amplify contamination:
\begin{table}[htbp]
\centering
\caption{Finetuning Hyperparameters}
\label{tab:finetuning_hyperparameters}
\begin{tabular}{ll}
\toprule
\textbf{Hyperparameter} & \textbf{Value} \\
\midrule
Optimizer & \texttt{adamw\_torch\_fused} \\
Betas & $\beta_1=0.9$, $\beta_2=0.95$ \\
Weight Decay & $0$ \\
Learning Rate Olmo 3 & 2.5e-5 \\
Learning Rate Schedule & Linear decay, $0.03$ warmup ratio \\
Max Grad Norm & 1.0 \\
Precision & \texttt{bf16} parameters, \texttt{fp32} reductions \\
Effective Batch Size & 64 sequences (per-device 24 $\times$ grad accum $\times$ \#GPUs) \\
Max Sequence Length & 2048 \\
Checkpoints & 1 per epoch (10 / model / corpus) \\
Loss Masking & Assistant response only (\texttt{completion\_only\_loss=True}) \\
\bottomrule
\end{tabular}
\end{table}

Loss is masked to the assistant response (\texttt{completion\_only\_loss=True})---prompt tokens contribute no gradient, matching the open-instruct SFT convention. All runs use the closest reproduction to the Olmo hyperparameter set for SFT, as they report.


\subsubsection{Reporting on top 100 cosine similarity matches instead of sampling from the top 0.1\%}\label{app:sem-dupe-wild-results-top100}

Complimenting Figure~\ref{fig:codeforces-sem-dupe-propensity-training-scheme}, see Figure~\ref{fig:top100-sem-dupe-wild} for the top 100 cosine similarity matches for CodeForces instead of 100 sampled from top 0.1\%.

\FloatBarrier
\section{Further Finetuning Results, Including Degradation Analysis}\label{app:finetuning-results}








The Opus 4.5 MuSR baseline accuracy is 91.6\%. The model does slightly worse on the first half of the data (that we train on in Table \ref{tab:app-musr-finetuning-models}) than on the second half, respectively Opus 4.5 gets 90.4\% on the first half and 92.8\% on the second half. 

The performance of \texttt{gpt-4.1-mini-2025-04-14} on MuSR benchmark data is 84.0, on our level0 semantic duplicates it is 79.8 and on level2 it is 76.4.
This is just the baseline performance of the teacher model GPT 4.1 mini without any finetuning.

Tables~\ref{tab:app-finetuning-degradation-musr} and \ref{tab:app-finetuning-degradation-zebra} show that our finetuned models do not degrade in performance across 6 other benchmarks.

\begin{table}[t]
\caption{Effect of finetuning Olmo 3 on semantic duplicates of MuSR Murder Mysteries reasoning traces. Olmo 3 was finetuned for 3 epochs.}
  \label{tab:app-musr-finetuning-models}
  \begin{center}
    \begin{small}
      \begin{sc}
\begin{tabular}{lccc}
\toprule
\makecell{\textbf{Duplication} \\ \textbf{level}} & \makecell{\textbf{Teacher: } \\ \textbf{Opus} $\mathbf{4.5}$}& \makecell{\textbf{Teacher: } \\ \textbf{GPT} $\mathbf{4.1}$ \textbf{mini}} \\
\midrule
Baseline & 66.0 & 66.0\\
\midrule
Exact Dupes  & 87.1 & 82.5\\
Regen & 86.8 &  82.4\\
$1$-branch &  86.1& 81.3 \\
$n$-branches & 85.8& 81.6 \\
\midrule
          \bottomrule
        \end{tabular}
      \end{sc}
    \end{small}
  \end{center}
  \vskip -0.1in
\end{table}

\begin{table}[t]
\caption{No degradation effect of finetuning Olmo 3 on semantic duplicates of MuSR Murder Mysteries reasoning traces. Olmo 3 was finetuned for 3 epochs.}
  \label{tab:app-finetuning-degradation-musr}
  \begin{center}
    \begin{small}
      \begin{sc}
\begin{tabular}{lcccccc}
\toprule
\makecell{\textbf{Duplication} \\ \textbf{level}} & \makecell{\textbf{Arc} \\ \textbf{Challenge}} & \makecell{\textbf{Arc } \\ \textbf{Easy}} & \makecell{\textbf{BoolQ} } & \makecell{\textbf{HellaSwag} } & \makecell{\textbf{Piqa} } & \makecell{\textbf{Winogrande} }\\
\midrule
Baseline & 50.1 & 78.2 & 75.7 & 57.5 & 75.0 & 65.3\\
\midrule
Exact Dupes  & 50.0 & 78.7 & 76.7 & 56.4 & 75.8 & 65.0\\
regen & 50.3 &  78.6 & 76.0 & 56.4 & 75.7 & 64.8\\
$1$-branch & 50.3 &  78.6 & 77.0 & 56.4 & 75.6 & 64.7\\
$n$-branches & 50.6 & 78.9 & 77.1 & 56.5 & 75.6  & 65.1\\
\midrule
          \bottomrule
        \end{tabular}
      \end{sc}
    \end{small}
  \end{center}
  \vskip -0.1in
\end{table}

\begin{table}[t]
\caption{No degradation effect of finetuning Olmo 3 on semantic duplicates of ZebraLogic reasoning traces. Olmo 3 was finetuned for 3 epochs.}
  \label{tab:app-finetuning-degradation-zebra}
  \begin{center}
    \begin{small}
      \begin{sc}
\begin{tabular}{lcccccc}
\toprule
\makecell{\textbf{Duplication} \\ \textbf{level}} & \makecell{\textbf{Arc} \\ \textbf{Challenge}} & \makecell{\textbf{Arc } \\ \textbf{Easy}} & \makecell{\textbf{BoolQ} } & \makecell{\textbf{HellaSwag} } & \makecell{\textbf{Piqa} } & \makecell{\textbf{Winogrande} }\\
\midrule
Baseline & 50.1 & 78.2 & 75.7 & 57.5 & 75.0 & 65.3\\
\midrule
Exact Dupes  & 49.5 & 78.0 & 77.0 & 56.8 & 75.0 & 64.7\\
Para & 49.3 &  77.7 & 78.3 & 56.4 & 74.5 & 64.1\\
Shuffle, Subs & 50.7 &  77.4 & 75.3 & 56.5 & 75.1 & 64.5\\
Shuffle, Para & 49.4 & 78.1 & 77.8 & 56.5 & 75.3  & 64.9\\
Shuffle, Subs, Para & 50.4 & 78.1 & 78.4 & 56.5 & 75.6  & 63.5\\
\midrule
          \bottomrule
        \end{tabular}
      \end{sc}
    \end{small}
  \end{center}
  \vskip -0.1in
\end{table}

\FloatBarrier
\section{Ecologically Finetuned Results}\label{app:eco-fine-results}

Table~\ref{tab:contamination_summary} complements our results in Figure~\ref{fig:accuracy-curves}, showing $p$ values.

\begin{table}[H]
\centering\small
\caption{Results from replicating our finetuning experiments on more models. Means are reported for each finetuning treatment on the \textsc{Seen} split (items the contaminated arm was trained on) and the \textsc{Unseen} held-out clean split. 
Seen corresponds to contaminated test, and unseen to clean test. 
$p$-values are from paired Wilcoxon signed-rank tests (asymptotic, with tie correction) on per-item mean accuracies (averaged over 20 evaluations) per question, comparing  against \textsc{Contaminated}.
$^{*}p<0.05,\ ^{**}p<0.01,\ ^{***}p<0.001$.
}
\label{tab:contamination_summary}
\renewcommand*{\arraystretch}{0.5}

\begin{tabular}{ll l ccc cc}
\toprule
 & & & \multicolumn{3}{c}{\textbf{Accuracy (\%)}} & \multicolumn{2}{c}{\textbf{$p$ vs.\ \textsc{Contam.}}} \\
\cmidrule(lr){4-6} \cmidrule(lr){7-8}
\textbf{Model} & $\mathbf{E}$ & \textbf{Split} & \textsc{Clean} & \textsc{Contam.} & \textsc{Exact} & \textsc{Clean} & \textsc{Exact} \\
\midrule
\multirow{6}{*}{\begin{turn}{90}Olmo-3-7B\end{turn}}
 & \multirow{2}{*}{Baseline} & Seen   & \multicolumn{3}{c}{49.9} & -- & -- \\
 &                           & Unseen & \multicolumn{3}{c}{49.8} & -- & -- \\
\cmidrule(l){2-8}
 & \multirow{2}{*}{5}  & Seen   & 56.3 & 58.1 & 71.1  & $0.689$           & $<\!0.001^{***}$ \\
 &                     & Unseen & 57.4 & 56.5 & 55.5  & $0.852$           & $0.652$ \\
\cmidrule(l){2-8}
 & \multirow{2}{*}{10} & Seen   & 52.5 & 69.4 & 75.5  & $0.001^{**}$      & $0.256$ \\
 &                     & Unseen & 51.0 & 62.4 & 51.2  & $0.037^{*}$       & $0.084$ \\
\midrule
\multirow{6}{*}{\begin{turn}{90}Qwen3-8B\end{turn}}
 & \multirow{2}{*}{Baseline} & Seen   & \multicolumn{3}{c}{55.4} & -- & -- \\
 &                           & Unseen & \multicolumn{3}{c}{51.2} & -- & -- \\
\cmidrule(l){2-8}
 & \multirow{2}{*}{5}  & Seen   & 59.5 & 59.4 & 71.9  & $0.951$           & $<\!0.001^{***}$ \\
 &                     & Unseen & 53.8 & 59.0 & 58.5  & $0.017^{*}$       & $0.803$ \\
\cmidrule(l){2-8}
 & \multirow{2}{*}{10} & Seen   & 60.4 & 73.6 & 99.2  & $<\!0.001^{***}$  & $<\!0.001^{***}$ \\
 &                     & Unseen & 59.0 & 68.9 & 65.6  & $<\!0.001^{***}$  & $0.334$ \\
\midrule
\multirow{6}{*}{\begin{turn}{90}Qwen3.5-9B\end{turn}}
 & \multirow{2}{*}{Baseline} & Seen   & \multicolumn{3}{c}{55.5} & -- & -- \\
 &                           & Unseen & \multicolumn{3}{c}{55.1} & -- & -- \\
\cmidrule(l){2-8}
 & \multirow{2}{*}{5}  & Seen   & 55.4 & 70.4 & 99.5  & $<\!0.001^{***}$  & $<\!0.001^{***}$ \\
 &                     & Unseen & 56.4 & 69.0 & 73.6  & $<\!0.001^{***}$  & $0.007^{**}$ \\
\cmidrule(l){2-8}
 & \multirow{2}{*}{10} & Seen   & 62.3 & 77.6 & 100.0 & $<\!0.001^{***}$  & $<\!0.001^{***}$ \\
 &                     & Unseen & 63.3 & 74.7 & 75.0  & $0.009^{**}$      & $0.754$ \\
\bottomrule
\end{tabular}
\end{table}

\FloatBarrier
\section{Decontamination}

\subsection{Decontamination methods}
\label{app:decontam_methods}

Table~\ref{tab:decontamination_methods_dense} shows the decontamination methods used in model releases from Anthropic, OpenAI, Meta, AllenAI, Gemini, Deepseek and Qwen. Newer releases either do not mention decontamination procedures or refer to the older releases for such details. Where a method has been used (\checkmark) we add the section where it is mentioned, for further verifiability. It is clear that the standard is n-grams and a mix of exact and subset string matching for decontamination methods.

\begin{table}[t]
\centering
\caption{\textbf{Disclosed decontamination procedures in SOTA model releases.} Every single named method is lexical; reported n-gram sizes range from 10 to 20 with no consensus; no surveyed release reports MinHash at the decontamination stage; the sole semantic mention (Gemini 2.5) names no method or implementation details. Notes: $^{*}$\,=\,remaining part of the pipeline is gated on the initial n-grams getting hits, N/A\,=\,the paper does not include said details.}
{\small
\setlength{\tabcolsep}{4pt}
\begin{tabular}{l l c c c c c}
\toprule
Model & Where? & Exact & Substring & n-gram & MinHash & Semantic \\
\midrule
Claude Opus 4.5 & \S2.2~\cite{anthropic2025opus45systemcard} & \checkmark & \checkmark & \checkmark{} (20) & -- & -- \\
Llama 3         & \S5.2~\cite{grattafiori2024llama}         & \checkmark & --         & --                & -- & -- \\
OLMo 3          & \S3.5.3~\cite{olmo2025olmo3}              & --         & --         & \checkmark{} (13)$^{*}$ & -- & -- \\
Qwen 2.5-Math   & \S4~\cite{yang2024qwen2}                  & --         & --         & \checkmark{} (13) & -- & -- \\
GPT-4           & App.~C~\cite{achiam2023gpt}               & --         & \checkmark & --                & -- & -- \\
GPT-3           & App.~C~\cite{brown2020language}           & --         & --         & \checkmark{} (13) & -- & -- \\
Gemini 2.5      & \S3~\cite{comanici2025gemini}             & --         & --         & \checkmark        & -- & \checkmark \\
DeepSeek-R1     & App.~D.1~\cite{guo2025deepseek}           & --         & --         & \checkmark{} (10) & -- & -- \\
Qwen            & \S5.2.6~\cite{bai2023qwen}                & --         & --         & \checkmark{} (13) & -- & -- \\
\bottomrule
\end{tabular}
}
\label{tab:decontamination_methods_dense}
\end{table}

\subsection{Types of semantic duplicate contamination}
Here we attempt to categorise the semantic duplicates we find in the wild for MBPP and CodeForces (coding test items). We classify 2,350 pairs: all 1,150 confirmed CodeForces pairs, and 1,200 MBPP pairs sampled at a cap of 5 per benchmark task. Because duplicate status was already established by the pipeline, the classifier (a strong LLM, forced single-choice over a fixed, mutually exclusive taxonomy) adjudicates only the form of duplication, not whether the pair is a duplicate.

We state the definition for each of the seven forms, ordered by increasing semantic distance between the two samples, from a standard paraphrase to different problem statements with identical algorithm:
\begin{itemize}
  \item \textbf{Verbatim restatement:} the corpus text reproduces the benchmark item essentially word-for-word, but not enough so that standard deduplication procedures would catch it. Many papers examine the effects of these, including ours.
  \item \textbf{Same-language paraphrase:} the same problem reworded in the same natural language (and, for code, the same programming language). This was previously well studied in the forms of semantic duplication.
  \item \textbf{Cross-programming-language:} the same task and solution rendered in a different programming language than the benchmark's canonical one.
  \item \textbf{Cross-natural-language:} the same problem written in a different natural language.
  \item \textbf{Logical complement:} the corpus item states the complementary problem; rendering logic for a solution is identical, up to complement.
  \item \textbf{Corpus strictly generalizes tests:} the corpus problem is a strict generalization; solving it solves the benchmark item as a special case (sort a list of length $N$, versus sort a list of length 5). We are careful of these, since certain problems require more efficient solutions than the general solution. Only when the general solution strictly covers the specific solution, up to triviality, do we consider this a semantic duplicate.
  \item \textbf{Identical algorithm, different problem statement:} surface-different statements solved by the same underlying algorithm, so a model that has learned one effectively has the other. That is, the training data contains in it the solution to the test problem, even if the problem statement isn't easily identifiable as the same problem.
\end{itemize}

Table~\ref{tab:prelim_sem_dupe_classification} shows the rates we find for each type.

\begin{table}[t]
  \centering
  \caption{A stratification of semantic duplicates detected in the wild by the main type of semantic duplication, which includes mechanisms such as language and logic relationship.}
  \label{tab:prelim_sem_dupe_classification}
  {\small
  \begin{tabular}{lrr}
    \toprule
    \textbf{Form of duplication} & \textbf{CodeForces} & \textbf{MBPP} \\
     & \textbf{($n=1{,}150$)} & \textbf{($n=1{,}200$)} \\
    \midrule
    Verbatim restatement                               & 15.3\% &  2.2\% \\
    Same-language paraphrase                           &  9.9\% & 19.3\% \\
    Cross-programming-language                         &  0.3\% &  7.9\% \\
    Cross-natural-language                             &  1.6\% &  1.0\% \\
    Logical complement                                 &  1.6\% &  4.1\% \\
    Corpus strictly generalizes test                   &  8.0\% & 22.6\% \\
    Identical algorithm, different problem statement   & 63.2\% & 42.9\% \\
    \bottomrule
  \end{tabular}
  }
\end{table}

\clearpage

\lstdefinestyle{mono}{
  basicstyle       = \ttfamily\tiny,
  breaklines       = true,
  showstringspaces = false,
  tabsize          = 4,
  numbers          = none,
  xleftmargin      = 0pt,
  aboveskip        = 4pt,
  belowskip        = 0pt,
  frame            = none,
}

\section{More qualitative examples labeled by our pipeline}\label{app:more-sem-dupe-examples}

\setlength{\columnsep}{0.8em}

\subsection{MBPP}

\subsubsection*{Example 1}
\label{mbpp:1}
Test sample ID: {\footnotesize \texttt{MBPP (125)}}.\qquad Match type: {\footnotesize \texttt{subset}}.

Corpus sample ID: {\footnotesize \texttt{dolci\_100pct,}}{\scriptsize \texttt{e581fb0748c03cf9c80f70cf5b371514706c073c0b3f4db7ae4aa4ccb47d9332}}.
\begin{paracol}{2}
\begin{testbox}
\scriptsize
Write a function to find the maximum difference between the number of 0s and number of 1s in any sub-string of the given binary string.

\lstinputlisting[style=mono, frame=none]{pair1_test_code.py}
\end{testbox}\switchcolumn\begin{corpusbox}
\scriptsize
Prompt: Given an array of integers, write a function that returns the maximum sum of any contiguous subarray within the array. For example, given the input array [2, 4, -3, 5], the function should return 11, which is the sum of the subarray [2, 4, -3, 5].

Response:
\lstinputlisting[style=mono, frame=none]{pair1_code.py}
\end{corpusbox}

\end{paracol}

\subsubsection*{Example 2}
\label{mbpp:2}
Test sample ID: {\footnotesize \texttt{MBPP (412)}}.\qquad Match type: {\footnotesize \texttt{equivalent}}.

Corpus sample ID: {\footnotesize \texttt{dolci\_100pct,}}{\scriptsize \texttt{a2838fb2d645df8018072f6fec43f3c40710d121692be022aab15eafe1d761ac}}.
\begin{paracol}{2}
\begin{testbox}
\scriptsize
Write a python function to remove odd numbers from a given list.

\begin{lstlisting}[style=mono, frame=none]
def remove_odd(l):
    for i in l:
        if i % 2 != 0:
            l.remove(i)
    return l
\end{lstlisting}
\end{testbox}\switchcolumn\begin{corpusbox}
\scriptsize
Prompt: You are tasked with creating a Python function that takes a list of integers and returns a new list containing only the even numbers from the original list. If there are no even numbers, the function should return an empty list. Implement the function \texttt{filter\_even\_numbers(arr: List[int]) -> List[int]} where \texttt{arr} is the input list of integers. The output should be a list of even integers extracted from the input list in the order they were encountered.

Response:
\lstinputlisting[style=mono, frame=none]{pair2_code.py}
\end{corpusbox}
\end{paracol}

\subsubsection*{Example 3}
\label{mbpp:3}
Test sample ID: {\footnotesize \texttt{MBPP (431)}}.\qquad Match type: {\footnotesize \texttt{subset}}.

Corpus sample ID: {\footnotesize \texttt{dolci\_100pct,}}{\scriptsize \texttt{84aaf5508ea33e8ca439094fb305328bf000127e3c22a65777bf7fdfac61ca80}}.
\begin{paracol}{2}
\begin{testbox}
\scriptsize
Write a function that takes two lists and returns true if they have at least one common element.

\lstinputlisting[style=mono, frame=none]{pair3_test.py}
\end{testbox}\switchcolumn\begin{corpusbox}
\scriptsize
Prompt: Given two collections of elements in the form of lists, you're required to produce a new list. This resultant list has to manifest exclusively those elements found identically present within both lists. The task involves multiple steps: identify, compare, and compile. The two source lists are as follow:

\texttt{list1 = [1, 2, 3, 4, 5]}\\
\texttt{list2 = [3, 4, 5, 6, 7]}

Response: You can solve this task in Python quite easily using list comprehension, here is how:

\lstinputlisting[style=mono, frame=none]{pair3_code.py}

When you run this program, it will print:

\texttt{[3, 4, 5]}

Explanation:

The expression \texttt{x for x in list1 if x in list2} goes through each element \texttt{x} in \texttt{list1} and checks \texttt{if x in list2}. If \texttt{True}, the element \texttt{x} is included in the \texttt{result} list.

Also, if you want the resulting list to be unique (i.e., no duplicate elements), you can change the list to a set which automatically removes duplicates in Python:

\lstinputlisting[style=mono, frame=none]{pair3_code_alt.py}

This will give you the same output but ensure that there are no duplicates.
\end{corpusbox}
\end{paracol}

\subsubsection*{Example 4}
\label{mbpp:4}
Test sample ID: {\footnotesize \texttt{MBPP (82)}}.\qquad Match type: {\footnotesize \texttt{equivalent}}.

Corpus sample ID: {\footnotesize \texttt{dolma\_1pct,}}{\scriptsize \texttt{ce549f574ae55532}}.
\begin{paracol}{2}
\begin{testbox}
\scriptsize
Write a function to find the volume of a sphere.

\lstinputlisting[style=mono, frame=none]{pair4_test_code.py}
\end{testbox}\switchcolumn\begin{corpusbox}
\scriptsize
\lstinputlisting[style=mono, frame=none, language=C++]{pair4_code.cpp}
\end{corpusbox}
\end{paracol}

\subsubsection*{Example 5}
\label{mbpp:5}
Test sample ID: {\footnotesize \texttt{MBPP (131)}}.\qquad Match type: {\footnotesize \texttt{equivalent}}.

Corpus sample ID: {\footnotesize \texttt{dolma\_1pct,}}{\scriptsize \texttt{4ec4fea5f7357828}}.
\begin{paracol}{2}
\begin{testbox}
\scriptsize
Write a python function to reverse only the vowels of a given string (where y is not a vowel).

\lstinputlisting[style=mono, frame=none]{pair5_test.py}
\end{testbox}\switchcolumn\begin{corpusbox}
\scriptsize
\lstinputlisting[style=mono, frame=none]{pair5_code.java}
\end{corpusbox}
\end{paracol}

\subsubsection*{Example 6}
\label{mbpp:6}
Test sample ID: {\footnotesize \texttt{MBPP (249)}}.\qquad Match type: {\footnotesize \texttt{subset}}.

Corpus sample ID: {\footnotesize \texttt{dolci\_100pct,}}{\scriptsize \texttt{0d3fd59142d8b0152ba5bb18935497e997d0811a06d860c7911b0af790e3aafe}}.
\begin{paracol}{2}
\begin{testbox}
\scriptsize
Write a function to find the intersection of two arrays.

\begin{lstlisting}[style=mono, frame=none]
def intersection_array(array_nums1,array_nums2):
    result = list(filter(lambda x: x in array_nums1, array_nums2))
    return result
\end{lstlisting}
\end{testbox}\switchcolumn\begin{corpusbox}
\scriptsize
Prompt: Write a program that reads two lists of integers from standard input and prints a list of their common elements.

Input:
\begin{itemize}\setlength\itemsep{0pt}
\item The first line contains space-separated integers representing the first list.
\item The second line contains space-separated integers representing the second list.
\end{itemize}

Output:
\begin{itemize}\setlength\itemsep{0pt}
\item Output, on a single line, a list of integers in the same order as they appear in the first input list, containing only the elements that are also in the second list. Print the list using the standard Python list format (e.g., \texttt{[2, 3]}).
\end{itemize}

Example:

Input:\\
\texttt{1 2 3 4}\\
\texttt{2 3 5 6}

Output:\\
\texttt{[2, 3]}

Response:
\lstinputlisting[style=mono, frame=none]{pair6_code.py}
\end{corpusbox}
\end{paracol}

\subsubsection*{Example 7}
\label{mbpp:7}
Test sample ID: {\footnotesize \texttt{MBPP (475)}}.\qquad Match type: {\footnotesize \texttt{equivalent}}.

Corpus sample ID: {\footnotesize \texttt{dolma\_1pct,}}{\scriptsize \texttt{160310d03a10d7fc}}.
\begin{paracol}{2}
\begin{testbox}
\scriptsize
Write a function to sort a dictionary by value.
\begin{lstlisting}[style=mono, frame=none]
from collections import Counter
def sort_counter(dict1):
    x = Counter(dict1)
    sort_counter = x.most_common()
    return sort_counter
\end{lstlisting}
\end{testbox}\switchcolumn\begin{corpusbox}
\scriptsize
{\ttfamily\# \begin{CJK}{UTF8}{gbsn}请按字典中的\end{CJK}value\begin{CJK}{UTF8}{gbsn}值进行排序\end{CJK}}
\lstinputlisting[style=mono, frame=none]{pair7_code.py}
\end{corpusbox}
\end{paracol}

\subsubsection*{Example 8}
\label{mbpp:8}
Test sample ID: {\footnotesize \texttt{MBPP (428)}}.\qquad Match type: {\footnotesize \texttt{equivalent}}.

Corpus sample ID: {\footnotesize \texttt{dolma\_1pct,}}{\scriptsize \texttt{a9ff098110261d30}}.
\begin{paracol}{2}
\begin{testbox}
\scriptsize
Write a function to sort the given array by using shell sort.
\lstinputlisting[style=mono, frame=none]{pair8_test_code.py}
\end{testbox}\switchcolumn\begin{corpusbox}
\scriptsize
\lstinputlisting[style=mono, frame=none]{pair8_code.js}
\end{corpusbox}
\end{paracol}

\subsubsection*{Example 9}
\label{mbpp:9}
Test sample ID: {\footnotesize \texttt{MBPP (65)}}.\qquad Match type: {\footnotesize \texttt{subset}}.

Corpus sample ID: {\footnotesize \texttt{dolci\_100pct,}}{\scriptsize \texttt{f98cda41dfe2a9ced6b9fb7843db3b25dc9e49ae042c6f6bc68675e4f49a0937}}.
\begin{paracol}{2}
\begin{testbox}
\scriptsize
Write a function to flatten a list and sum all of its elements.

\begin{lstlisting}[style=mono, frame=none]
def recursive_list_sum(data_list):
    total = 0
    for element in data_list:
        if type(element) == type([]):
            total = total + recursive_list_sum(element)
        else:
            total = total + element
    return total
\end{lstlisting}
\end{testbox}\switchcolumn\begin{corpusbox}
\scriptsize
Prompt: Implement a function that calculates the total sum of all integers within a nested list structure. The list may contain integers, strings representing integers, and other lists. Your function should handle any level of nesting. For example, the input \lstinline[style=mono]|['1', '2', 3, [4, '5']]| should return 15. The function signature should be: \lstinline[style=mono]|def sum_nested(lst: List[Union[int, str, List]]) -> int:|

Response:
\lstinputlisting[style=mono, frame=none]{pair9_code.py}
\end{corpusbox}
\end{paracol}

\subsubsection*{Example 10}
\label{mbpp:10}
Test sample ID: {\footnotesize \texttt{MBPP (297)}}.\qquad Match type: {\footnotesize \texttt{equivalent}}.

Corpus sample ID: {\footnotesize \texttt{dolci\_100pct,}}{\scriptsize \texttt{e04feb161678637b214f7f504bc70157685665afbebdb1deefeb7ebbf240875f}}.
\begin{paracol}{2}
\begin{testbox}
\scriptsize
Write a function to flatten a given nested list structure.

\lstinputlisting[style=mono, frame=none]{pair10_test_code.py}
\end{testbox}\switchcolumn\begin{corpusbox}
\scriptsize
Prompt: Given a nested list of integers, write a function \texttt{flatten\_list(nested\_list)} that takes a single argument \texttt{nested\_list}, which can contain integers and other lists (which can themselves contain integers or more lists). Your function should return a flat list containing all the integers from the nested structure in the order they appear. If the input list is empty, return an empty list. Note that the input will never contain strings or bytes, only integers and lists of integers. For example, calling \texttt{flatten\_list([1, [2, 3], [[4], 5]])} should return \texttt{[1, 2, 3, 4, 5]}.

Response:
\lstinputlisting[style=mono, frame=none]{pair10_code.py}
\end{corpusbox}
\end{paracol}

\subsection{CodeForces}

\subsubsection*{Example 1}
\label{codeforces:1}
Test sample ID: {\footnotesize \texttt{CodeForces (1987/D)}}.\qquad Match type: {\footnotesize \texttt{exact}}.

Corpus sample ID: {\footnotesize \texttt{dolci\_sft,}}{\scriptsize \texttt{55333872e569f21088eac6eef7aa8093d7ee90d223bc9865eafbcec4e8faf178}}.
\begin{testbox}
\scriptsize
Alice and Bob are playing a game. Initially, there are $n$ cakes, with the $i$-th cake having a tastiness value of $a_i$.

Alice and Bob take turns eating them, with Alice starting first:
\begin{itemize}\setlength\itemsep{0pt}
\item In her turn, Alice chooses and eats any remaining cake whose tastiness is strictly greater than the maximum tastiness of any of the cakes she's eaten before that. Note that on the first turn, she can choose any cake.
\item In his turn, Bob chooses any remaining cake and eats it.
\end{itemize}

The game ends when the current player can't eat a suitable cake. Let $x$ be the number of cakes that Alice ate. Then, Alice wants to maximize $x$, while Bob wants to minimize $x$.

Find out how many cakes Alice will eat if both players play optimally.

Each test contains multiple test cases. The first line of input contains a single integer $t$ ($1 \le t \le 500$) --- the number of test cases. The description of the test cases follows.

The first line of each test case contains a single integer $n$ ($1 \le n \le 5000$) --- the number of cakes.

The second line of each test case contains $n$ integers $a_1, a_2, \ldots, a_n$ ($1 \le a_i \le n$) --- the tastiness values of the cakes.

It is guaranteed that the sum of $n$ over all test cases does not exceed $5000$.

For each test case, output a single integer --- the number of cakes Alice will eat if both players play optimally.
\end{testbox}
\begin{corpusbox}
\scriptsize
Prompt: Alice and Bob are playing a game with cakes.

Input

The first line contains an integer $t$ ($1 \le t \le 500$) -- the number of test cases. For each test case:
\begin{itemize}\setlength\itemsep{0pt}
\item The first line contains an integer $n$ ($1 \le n \le 5000$) -- the number of cakes.
\item The second line contains $n$ integers $a_1 \ldots a_n$ ($1 \le a_i \le n$) -- the tastiness of each cake.
\end{itemize}
The sum of $n$ over all test cases does not exceed $5000$.

Rules of the game
\begin{itemize}\setlength\itemsep{0pt}
\item Alice moves first, then they alternate.
\item When it is Alice's turn she must choose a remaining cake whose tastiness is strictly greater than every cake she has previously eaten. On her first turn she may take any cake.
\item When it is Bob's turn he may take any remaining cake.
\item If a player cannot move the game ends.
\end{itemize}

Both players play optimally: Alice tries to maximize the number $x$ of cakes she eats, Bob tries to minimize $x$.

Output

For each test case output a single integer -- the value of $x$ when both play optimally.

Response: To solve this problem, we need to simulate a game where \textbf{Alice and Bob} take turns eating cakes under specific rules:
\begin{itemize}\setlength\itemsep{0pt}
\item \textbf{Alice} can only eat a cake that has \textbf{strictly greater tastiness} than the last one she ate.
\item \textbf{Bob} can eat any remaining cake.
\item The goal is to \textbf{maximize the number of cakes Alice eats}, assuming both play optimally.
\end{itemize}

\par\noindent\rule{\linewidth}{0.4pt}\par

\textbf{Key Observations:}
\begin{enumerate}\setlength\itemsep{0pt}
\item \textbf{Alice's constraint} is that she must eat strictly increasing tastiness values.
\item \textbf{Bob's move} is to reduce Alice's future options by taking a cake that might be useful for her.
\item Since the total number of cakes is small ($\le 5000$), we can use \textbf{dynamic programming with memoization}.
\end{enumerate}

\par\noindent\rule{\linewidth}{0.4pt}\par

\textbf{Strategy:}

We define a recursive function with memoization:

\texttt{def dfs(last\_a, remaining\_cakes)}

Where:
\begin{itemize}\setlength\itemsep{0pt}
\item \texttt{last\_a} is the last tastiness value that Alice ate.
\item \texttt{remaining\_cakes} is a list of remaining cake tastiness values (sorted for easier processing).
\end{itemize}

At each step:
\begin{itemize}\setlength\itemsep{0pt}
\item If it's \textbf{Alice's turn}, she can only pick a cake with tastiness \textbf{greater than \texttt{last\_a}}.
\item If it's \textbf{Bob's turn}, he can pick \textbf{any} remaining cake.
\end{itemize}

We simulate both players' turns, alternating, and memoize the result based on:
\begin{itemize}\setlength\itemsep{0pt}
\item The set of remaining cakes (as a frozenset for memoization).
\item The last value Alice ate.
\end{itemize}

\par\noindent\rule{\linewidth}{0.4pt}\par

\textbf{Python Code:}

\lstinputlisting[style=mono, frame=none]{pair1_code.py}

\par\noindent\rule{\linewidth}{0.4pt}\par

\textbf{Explanation:}
\begin{itemize}\setlength\itemsep{0pt}
\item We \textbf{sort the cakes} to process in increasing order (easier to handle Alice's increasing constraint).
\item We use \textbf{memoization} via \texttt{@lru\_cache} to avoid recomputation.
\item We simulate all possible first moves for Alice and pick the one that gives her the \textbf{maximum number of cakes}.
\end{itemize}

\par\noindent\rule{\linewidth}{0.4pt}\par

\textbf{Time Complexity:}
\begin{itemize}\setlength\itemsep{0pt}
\item For each test case, the number of unique tastiness values is at most $n$, and the number of states in the DP is $O(n^2)$ in the worst case.
\item Since the total $n$ across all test cases is $\le 5000$, this is efficient enough.
\end{itemize}

\par\noindent\rule{\linewidth}{0.4pt}\par

\textbf{Example:}

\textbf{Input:}
\begin{lstlisting}[style=mono, frame=none]
1
5
1 2 3 4 5
\end{lstlisting}

\textbf{Output:}
\begin{lstlisting}[style=mono, frame=none]
3
\end{lstlisting}

\textbf{Explanation:}
\begin{itemize}\setlength\itemsep{0pt}
\item Alice can eat 1, 3, 5 (or similar increasing sequences).
\item Bob will block the best options to minimize Alice's total.
\end{itemize}

\par\noindent\rule{\linewidth}{0.4pt}\par

Let me know if you'd like a version with full test case handling or more explanation!
\end{corpusbox}

\subsubsection*{Example 2}
\label{codeforces:2}
Test sample ID: {\footnotesize \texttt{CodeForces (1984/D)}}.\qquad Match type: {\footnotesize \texttt{exact}}.

Corpus sample ID: {\footnotesize \texttt{dolci\_sft,}}{\scriptsize \texttt{32954b54484de7058cb6a10376ed889752a1f60f534eac8192a7c55c2d3cdf26}}.
\begin{testbox}
\scriptsize
You are given a string $s$ consisting of lowercase Latin characters. Count the number of nonempty strings $t \neq$ \texttt{"a"} such that it is possible to partition$^\dagger$ $s$ into some substrings satisfying the following conditions:
\begin{itemize}
\setlength\itemsep{0pt}
\item each substring either equals $t$ or \texttt{"a"}, and
\item at least one substring equals $t$.
\end{itemize}

$^\dagger$ A partition of a string $s$ is an ordered sequence of some $k$ strings $t_1, t_2, \ldots, t_k$ (called substrings) such that $t_1 + t_2 + \cdots + t_k = s$, where $+$ represents the concatenation operation.

The first line contains a single integer $t$ ($1 \le t \le 10^4$) --- the number of test cases.

The only line of each test case contains a string $s$ consisting of lowercase Latin characters ($2 \le |s| \le 2 \times 10^5$).

The sum of $|s|$ over all test cases does not exceed $3 \times 10^5$.

For each test case, output a single integer --- the number of nonempty strings $t \neq$ \texttt{"a"} that satisfy all constraints.
\end{testbox}
\begin{corpusbox}
\scriptsize
Prompt: You are given a string $s$ consisting only of lowercase Latin letters.

Your task is to count how many non-empty strings $t$ (with $t \neq$ \texttt{"a"}) have the following property:

It is possible to split $s$ into consecutive substrings so that every substring is either exactly $t$ or exactly \texttt{"a"}, and at least one of the substrings is $t$.

Input

The first line contains an integer $t$ ($1 \le t \le 10^4$) --- the number of test cases.

Each of the next $t$ lines contains a string $s$ ($2 \le |s| \le 2 \cdot 10^5$). The total length of all given strings does not exceed $3 \cdot 10^5$.

Output

For each test case output a single integer --- the number of different strings $t$ satisfying the conditions for the corresponding $s$.

Response:
\lstinputlisting[style=mono, frame=none]{pair2_code.py}
\end{corpusbox}

\subsubsection*{Example 3}
\label{codeforces:3}
Test sample ID: {\footnotesize \texttt{CodeForces (1978/C)}}.\qquad Match type: {\footnotesize \texttt{equivalent}}.

Corpus sample ID: {\footnotesize \texttt{dolci\_sft,}}{\scriptsize \texttt{311e326c774442b2cac0b1665be99dfc9ca0b884f6358a713a0c5fe33058a891}}.
\begin{testbox}
\scriptsize
Let's call the Manhattan value of a permutation$^\dagger$ $p$ the value of the expression $|p_1 - 1| + |p_2 - 2| + \cdots + |p_n - n|$.

For example, for the permutation $[1, 2, 3]$, the Manhattan value is $|1 - 1| + |2 - 2| + |3 - 3| = 0$, and for the permutation $[3, 1, 2]$, the Manhattan value is $|3 - 1| + |1 - 2| + |2 - 3| = 2 + 1 + 1 = 4$.

You are given integers $n$ and $k$. Find a permutation $p$ of length $n$ such that its Manhattan value is equal to $k$, or determine that no such permutation exists.

$^\dagger$A permutation of length $n$ is an array consisting of $n$ distinct integers from $1$ to $n$ in arbitrary order. For example, $[2,3,1,5,4]$ is a permutation, but $[1,2,2]$ is not a permutation ($2$ appears twice in the array), and $[1,3,4]$ is also not a permutation ($n=3$ but there is $4$ in the array).

Each test consists of multiple test cases. The first line contains a single integer $t$ ($1 \le t \le 10^{4}$) --- the number of test cases. The description of the test cases follows.

The only line of each test case contains two integers $n$ and $k$ ($1 \le n \le 2 \times 10^{5}$, $0 \le k \le 10^{12}$) --- the length of the permutation and the required Manhattan value.

It is guaranteed that the sum of $n$ over all test cases does not exceed $2 \times 10^{5}$.

For each test case, if there is no suitable permutation, output \texttt{"No"}. Otherwise, in the first line, output \texttt{"Yes"}, and in the second line, output $n$ distinct integers $p_1, p_2, \ldots, p_n$ ($1 \le p_i \le n$) --- a suitable permutation.

If there are multiple solutions, output any of them.

You can output the answer in any case (for example, the strings \texttt{"yEs"}, \texttt{"yes"}, \texttt{"Yes"}, and \texttt{"YES"} will be recognized as a positive answer).
\end{testbox}
\begin{corpusbox}
\scriptsize
Prompt: We define the Manhattan value of a permutation $p$ of length $n$ as
$|p_1-1|+|p_2-2|+\cdots+|p_n-n|$.

Given integers $n$ and $k$ you must decide whether there exists a permutation of $1\ldots n$ whose Manhattan value equals $k$.

Input (via standard input)

The first line contains $t$ ($1 \le t \le 10^4$) -- the number of test cases.

Each of the next $t$ lines contains two integers $n$ and $k$ ($1 \le n \le 2 \cdot 10^5$, $0 \le k \le 10^{12}$).

The sum of $n$ over all test cases does not exceed $2 \cdot 10^5$.

Output (via standard output)

For every test case output either

No

if no such permutation exists, or

Yes

$p_1\;p_2\;\ldots\;p_n$

in two separate lines if a suitable permutation exists. Any valid permutation may be printed.

Response:
\lstinputlisting[style=mono, frame=none]{pair3_code.py}
\end{corpusbox}

\subsubsection*{Example 4}
\label{codeforces:4}
Test sample ID: {\footnotesize \texttt{CodeForces (1996/B)}}.\qquad Match type: {\footnotesize \texttt{equivalent}}.

Corpus sample ID: {\footnotesize \texttt{dolci\_sft,}}{\scriptsize \texttt{0dbc4ecb2bbe6499ac5bc74522c971243f1bc0492c25048aaae49ca59fa5da17}}.
\begin{testbox}
\scriptsize
Tina has a square grid with $n$ rows and $n$ columns. Each cell in the grid is either 0 or 1.

Tina wants to reduce the grid by a factor of $k$ ($k$ is a divisor of $n$). To do this, Tina splits the grid into $k \times k$ nonoverlapping blocks of cells such that every cell belongs to exactly one block.

Tina then replaces each block of cells with a single cell equal to the value of the cells in the block. It is guaranteed that every cell in the same block has the same value.

For example, the following demonstration shows a grid being reduced by factor of 3.

Original grid \texttt{000111000111000111111000111000111000}

Reduced grid \texttt{0110}

Help Tina reduce the grid by a factor of $k$.

The first line contains $t$ ($1 \le t \le 100$) --- the number of test cases.

The first line of each test case contains two integers $n$ and $k$ ($1 \le n \le 1000$, $1 \le k \le n$, $k$ is a divisor of $n$) --- the number of rows and columns of the grid, and the factor that Tina wants to reduce the grid by.

Each of the following $n$ lines contain $n$ characters describing the cells of the grid. Each character is either 0 or 1. It is guaranteed every $k$ by $k$ block has the same value.

It is guaranteed the sum of $n$ over all test cases does not exceed 1000.

For each test case, output the grid reduced by a factor of $k$ on a new line.
\end{testbox}
\begin{corpusbox}
\scriptsize
Prompt: Tina has a square grid with $n$ rows and $n$ columns ($1 \le n \le 1000$). Every cell contains either the character 0 or 1.

Tina wants to reduce the grid by a factor of $k$, where $k$ is a divisor of $n$ ($1 \le k \le n$). She divides the grid into non-overlapping $k \times k$ blocks so that each cell belongs to exactly one block. It is guaranteed that within every $k \times k$ block all cells have the same value. Each block is replaced by a single cell equal to that common value, producing a new grid of size $(n / k) \times (n / k)$.

Input

The first line contains $t$ ($1 \le t \le 100$) --- the number of test cases.

For each test case:
\begin{itemize}\setlength\itemsep{0pt}
\item The first line contains two integers $n$ and $k$.
\item The next $n$ lines each contain a string of $n$ characters (0 or 1) describing the grid.
\end{itemize}

It is guaranteed that the sum of $n$ over all test cases does not exceed 1000.

Output

For every test case output the reduced grid, each row on its own line. Output test cases one after another without extra blank lines.

Example

Input\\
\texttt{4}\\
\texttt{4 4}\\
\texttt{0000}\\
\texttt{0000}\\
\texttt{0000}\\
\texttt{0000}\\
\texttt{6 3}\\
\texttt{000111}\\
\texttt{000111}\\
\texttt{000111}\\
\texttt{111000}\\
\texttt{111000}\\
\texttt{111000}\\
\texttt{6 2}\\
\texttt{001100}\\
\texttt{001100}\\
\texttt{111111}\\
\texttt{111111}\\
\texttt{110000}\\
\texttt{110000}\\
\texttt{8 1}\\
\texttt{11111111}\\
\texttt{11111111}\\
\texttt{11111111}\\
\texttt{11111111}\\
\texttt{11111111}\\
\texttt{11111111}\\
\texttt{11111111}\\
\texttt{11111111}

Output\\
\texttt{0}\\
\texttt{01}\\
\texttt{10}\\
\texttt{010}\\
\texttt{111}\\
\texttt{100}\\
\texttt{11111111}\\
\texttt{11111111}\\
\texttt{11111111}\\
\texttt{11111111}\\
\texttt{11111111}\\
\texttt{11111111}\\
\texttt{11111111}\\
\texttt{11111111}

Response:
\lstinputlisting[style=mono, frame=none]{pair4_code.py}
\end{corpusbox}

\subsubsection*{Example 5}
\label{codeforces:5}
Test sample ID: {\footnotesize \texttt{CodeForces (1942/C1)}}.\qquad Match type: {\footnotesize \texttt{subset}}.

Corpus sample ID: {\footnotesize \texttt{dolci\_sft,}}{\scriptsize \texttt{239cfdbeca54efec1ef280a8c44dbf2997090d1ff21a5b03287db04769a83811}}.
\begin{testbox}
\scriptsize
This is the easy version of the problem. The only difference between the two versions is the constraint on $y$. In this version $y = 0$. You can make hacks only if both versions are solved.

Bessie has received a birthday cake from her best friend Elsie, and it came in the form of a regular polygon with $n$ sides. The vertices of the cake are numbered from $1$ to $n$ clockwise. You and Bessie are going to choose some of those vertices to cut non-intersecting diagonals into the cake. In other words, the endpoints of the diagonals must be part of the chosen vertices.

Bessie would only like to give out pieces of cake which result in a triangle to keep consistency. The size of the pieces doesn't matter, and the whole cake does not have to be separated into all triangles (other shapes are allowed in the cake, but those will not be counted).

Bessie has already chosen $x$ of those vertices that can be used to form diagonals. She wants you to choose no more than $y$ other vertices such that the number of triangular pieces of cake she can give out is maximized.

What is the maximum number of triangular pieces of cake Bessie can give out?

The first line contains a single integer $t$ ($1 \le t \le 10^4$) --- the number of test cases.

The first line of each test case consists of three integers, $n$, $x$, and $y$ ($4 \le n \le 10^9$, $2 \le x \le \min(n, 2 \times 10^5)$, $y = 0$) --- the number of sides of the polygon, number of vertices Bessie has chosen, and the maximum number of other vertices you can choose.

The second line consists of $x$ distinct integers from $1$ to $n$, representing the vertices Bessie has chosen.

It is guaranteed the sum of $x$ over all test cases does not exceed $2 \times 10^5$.

For each test case, output a single integer: the maximum number of non-intersecting triangular pieces of cake she can give out.
\end{testbox}
\begin{corpusbox}
\scriptsize
Prompt: Birthday Cake Triangles (hard version)\\
\noindent\rule{\linewidth}{0.4pt}

Bessie's cake is a regular polygon with $n$ vertices numbered $1..n$ clockwise ($4 \le n \le 10^9$).
Bessie has already selected $x$ of those vertices ($2 \le x \le \min(n, 2 \cdot 10^5)$).
You may additionally mark at most $y$ other vertices ($0 \le y \le n - x$).
After all marked vertices are fixed we are allowed to draw non-intersecting diagonals but every diagonal must have both endpoints among the marked vertices.
From the regions obtained we only count those that are triangles; other shapes are ignored.
For every test case output the maximum possible number of counted triangles.

Input\\
\noindent\rule{\linewidth}{0.4pt}

The first line contains $t$ ($1 \le t \le 10^4$) -- the number of test cases.

Each test case consists of two lines:
\begin{itemize}\setlength\itemsep{0pt}
\item line 1: $n$ $x$ $y$
\item line 2: $x$ distinct integers -- the vertices already chosen by Bessie (in any order)
\end{itemize}

It is guaranteed that the sum of $x$ over all test cases does not exceed $2 \cdot 10^5$.

Output\\
\noindent\rule{\linewidth}{0.4pt}

For each test case output one line -- the maximum number of triangles that can be obtained.

Example\\
\noindent\rule{\linewidth}{0.4pt}

Input
\begin{lstlisting}[style=mono, frame=none]
3
8 4 2
1 6 2 5
7 3 1
6 4 3
4 2 2
1 3
\end{lstlisting}
Output
\begin{lstlisting}[style=mono, frame=none]
6
5
2
\end{lstlisting}

Response:
\lstinputlisting[style=mono, frame=none]{pair5_code.py}
\end{corpusbox}

\subsubsection*{Example 6}
\label{codeforces:6}
Test sample ID: {\footnotesize \texttt{CodeForces (1986/G2)}}.\qquad Match type: {\footnotesize \texttt{subset}}.

Corpus sample ID: {\footnotesize \texttt{dolci\_sft,}}{\scriptsize \texttt{373fbbd64cb65ba1ddfcb78c89bf2b3e62e8b4eba47f38cacec16a39c7e7c83c}}.
\begin{testbox}
\scriptsize
This is the hard version of the problem. The only difference is that in this version $n \le 5 \times 10^5$ and the sum of $n$ for all sets of input data does not exceed $5 \times 10^5$.

You are given a permutation $p$ of length $n$. Calculate the number of index pairs $1 \le i < j \le n$ such that $p_i \cdot p_j$ is divisible by $i \cdot j$ without remainder.

A permutation is a sequence of $n$ integers, in which each integer from $1$ to $n$ occurs exactly once. For example, $[1]$, $[3,5,2,1,4]$, $[1,3,2]$ are permutations, while $[2,3,2]$, $[4,3,1]$, $[0]$ are not.

Each test consists of several sets of input data. The first line contains a single integer $t$ ($1 \le t \le 10^4$) --- the number of sets of input data. Then follows their description.

The first line of each set of input data contains a single integer $n$ ($1 \le n \le 5 \times 10^5$) --- the length of the permutation $p$.

The second line of each set of input data contains $n$ distinct integers $p_1, p_2, \ldots, p_n$ ($1 \le p_i \le n$) --- the permutation $p$.

It is guaranteed that the sum of $n$ for all sets of input data does not exceed $5 \times 10^5$.

For each set of input data, output the number of index pairs $1 \le i < j \le n$ such that $p_i \cdot p_j$ is divisible by $i \cdot j$ without remainder.
\end{testbox}
\begin{corpusbox}
\scriptsize
Prompt: Chef has an array $A$ of $N$ positive integers. For every ordered pair of indices $(i, j)$ with $1 \le i < j \le N$, Chef is interested in whether the product of the two corresponding array elements is divisible by the product of the indices, i.e.

$A_i \cdot A_j$ is divisible by $i \cdot j$

For each test case output how many pairs $(i, j)$ satisfy the condition.

Input

The first line contains a single integer $T$ -- the number of test cases. For each test case:
\begin{itemize}
\setlength\itemsep{0pt}
\item The first line contains an integer $N$ -- the length of the array ($2 \le N \le 10^5$).
\item The second line contains $N$ space-separated integers $A_1 \ldots A_N$ ($1 \le A_i \le 10^5$).
\end{itemize}
The sum of $N$ over all test cases does not exceed $10^5$.

Output

For every test case print a single line with a single integer: the number of index pairs $(i, j)$ ($1 \le i < j \le N$) for which $A_i \cdot A_j$ is divisible by $i \cdot j$.

Example

Input
\begin{lstlisting}[style=mono, frame=none]
3
3
1 2 3
4
2 3 5 6
5
3 9 14 6 10
\end{lstlisting}
Output
\begin{lstlisting}[style=mono, frame=none]
3
2
6
\end{lstlisting}

Response:
\lstinputlisting[style=mono, frame=none]{pair6_code.py}
\end{corpusbox}

\subsubsection*{Example 7}
\label{codeforces:7}
Test sample ID: {\footnotesize \texttt{CodeForces (1989/B)}}.\qquad Match type: {\footnotesize \texttt{superset}}.

Corpus sample ID: {\footnotesize \texttt{dolci\_sft,}}{\scriptsize \texttt{bc86d7050afb1b4d2336493c5d4cf5c07d3e8d038d45c80e9151d681d064313d}}.
\begin{testbox}
\scriptsize
You are given two strings $a$ and $b$, both consisting of lowercase Latin letters.

A subsequence of a string is a string which can be obtained by removing several (possibly zero) characters from the original string. A substring of a string is a contiguous subsequence of that string.

For example, consider the string \texttt{abac}:
\begin{itemize}\setlength\itemsep{0pt}
\item \texttt{a}, \texttt{b}, \texttt{c}, \texttt{ab}, \texttt{aa}, \texttt{ac}, \texttt{ba}, \texttt{bc}, \texttt{aba}, \texttt{abc}, \texttt{aac}, \texttt{bac} and \texttt{abac} are its subsequences;
\item \texttt{a}, \texttt{b}, \texttt{c}, \texttt{ab}, \texttt{ba}, \texttt{ac}, \texttt{aba}, \texttt{bac} and \texttt{abac} are its substrings.
\end{itemize}

Your task is to calculate the minimum possible length of the string that contains $a$ as a substring and $b$ as a subsequence.

The first line contains a single integer $t$ ($1 \le t \le 10^3$) --- the number of test cases.

The first line of each test case contains a string $a$ ($1 \le |a| \le 100$), consisting of lowercase Latin letters.

The second line of each test case contains a string $b$ ($1 \le |b| \le 100$), consisting of lowercase Latin letters.

For each test case, print a single integer --- the minimum possible length of the string that contains $a$ as a substring and $b$ as a subsequence.
\end{testbox}
\begin{corpusbox}
\scriptsize
Prompt: You are given two strings $a$ and $b$ consisting of lowercase Latin letters. In one operation you may delete the first character or the last character of either string (if the string is not empty). You may perform any number of operations in any order.

After a sequence of operations the two remaining strings (each is some contiguous substring of the original one and may be empty) must become exactly equal. Your task is to find the minimum possible number of operations required to achieve this.

Input

The first line contains an integer $t$ ($1 \le t \le 100$) --- the number of test cases.

For every test case two lines follow: the first line contains the string $a$ ($1 \le |a| \le 20$), the second line contains the string $b$ ($1 \le |b| \le 20$). Both strings consist only of lowercase Latin letters.

Output

For each test case output a single integer --- the minimum number of operations needed to make $a$ and $b$ equal (empty strings are considered equal).

Example

Input

5

a

a

abcd

bc

hello

codeforces

hello

helo

dhjakjsnasjhfksafasd

adjsnasjhfksvdafdser

Output

0

2

13

3

20

Response:
\lstinputlisting[style=mono, frame=none]{pair7_code.py}
\end{corpusbox}

\subsubsection*{Example 8}

\label{codeforces:8}

Test sample ID: {\footnotesize \texttt{CodeForces (1991/F)}}.\qquad Match type: {\footnotesize \texttt{superset}}.

Corpus sample ID: {\footnotesize \texttt{dolci\_sft,}}{\scriptsize \texttt{0361b4c13bf295f2f7293a8917c35b82ea23514708e283101f2638eaae1279c7}}.


\begin{testbox}

\scriptsize

You are given $n$ sticks, numbered from $1$ to $n$. The length of the $i$-th stick is $a_i$.

You need to answer $q$ queries. In each query, you are given two integers $l$ and $r$ ($1 \le l < r \le n$, $r - l + 1 \ge 6$). Determine whether it is possible to choose $6$ distinct sticks from the sticks numbered $l$ to $r$, to form $2$ non-degenerate triangles$^\ast$.

The first line contains two integers $n$ and $q$ ($6 \le n \le 10^5$, $1 \le q \le 10^5$) --- the number of sticks and the number of queries respectively.

The second line contains $n$ integers $a_1$, $a_2$, ..., $a_n$ ($1 \le a_i \le 10^9$) --- $a_i$ denotes the length of the $i$-th stick.

Each of the following $q$ lines contains two integers $l$ and $r$ ($1 \le l < r \le n$, $r - l + 1 \ge 6$) --- the parameters of each query.

For each query, output \texttt{""YES""} (without quotes) if it is possible to form $2$ triangles, and \texttt{""NO""} (without quotes) otherwise.

You can output the answer in any case (upper or lower). For example, the strings \texttt{""yEs""}, \texttt{""yes""}, \texttt{""Yes""}, and \texttt{""YES""} will be recognized as positive responses.

\end{testbox}


\begin{corpusbox}

\scriptsize

Prompt: You are given the lengths of $N$ wooden sticks. You want to know whether you can choose exactly three of them that form a non-degenerate triangle (a triangle with strictly positive area, i.e. the sum of the two shorter sides is strictly greater than the longest side).

Input

The first line contains an integer $N$ ($3 \le N \le 20000$) – the number of sticks collected. The second line contains $N$ positive integers, each strictly less than $2^60$, giving the lengths of the sticks.

Output

Print exactly one line containing the single word \texttt{""possible""} if there exists a triple of sticks that can form a non-degenerate triangle, and \texttt{""impossible""} otherwise.

Example

Input

$3$

$1$ $1$ $1$

Output

possible

Response:

\lstinputlisting[style=mono, frame=none]{pair8_code.py}

\end{corpusbox}


\subsubsection*{Example 9}

\label{codeforces:9}

Test sample ID: {\footnotesize \texttt{CodeForces (1991/D)}}.\qquad Match type: {\footnotesize \texttt{related}}.

Corpus sample ID: {\footnotesize \texttt{dolci\_sft,}}{\scriptsize \texttt{d007fafb1db5925fc9894dba5be6d092c5f2042bb5794c3a6c664b3f0cfd1b02}}.


\begin{testbox}

\scriptsize

You are given an undirected graph with $n$ vertices, numbered from $1$ to $n$. There is an edge between vertices $u$ and $v$ if and only if $u$ XOR $v$ is a prime number, where XOR denotes the bitwise XOR operator.

Color all vertices of the graph using the minimum number of colors, such that no two vertices directly connected by an edge have the same color.

Each test contains multiple test cases. The first line contains the number of test cases $t$ ($1 \le t \le 500$). The description of test cases follows.

The only line contains a single integer $n$ ($1 \le n \le 2 \times 10^5$) --- the number of vertices in the graph.

It is guaranteed that the sum of $n$ over all test cases does not exceed $2 \times 10^5$.

For each test case, output two lines.

The first line should contain a single integer $k$ ($1 \le k \le n$) --- the minimum number of colors required.

The second line should contain $n$ integers $c_1$, $c_2$, $\ldots$, $c_n$ ($1 \le c_i \le k$) --- the color of each vertex.

If there are multiple solutions, output any of them.

\end{testbox}


\begin{corpusbox}

\scriptsize

Prompt: You are given an \textbf{undirected graph} with $54$ vertices, labeled from \texttt{`1`} to \texttt{`54`}. Two vertices $u$ and $v$ are connected by an edge \textbf{if and only if} the absolute difference $|u - v|$ is a prime number.

Your task is to assign a \textbf{non-negative integer color} to each vertex, represented as \texttt{`c[1], c[2], ..., c[54]`}, such that:

\begin{itemize}
\setlength\itemsep{0pt}
\item For every edge $(u, v)$ in the graph, $c[u] \ne c[v]$ --- adjacent vertices must have different colors.

\item The total number of \textbf{distinct colors used} (i.e., the number of unique values among \texttt{`c[1]`} to \texttt{`c[54]`}) is \textbf{minimized} - try your best to find a valid coloring using as few colors as possible.
\end{itemize}

\textbf{Output Format:} Your final answer should be a single line containing the color of each vertex in order: \texttt{`c[1], c[2], ..., c[54]`}, separated by \textbf{spaces}.

Response: We use the well‐known $4$-coloring of the “prime-distance” graph on the integers: since no prime is divisible by $4$ (all primes $p$ satisfy $p$ mod $4 \ne 0$), any two numbers at prime distance have different residues mod $4$. Hence the assignment $c[i]=i$ mod $4$ is a proper coloring using exactly $4$ colors, which is minimal (one can’t do it with $3$ colors because of difference $3$). We now list $c[1]$ through $c[54]$.

Answer:

$1$ $2$ $3$ $0$

$1$ $2$ $3$ $0$

$1$ $2$ $3$ $0$

$1$ $2$ $3$ $0$

$1$ $2$ $3$ $0$

$1$ $2$ $3$ $0$

$1$ $2$ $3$ $0$

$1$ $2$ $3$ $0$

$1$ $2$ $3$ $0$

$1$ $2$ $3$ $0$

$1$ $2$ $3$ $0$

$1$ $2$ $3$ $0$

$1$ $2$ $3$ $0$ $1$ $2$

\end{corpusbox}


\subsubsection*{Example 10}

\label{codeforces:10}

Test sample ID: {\footnotesize \texttt{CodeForces (1971/G)}}.\qquad Match type: {\footnotesize \texttt{related}}.

Corpus sample ID: {\footnotesize \texttt{dolci\_sft,}}{\scriptsize \texttt{7f575b5b3298b8690432a1a8ecf96cb6c97a7d62b45ee7ee27a6e0b2dccf5aa6}}.


\begin{testbox}

\scriptsize

You are given an array $a$ consisting of $n$ nonnegative integers.

You can swap the elements at positions $i$ and $j$ if $a_i~\mathsf{XOR}~a_j < 4$, where $\mathsf{XOR}$ is the bitwise XOR operation.

Find the lexicographically smallest array that can be made with any number of swaps.

An array $x$ is lexicographically smaller than an array $y$ if in the first position where $x$ and $y$ differ, $x_i < y_i$.

The first line contains a single integer $t$ ($1 \le t \le 10^4$) --- the number of test cases.

The first line of each test case contains a single integer $n$ ($1 \le n \le 2\cdot10^5$) --- the length of the array.

The second line of each test case contains $n$ integers $a_i$ ($0 \le a_i \le 10^9$) --- the elements of the array.

It is guaranteed that the sum of $n$ over all test cases does not exceed $2 \times 10^5$.

For each test case, output $n$ integers --- the lexicographically smallest array that can be made with any number of swaps.

\end{testbox}


\begin{corpusbox}

\scriptsize

Prompt: You are given an integer array $a$ of length $n$.

Operation: you may repeatedly pick two indices $i$ and $j$ ($1 \le i, j \le n$) such that $a_i + a_j$ is odd, and swap the two elements $a_i$ and $a_j$.

Determine the lexicographically smallest array that can be obtained after performing any number of such operations.

Input

The first line contains a single integer $n$ ($1 \le n \le 10^5$).

The second line contains $n$ space-separated integers $a_1, a_2, \ldots, a_n$ ($1 \le a_i \le 10^9$).

Output

Output the lexicographically smallest array you can obtain, as $n$ space-separated integers on one line.

An array $x$ is lexicographically smaller than an array $y$ if, at the first position where they differ, the element in $x$ is smaller than the corresponding element in $y$.

Example

Input

3

4 1 7

Output

1 4 7

In the example, swap 4 and 1 (their sum is 5, which is odd) to get the lexicographically smallest array.

Response:

\lstinputlisting[style=mono, frame=none]{pair10_code.py}

\end{corpusbox}


\section*{NeurIPS Paper Checklist}

\begin{enumerate}

\item {\bf Claims}
    \item[] Question: Do the main claims made in the abstract and introduction accurately reflect the paper's contributions and scope?
    \item[] Answer: \answerYes{} 
    \item[] Justification: The abstract mentions our main headline numerical results, emphasises that the main prevalence result is on one open model, and summarises the resulting direction of evidence with respect to our main `shallow generalization' hypothesis. No overclaiming.

\item {\bf Limitations}
    \item[] Question: Does the paper discuss the limitations of the work performed by the authors?
    \item[] Answer: \answerYes{} 
    \item[] Justification: See Section~\ref{sec:limits}

\item {\bf Theory assumptions and proofs}
    \item[] Question: For each theoretical result, does the paper provide the full set of assumptions and a complete (and correct) proof?
    \item[] Answer: \answerNA{} 
    \item[] Justification: No derivations.

    \item {\bf Experimental result reproducibility}
    \item[] Question: Does the paper fully disclose all the information needed to reproduce the main experimental results of the paper to the extent that it affects the main claims and/or conclusions of the paper (regardless of whether the code and data are provided or not)?
    \item[] Answer: \answerYes{} 
    \item[] Justification: See Section~3, Appendix A, and the GitHub repository.

\item {\bf Open access to data and code}
    \item[] Question: Does the paper provide open access to the data and code, with sufficient instructions to faithfully reproduce the main experimental results, as described in supplemental material?
    \item[] Answer: \answerYes{} 
    \item[] Justification: \texttt{\href{https://github.com/AriSpiesberger/Soft-Contamination-Prevelance}{https://github.com/AriSpiesberger/Soft-Contamination-Prevelance}}

\item {\bf Experimental setting/details}
    \item[] Question: Does the paper specify all the training and test details (e.g., data splits, hyperparameters, how they were chosen, type of optimizer) necessary to understand the results?
    \item[] Answer: \answerYes{} 
    \item[] Justification: See Appendix A and B.

\item {\bf Experiment statistical significance}
    \item[] Question: Does the paper report error bars suitably and correctly defined or other appropriate information about the statistical significance of the experiments?
    \item[] Answer: \answerYes{} 
    \item[] Justification: See figures and tables. Appendix A.4 for details.

\item {\bf Experiments compute resources}
    \item[] Question: For each experiment, does the paper provide sufficient information on the computer resources (type of compute workers, memory, time of execution) needed to reproduce the experiments?
    \item[] Answer: \answerYes{} 
    \item[] Justification: Experiment hardware and time is listed in the paper, code supports one or many gpus and will run on other industrial grade GPUS. 
    
\item {\bf Code of ethics}
    \item[] Question: Does the research conducted in the paper conform, in every respect, with the NeurIPS Code of Ethics \url{https://neurips.cc/public/EthicsGuidelines}?
    \item[] Answer: \answerYes{} 
    \item[] Justification: No human subjects, no private, deprecated, or copyrighted data.

\item {\bf Broader impacts}
    \item[] Question: Does the paper discuss both potential positive societal impacts and negative societal impacts of the work performed?
    \item[] Answer: \answerYes{} 
    \item[] Justification: See Impact Statement.
    
\item {\bf Safeguards}
    \item[] Question: Does the paper describe safeguards that have been put in place for responsible release of data or models that have a high risk for misuse (e.g., pre-trained language models, image generators, or scraped datasets)?
    \item[] Answer: \answerNA{} 
    \item[] Justification: No original data or models released.

\item {\bf Licenses for existing assets}
    \item[] Question: Are the creators or original owners of assets (e.g., code, data, models), used in the paper, properly credited and are the license and terms of use explicitly mentioned and properly respected?
    \item[] Answer: \answerYes{} 
    \item[] Justification: All cited appropriately.

\item {\bf New assets}
    \item[] Question: Are new assets introduced in the paper well documented and is the documentation provided alongside the assets?
    \item[] Answer: \answerNA{} 
    \item[] Justification: N/A.

\item {\bf Crowdsourcing and research with human subjects}
    \item[] Question: For crowdsourcing experiments and research with human subjects, does the paper include the full text of instructions given to participants and screenshots, if applicable, as well as details about compensation (if any)? 
    \item[] Answer: \answerNA{} 
    \item[] Justification: N/A

\item {\bf Institutional review board (IRB) approvals or equivalent for research with human subjects}
    \item[] Question: Does the paper describe potential risks incurred by study participants, whether such risks were disclosed to the subjects, and whether Institutional Review Board (IRB) approvals (or an equivalent approval/review based on the requirements of your country or institution) were obtained?
    \item[] Answer: \answerNA{} 
    \item[] Justification: N/A

\item {\bf Declaration of LLM usage}
    \item[] Question: Does the paper describe the usage of LLMs if it is an important, original, or non-standard component of the core methods in this research? Note that if the LLM is used only for writing, editing, or formatting purposes and does \emph{not} impact the core methodology, scientific rigor, or originality of the research, declaration is not required.
    \item[] Answer: \answerYes{} 
    \item[] Justification: We use LLMs as a judge in several places of our pipeline, see Figure~\ref{fig:method_pipeline}.

\end{enumerate}

\end{document}